\documentclass[10pt,journal,compsoc]{IEEEtran}

\usepackage{tikz}
\usepackage{fancyhdr}
\usepackage{times}
\usepackage{epsfig}
\usepackage{graphicx}
\usepackage{amsmath}
\usepackage{amssymb}
\usepackage{multirow}
\usepackage{verbatim}
\usepackage{color}
\usepackage{aliascnt}
\usepackage{amsthm}
\usepackage{caption}
\usepackage{subcaption}
\usepackage{multirow}
\usepackage{wrapfig}
\usepackage{footnote}
\usepackage{cancel}
\usepackage{array}

\usepackage[ruled,titlenotnumbered]{algorithm2e}
\providecommand{\SetAlgoLined}{\SetLine}

\DeclareMathOperator*{\argmin}{arg\,min}

 \makeatletter
\def\ifundefined{\@ifundefined}
\makeatother

\newtheorem{defn}{Definition}
\newtheorem{coro}{Corollary}

\newtheorem{lem}{Lemma}

\newtheorem{prop}{Proposition}
\newtheorem{proper}{Property}
\newtheorem{theor}{Theorem}

\def\utc{\mbox{$\;\overset{c}{=}\;$}}
\def\approxutc{\mbox{$\;\overset{c}{\approx}\;$}}

\def\P{\mbox{${\cal P}$}}

\def\eg{{\em e.g.~}}
\def\ie{{\em i.e.~}}
\def\etc{{\em etc}}

\def\Identity{\mathbf{I}}

\def\S{\mathbf{S}}
\def\barS{\bar{\mathbf{S}}}

\def\dS{\mbox{$d_{\mathbf{s}}$}}
\def\dSk{\mbox{$d^k_{\mathbf{s}}$}}
\def\dSbar{\mbox{$d_{\bar{\mathbf{s}}}$}}
\def\thS{\mbox{$\theta_{\mathbf{s}}$}}

\def\muSk{\mbox{$\mu_{S^k}$}}

\usepackage{MnSymbol}
\def\e{\mbox{$\wideparen{e}$}}   
\def\rp{\mbox{$\gamma$}}           
\def\rs{\mbox{$\beta$}}                
\def\ds{\mbox{$\delta$}}              
\def\pbp{\mbox{$\eta$}}              

\newcommand{\F}{{\cal F}}
\newcommand{\E}{{\mathbf E}}
\newcommand{\N}{{\cal N}}
\def\kp{\mbox{$A$}}

\def\KNN/{\mbox{$K\!N\!N$}}
\def\KM/{{\sf KM}}
\def\pKM/{{\sf pKM}}
\def\kKM/{$k${\sf KM}}
\def\wkKM/{$wk${\sf KM}}
\def\AA/{{\sf AA}}
\def\AD/{{\sf AD}}
\def\NC/{{\sf NC}}
\def\AC/{{\sf AC}}

\usepackage[pagebackref=true,breaklinks=true,letterpaper=true,colorlinks,bookmarks=false]{hyperref}

\begin{document}

\title{{\em Kernel Cuts}: \\ MRF meets Kernel \& Spectral Clustering}

\author{Meng Tang$^*$ \hspace{5ex} Dmitrii Marin$^*$ \hspace{5ex} 
Ismail Ben Ayed$^{\dag}$ \hspace{5ex}  Yuri Boykov$^*$ \\[0.5ex]
{\small $^*$Computer Science, University of Western Ontario, Canada  \hspace{3ex} $^\dag$\'Ecole de Technologie Sup\'erieure, University of Quebec, Canada} \\[-1ex]
{\tt\small mtang73@csd.uwo.ca     dmitrii.a.marin@gmail.com    ismail.benayed@etsmtl.ca   yuri@csd.uwo.ca }
}

\maketitle

\begin{abstract}
We propose a new segmentation model combining common regularization energies, \eg 
Markov Random Field (MRF) potentials, and standard pairwise clustering criteria like Normalized Cut (\NC/),
average association (\AA/), \etc. These clustering and regularization models are widely used
in machine learning and computer vision, but they were not combined before due to significant differences 
in the corresponding optimization, \eg spectral relaxation and combinatorial max-flow techniques. 
On the one hand, we show that many common applications using MRF segmentation energies 
can benefit from a high-order NC term, \eg enforcing balanced clustering of arbitrary high-dimensional 
image features combining color,  texture, location, depth, motion, etc. 
On the other hand, standard clustering applications can benefit from an inclusion of common 
pairwise or higher-order MRF constraints, \eg edge alignment, bin-consistency, label cost, etc. 
To address joint energies like NC+MRF,  we propose efficient {\em Kernel Cut} algorithms based on 
bound optimization. While focusing on graph cut and move-making techniques, our new unary (linear)
{\em kernel} and {\em spectral} bound formulations for common pairwise clustering criteria allow to integrate them 
with any regularization functionals with existing discrete or continuous solvers.
\end{abstract}

\section{Introduction and Motivation} 

The terms {\em clustering} and {\em segmentation} are largely synonyms. 
The latter is common specifically for computer vision when data points are intensities 
or other features $I_p\in\mathcal{R}^N$ sampled at regularly spaced image pixels $p \in \mathcal{R}^M$.
The pixel's location is essential information. Many image segmentation methods treat $I_p$ 
as a function $I:\mathcal{R}^M \rightarrow\mathcal{R}^N$ and process intensity and 
location in fundamentally different ways. This applies to many regularization methods
including discrete MRF-based techniques \cite{Geman:84} and continuous variational methods \cite{Mumford-Shah-89}. 
For example, such methods often use pixel locations to represent the segment's geometry/shape and 
intensities to represent its appearance \cite{Sapiro:97,BJ:01,BK:ICCV03,pock:CVPR09}.

The term {\em clustering} is often used in the general context where data point $I_p$
is an arbitrary observation indexed by integer $p$. Data clustering techniques \cite{Aggarwal:clustering14} 
are widely used outside of vision. Variants of {\em K-means}, spectral or other
clustering methods are also used for image segmentation. They often join the pixel's location and its feature
into one combined data point in $\mathcal{R}^{N+M}$. We focus on a well-known general group of
{\em pairwise clustering} methods \cite{Shi2000} based on some estimated {\em affinities} between all pairs of points.

While independently developed from different methodologies, standard regularization and pairwise clustering methods 
are defined by objective functions that have many complementary properties reviewed later. 
Our goal is to combine these functions into a joint energy applicable to image segmentation or general clustering problems. 
Such objectives could not be combined before due to significant differences in the underlying 
optimization methods, e.g. combinatorial graph cut versus spectral relaxation. 
While focused on MRF regularization, our approach to integrating pairwise clustering is based on a bound formulation
that easily extends to any regularization functional with existing solvers.

We use basic notation applicable to either image segmentation or general data clustering.
Let $\Omega$ be a set of pixels, voxels, or any other points $p$. For example,
for 2D images $\Omega$ could be a subset of regularly spaced points in $\mathcal{R}^2$.
Set $\Omega$ could also represent data points indices. Assume that every 
$p\in\Omega$ comes with an observed feature $I_p\in\mathcal{R}^N$.
For example, $I_p$ could be a greyscale intensity in $\mathcal{R}^1$, 
an RGB color in $\mathcal{R}^3$, or any other high-dimensional observation. 
If needed, $I_p$ could also include the pixel's location.
 
A segmentation of $\Omega$ can be equivalently 
represented either as a labeling $S:=(S_p|p\in\Omega)$ including integer node labels $1 \leq S_p \leq K$ or 
as a partitioning $\{S^k\}$ of set $\Omega$  into $K$ non-overlapping subsets or 
segments $S^k := \{p \in \Omega|S_p=k\}$. With minor abuse of notation, we will also use 
$S^k$ as indicator vectors in $\{0,1\}^{|\Omega|}$.

We combine standard pairwise clustering criteria such as {\em Average Association} (\AA/) or
{\em Normalized Cut} (\NC/) \cite{Shi2000}  and common regularization functionals such as pairwise 
or high-order MRF potentials \cite{Geman:84,Li:MRFBook}. 
The general form of our joint energy is
\begin{equation} \label{eq:EA+MRF}
E(S) \;=\;\; E_A(S)  \;\;+\;\;\rp\;\; \sum_{c\in\F}  E_{c} (S_{c})
\end{equation} 
where  the first term is some pairwise clustering objective based on data {\em affinity matrix} 
or {\em kernel} $A:=[\kp_{pq}]$ with $\kp_{pq} :=\kp(I_p,I_q)$ defined by some similarity 
function $\kp(\cdot,\cdot)$.  
The second term in \eqref{eq:EA+MRF} is a general formulation of MRF {\em potentials} 
\cite{BVZ:PAMI01,kohli2009robust,LabelCosts:IJCV12}. 
Constant $\rp$ is a relative weight of this term. Subset $c\subset\Omega$ represents a {\em factor}
typically consisting of pixels with nearby locations. Factor labels $S_c:=(S_p|p\in c)$ is a restriction of labeling $S$ to $c$. 
Potentials $E_c(S_c)$ for a given set of factors $\F$ represent various forms 
of unary, second, or higher order constraints, where factor size $|c|$ defines the order. 
Factor features $\{I_p|p\in c\}$ often work as parameters for potential $E_c$.
Section \ref{sec:MRF_overview} reviews standard MRF potentials.
Note that standard clustering methods encourage balanced segments using ratio-based objectives. 
They correspond to high-order potential $E_{\cal A}(S)$ of order $|\Omega|$ in \eqref{eq:EA+MRF}.  
Sections \ref{sec:Kmeans_overview} and \ref{sec:NC_overview} 
review standard clustering objectives used in our paper.

\subsection{Overview of MRF regularization} \label{sec:MRF_overview}

Probably the most basic MRF regularization potential corresponds to the second-order Potts model \cite{BVZ:PAMI01}
used for {\bf edge alignment} 
\begin{equation} \label{eq:potts}
\sum_{c\in\F}  E_{c} (S_{c}) \;=\; \sum_{pq\in\N}  w_{pq}\cdot[S_p\neq S_q] \;\; \approx \;\; ||\partial S||
\end{equation}
where a set of pairwise factors $\F=\N$ includes {\em edges} $c=\{pq\}$ between pairs of neighboring nodes and
$[\cdot]$ are {\em Iverson brackets}. 
Weight $w_{pq}$ is a discontinuity penalty between $p$ and $q$. It could be a constant or may
be set by a decreasing function of intensity difference  $I_p-I_q$ attracting the segmentation boundary 
to image contrast edges \cite{BJ:01}. This is similar to the image-based boundary length 
in geodesic contours \cite{Sapiro:97,BK:ICCV03}.

A useful {\bf bin consistency} constraint enforced by the $P^n$-Potts model \cite{kohli2009robust} is defined over an arbitrary 
collection of high-order factors $\F$.  Factors $c\in\F$ correspond to predefined subsets of nodes 
such as {\em superpixels} \cite{kohli2009robust} or {\em bins} of pixels with the same color/feature \cite{Park:ECCV2012,onecut:iccv13}.
The model penalizes inconsistency in segmentation of each factor 
\begin{equation} \label{eq:pnpotts}
\sum_{c\in\F}  E_{c} (S_{c}) \;\;=\;\; \sum_{c\in\F}  \min\{T,|c|-|S_c|^*\}
\end{equation}
where $T$ is some threshold and $|S_c|^*:=\max_k|S^k\cap c|$ is the cardinality of the largest segment inside $c$.
Potential \eqref{eq:pnpotts} has its lowest value (zero) when all nodes in each factor are within the same segment. 

Standard {\bf label cost} \cite{LabelCosts:IJCV12}  is a sparsity potential defined for a single high-order factor $c=\Omega$.
In its simplest form this potential penalizes the number of distinct segments (labels) in $S$
\begin{equation} \label{eq:labelcost}
E_{\Omega} (S) \;\;=\;\; \sum_{k}  h_k \cdot [ |S^k|>0 ]
\end{equation}
where $h_k$ could be a constant or a cost for each specific label.

Potentials \eqref{eq:potts}, \eqref{eq:pnpotts}, \eqref{eq:labelcost} are only a few examples of 
regularization terms widely used in combination with powerful discrete solvers like
graph cut  \cite{BVZ:PAMI01}, belief propagation \cite{Weiss:05}, 
TRWS \cite{trws}, LP relaxation \cite{werner2007linear,kappes2015comparative}, or
continuous methods \cite{chambolle2004algorithm,chambolle2011first,cremers2007review}.

Image segmentation methods often combine regularization with a {\bf likelihood term} 
integrating segments/objects color models. For example, \cite{BJ:01,BF:IJCV06} used graph cuts to
combine second-order edge alignment \eqref{eq:potts} with a unary (first-order)  appearance term
\begin{equation} \label{eq:E_PS}
  - \sum_k\sum_{p\in S^k} \log P^k(I_p)
\end{equation}
where $\{P^k\}$ are given probability distributions. Unary terms like \eqref{eq:E_PS} are easy 
to integrate into any of the solvers above.

If unknown, parameters of the models $\{P^k\}$ in a regularization energy including \eqref{eq:E_PS} 
are often estimated by iteratively minimizing the energy with respect to $S$ and model parameters 
\cite{Zhu:96,Chan-Vese-01b,ismail:pami05,GrabCuts:SIGGRAPH04,LabelCosts:IJCV12}. 
In presence of variable model parameters, \eqref{eq:E_PS} can be seen as
a {\em maximum likelihood} (ML) model-fitting term or a {\em probabilistic K-means} clustering objective \cite{UAI:97}. 
The next section reviews K-means and other standard clustering methods.

\subsection{Overview of K-means and clustering} \label{sec:Kmeans_overview}

Many clustering methods are based on K-means (\KM/). The most basic iterative \KM/ algorithm \cite{Duda:01} 
can be described as the {\em block-coordinate descent} for the following {\em mixed} objective
\begin{equation} \label{eq:mixed}
F(S,m) \;\;:=\;\; \sum_k \sum_{p \in S^k} \|I_p- m_k \|^2  
\end{equation}
combining discrete variables $S=\{S^k\}_{k=1}^K$ with continuous variables $m=\{m_k\}_{k=1}^K$ 
representing cluster ``centers''. Norm $\| . \| $ denotes the Euclidean metric. 
For any given $S$ the optimal centers $\argmin_{m} F(S,m)$ are the means
\begin{equation} \label{eq:mean}
\muSk \;\; := \;\; \frac{\sum_{q \in S^k} I_q}{|S^k| }
\end{equation}
where $|S^k|$ denotes the segment's cardinality. Assuming current segments $S^k_t$ 
the update operation giving $\argmin_S F(S,\mu_{S_t})$
\begin{equation} \label{eq:KMproc}
\left(\parbox{10ex}{\centering\bf\small basic KM \\ procedure}\right)\quad\quad
S_p \;\;\;\leftarrow\;\;\; \arg\min_k \|  I_p - \mu_{S^k_t}  \| \quad\quad\quad
\end{equation}
defines the next solution $S_{t+1}$ as per standard K-means algorithm. 
This greedy descent technique converges only to a local minimum of \KM/ objective \eqref{eq:mixed},
which is known to be NP hard to optimize. There are also other approximation methods.
Below we review the properties of \KM/ objective \eqref{eq:mixed} independently of optimization.

The optimal centers $m_k$ in \eqref{eq:mean} allow to represent \eqref{eq:mixed} 
via an equivalent objective of a single argument $S$
\begin{equation} \label{eq:kmeans1}
\sum_k \sum_{p\in S^k} \|I_p-\muSk\|^2 \;\;\; \equiv \;\;\; \sum_k |S^k|\cdot var(S^k).
\end{equation}
The sum of squared distances between data points $\{I_p|p\in S^k\}$  and mean $\muSk$
normalized by $|S^k|$ gives the {\em sample variance} denoted by  $var(S^k)$. 
Formulation \eqref{eq:kmeans1} presents the basic \KM/ objective as a standard {\em variance criterion} for clustering. That is, K-means attempts to find K compact clusters with small variance.
 
K-means can also be presented as a pairwise clustering criteria with Euclidean affinities.
The {\em sample variance} can be expressed as the sum of distances between all pairs of the points.
For example, plugging \eqref{eq:mean} into \eqref{eq:kmeans1} reduces this \KM/ objective to
\begin{equation} \label{eq:kmeans2}
\sum_k \frac{\sum_{pq \in S^k}\|I_p-I_q \|^2}{2\;|S^k|}. 
\end{equation}
Taking the square in the denominator transforms \eqref{eq:kmeans2} to another equivalent
\KM/ energy with Euclidean dot-product affinities
\begin{equation} \label{eq:kmeans3}
\utc \;\; - \sum_k \frac{\sum_{pq \in S^k}\langle I_p,I_q \rangle}{\;|S^k|}. 
\end{equation}
Note that we use \utc and \approxutc for ``up to additive constant'' relations.

Alternatively, K-means clustering can be seen as Gaussian model fitting. 
Formula \eqref{eq:E_PS} for normal distributions 
with variable means $m_k$ and some fixed variance
\begin{equation} \label{eq:Gmodels}
  - \sum_k\sum_{p\in S^k} \log N(I_p|m_k)
\end{equation}
equals objective \eqref{eq:mixed} up to a constant.

Various extensions of objectives \eqref{eq:mixed}, \eqref{eq:kmeans1}, \eqref{eq:kmeans2}, \eqref{eq:kmeans3}, or \eqref{eq:Gmodels} 
lead to many powerful clustering methods such as kernel K-means,
average association, and Normalized Cut, see Tab.\ref{tab:kmeans} and Fig.\ref{fig:dclustering}.

\subsubsection{Probabilistic K-means (\pKM/) and model fitting} \label{sec:pKM}

One way to generalize K-means is to replace squared Euclidean distance in \eqref{eq:mixed} 
by other {\em distortion} measures $\|\|_d$ leading to a general {\em distortion energy} commonly used for clustering
\begin{equation} \label{eq:distortion}
\sum_k  \sum_{p \in S^k}\|I_p - m_k \|_d.
\end{equation}
The optimal value of parameter $m_k$ may no longer correspond to a {\em mean}. For example,
the optimal $m_k$ for non-squared $L_2$ metric is a {\em geometric median}. For exponential distortions the
optimal $m_k$ may correspond to {\em modes} \cite{Ismail:tip10,carreira:arXiv13}, see Appendix B.

\begin{figure}
        \centering
        \begin{subfigure}[b]{0.18\textwidth}
                \setlength{\fboxsep}{1pt}\fbox{\includegraphics[width=\textwidth]{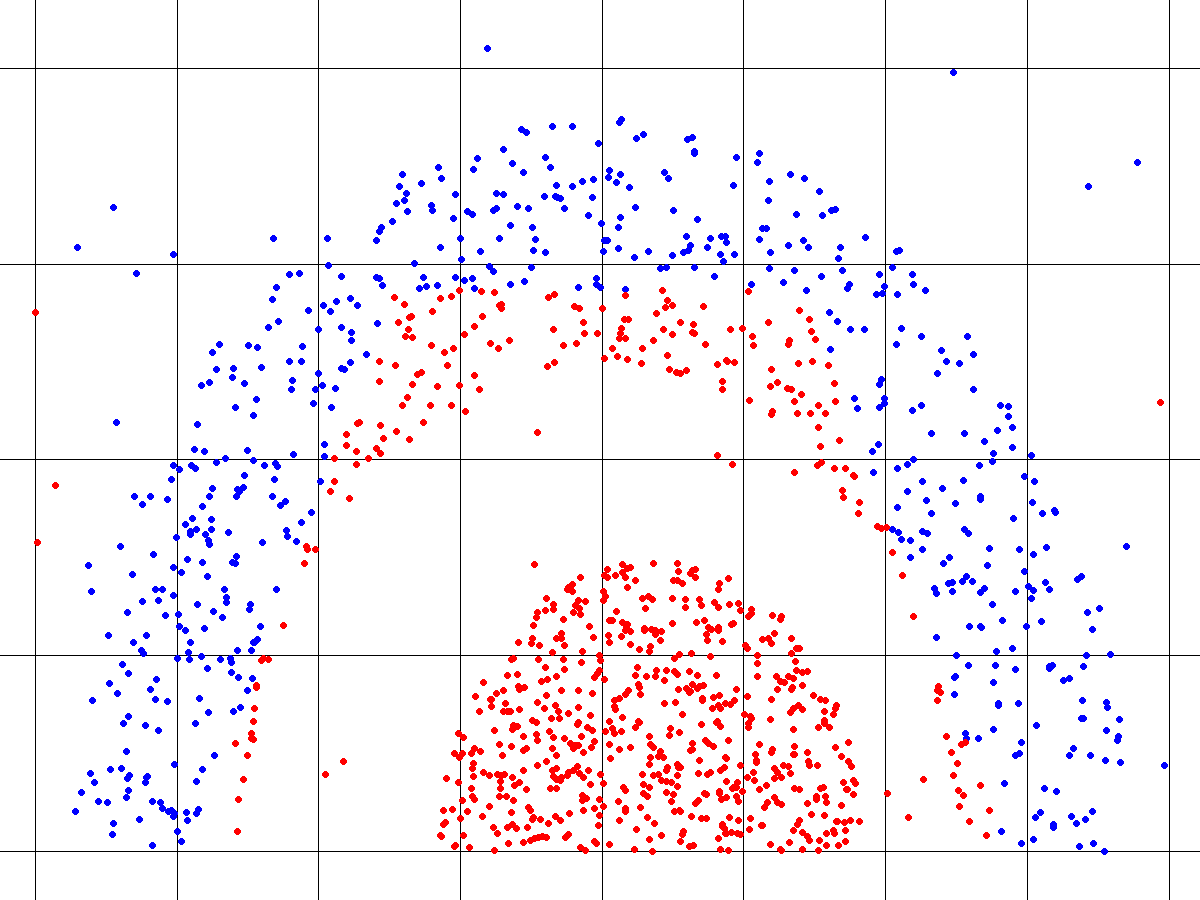}}
                \caption{initialization}
        \end{subfigure}%
        ~ 
        \begin{subfigure}[b]{0.18\textwidth}
                \setlength{\fboxsep}{1pt}\fbox{\includegraphics[width=\textwidth]{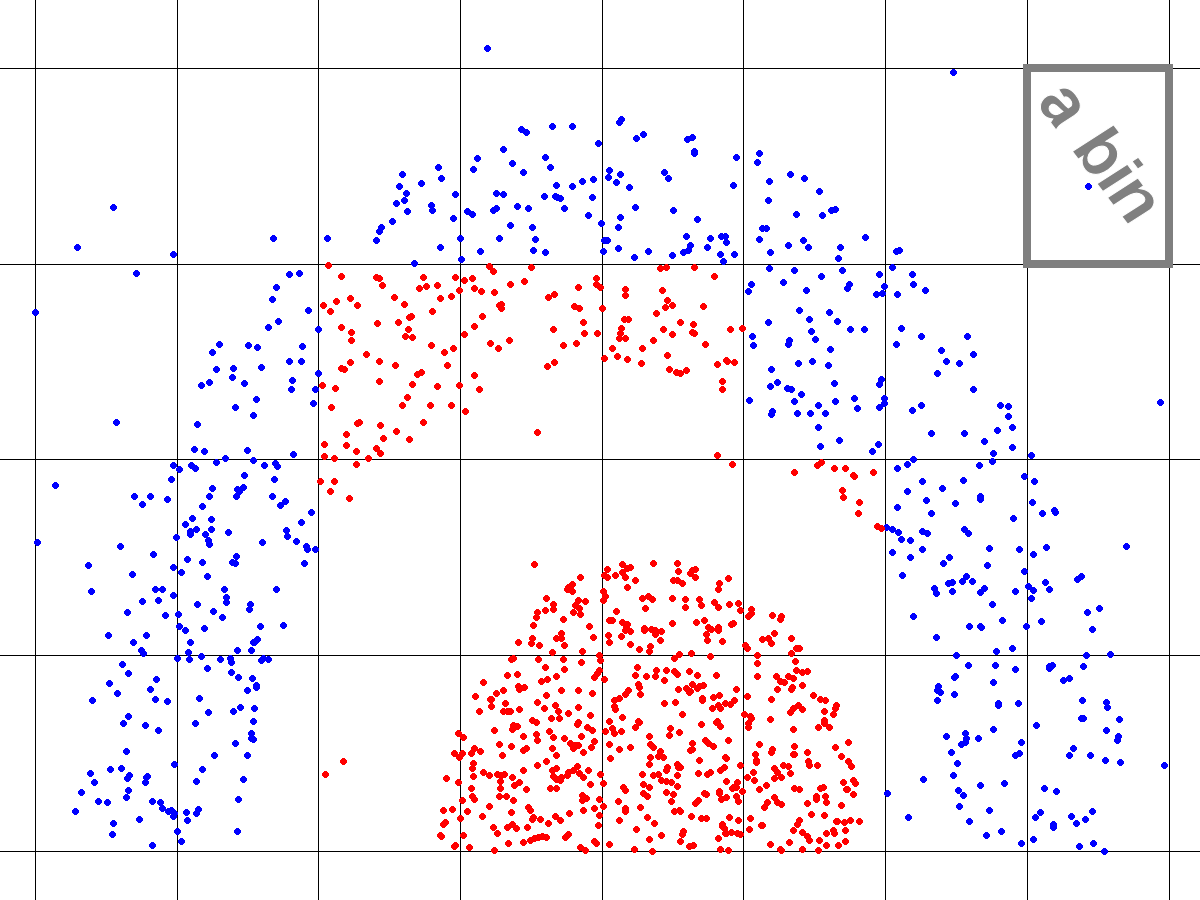}}
                \caption{histogram fitting}
        \end{subfigure}
         \\[1ex]
         \begin{subfigure}[b]{0.18\textwidth}
                \setlength{\fboxsep}{1pt}\fbox{\includegraphics[width=\textwidth]{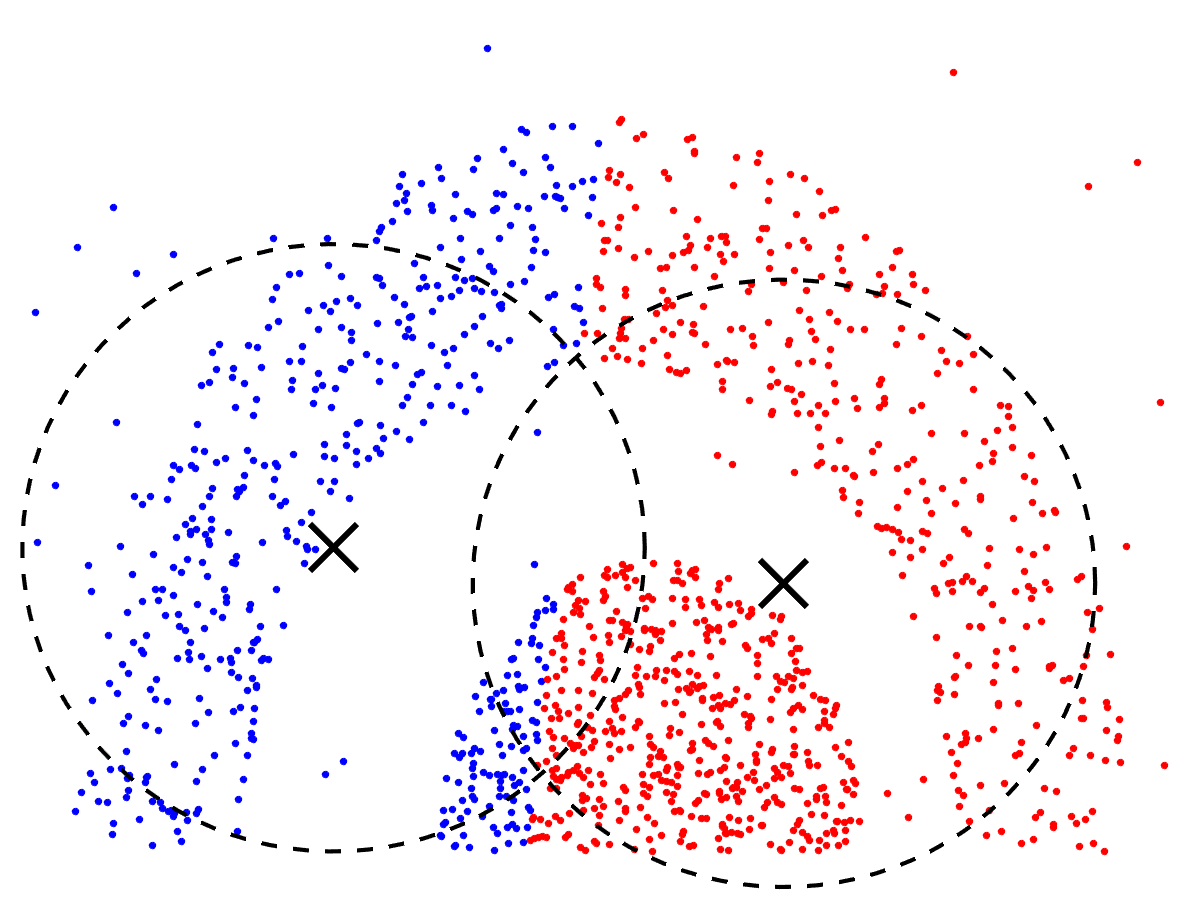}}
                \caption{basic K-means}
        \end{subfigure}%
       ~ 
        \begin{subfigure}[b]{0.18\textwidth}
                \setlength{\fboxsep}{1pt}\fbox{\includegraphics[width=\textwidth]{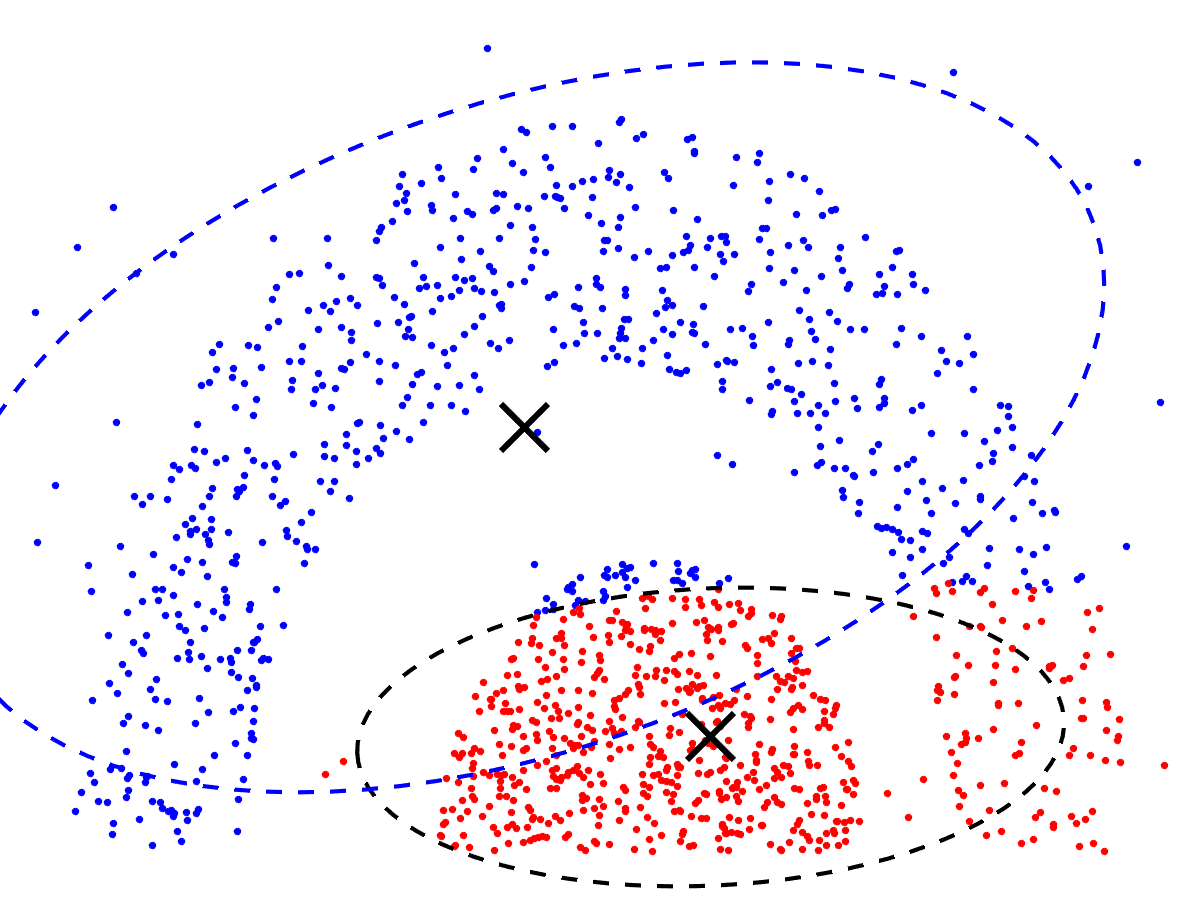}}
                \caption{elliptic K-means}
        \end{subfigure}%
         \\[1ex]
         \begin{subfigure}[b]{0.18\textwidth}
                \setlength{\fboxsep}{1pt}\fbox{\includegraphics[width=\textwidth]{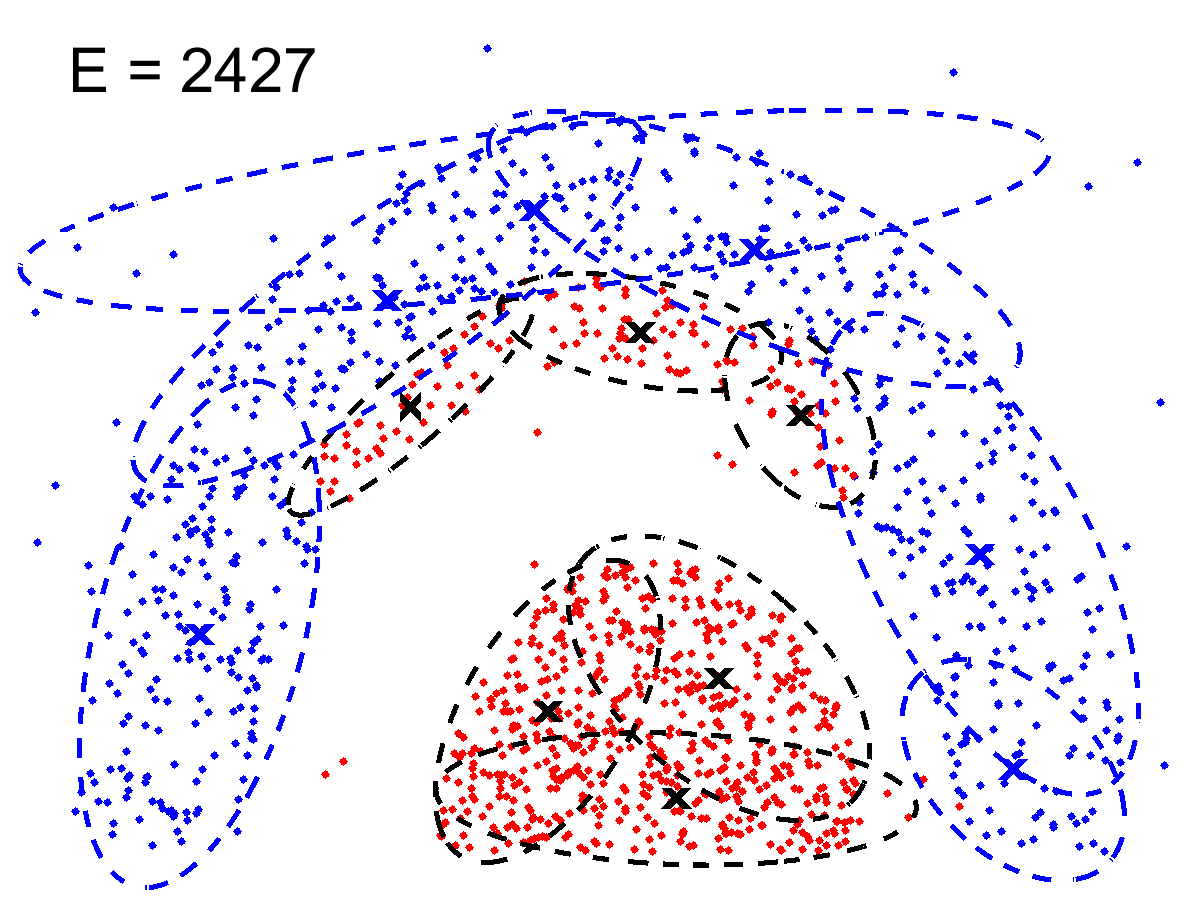}}
                \caption{GMM: local min}
        \end{subfigure}%
       ~ 
        \begin{subfigure}[b]{0.18\textwidth}
                \setlength{\fboxsep}{1pt}\fbox{\includegraphics[width=\textwidth]{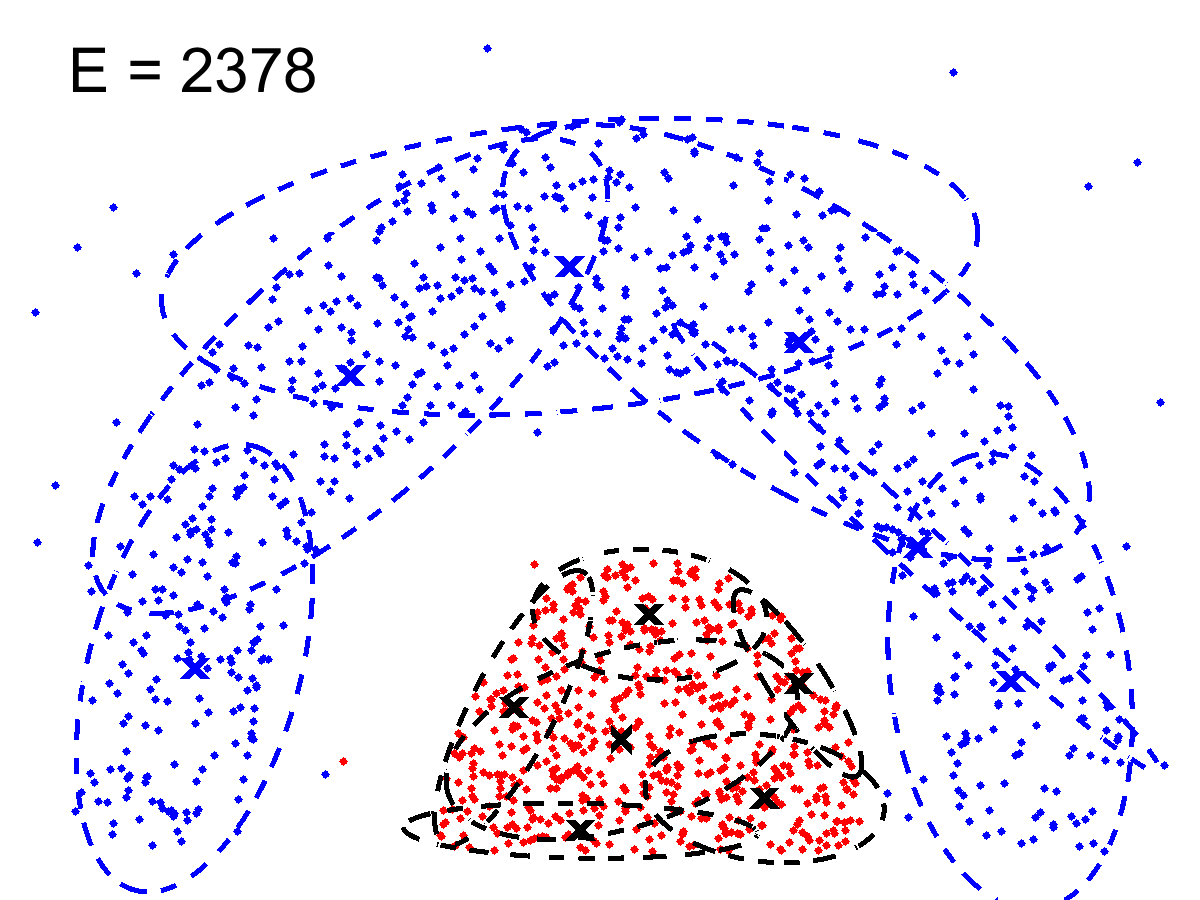}}
                \caption{GMM: from gr. truth}
        \end{subfigure}%
         \\[1ex]
         \begin{subfigure}[b]{0.18\textwidth}
                \setlength{\fboxsep}{1pt}\fbox{\includegraphics[width=\textwidth]{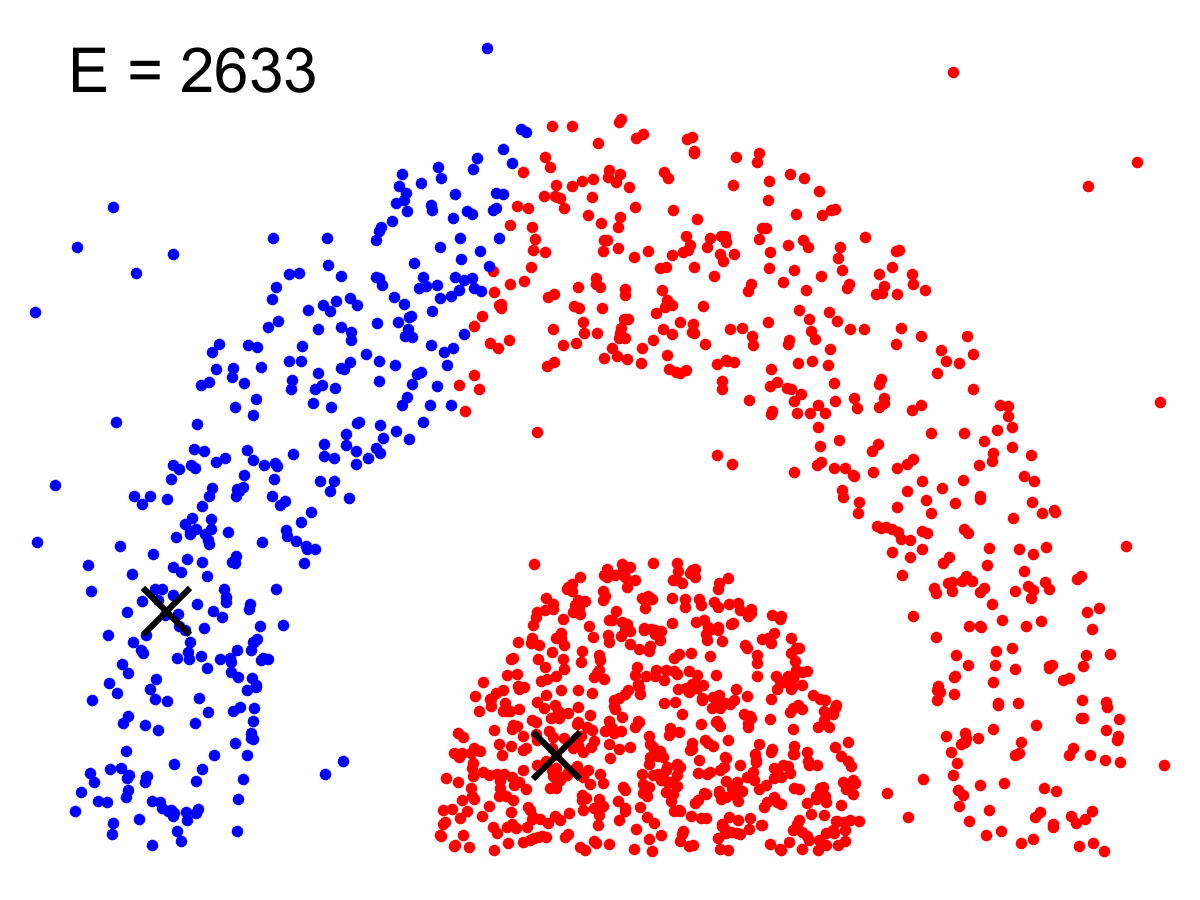}}
                \caption{K-modes $\sim$ {\em mean-shift}}
        \end{subfigure}%
       ~ 
        \begin{subfigure}[b]{0.18\textwidth}
                \setlength{\fboxsep}{1pt}\fbox{\includegraphics[width=\textwidth]{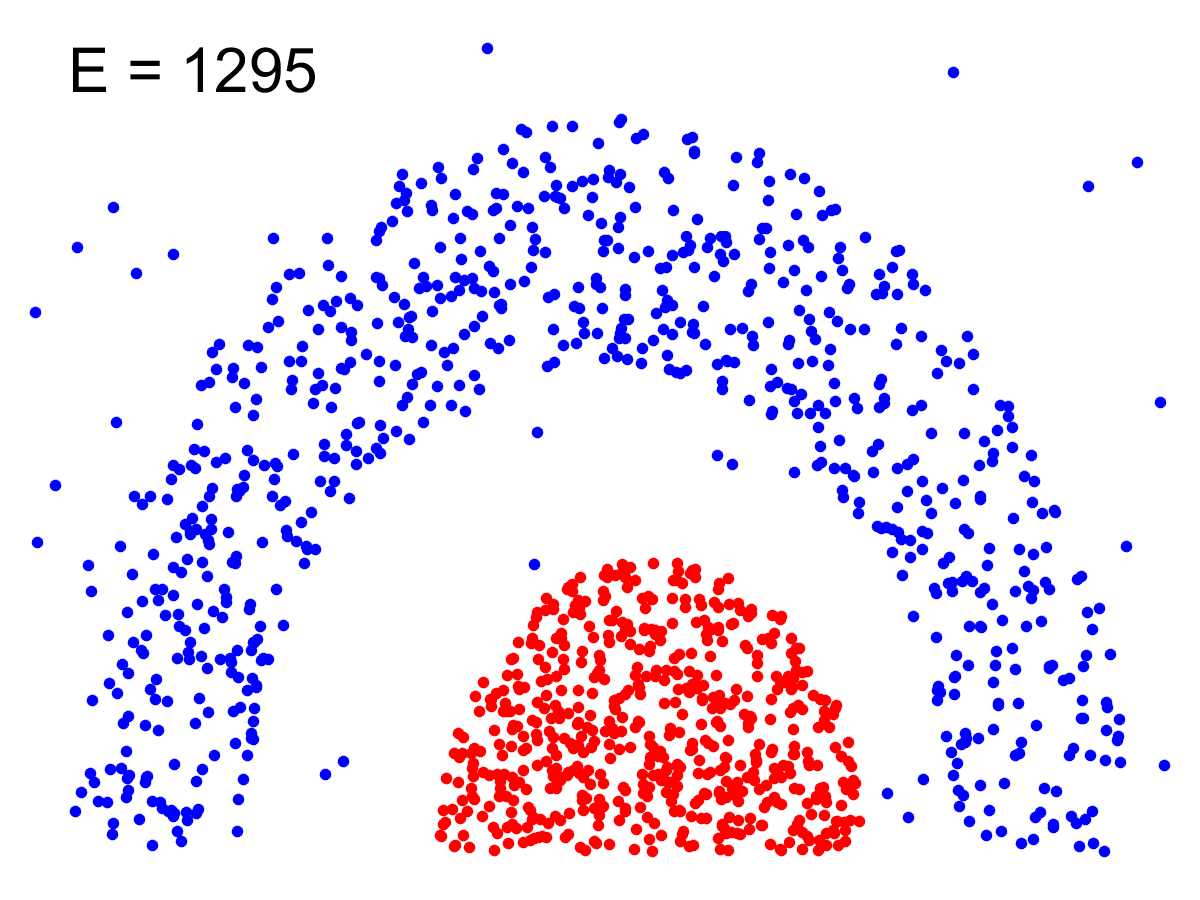}}
                \caption{kernel K-means}
        \end{subfigure}%
        \caption{{\em Model fitting (\pKM/) \eqref{eq:pKmeans} vs kernel K-means (\kKM/) \eqref{eq:kKmeans3}}. 
Histogram fitting converges in one step assigning initially dominant  bin label (a) to all points in the bin (b):
energy (\ref{eq:pKmeans},\ref{eq:entropy}) is minimal at any volume-balanced solution
with one label inside each bin \cite{UAI:97}. Basic and elliptic K-means (one mode GMM) under-fit the data (c,d). 
Six mode GMMs over-fit (e) as in  (b). GMMs have local minima issues;
ground-truth initialization (f) yields lower energy (\ref{eq:pKmeans},\ref{eq:entropy}).
Kernel K-means (\ref{eq:kKmeans2},\ref{eq:kKmeans3}) with Gaussian kernel $k$ in (h) 
outperforms \pKM/ with distortion $\|\|_k$ in (g) related to K-modes or mean-shift ({\em weak} \kKM/, see Sec.\ref{sec:pKM-vs-kKM}).}
        \label{fig:synexp}
\end{figure}


\begin{figure}
        \centering
        \begin{subfigure}[b]{0.5\textwidth}
        \centering
                \includegraphics[width=0.355\textwidth]{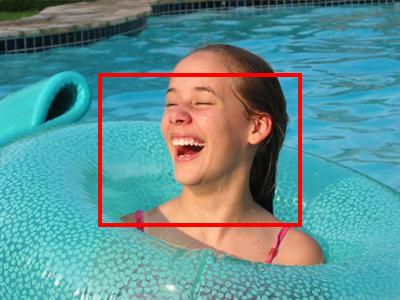}~\includegraphics[width=0.4\textwidth]{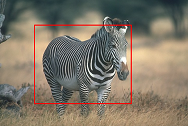}
                \caption{Input and initialization}
        \end{subfigure} \\[1ex]
        \begin{subfigure}[b]{0.5\textwidth}
        \centering
                \setlength{\fboxsep}{1pt}\fbox{\includegraphics[width=0.355\textwidth]{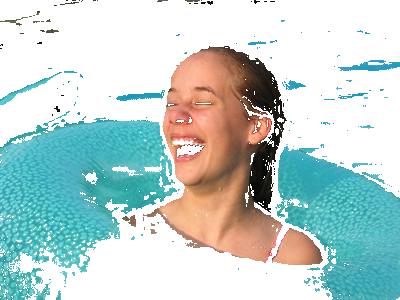}}~\setlength{\fboxsep}{1pt}\fbox{\includegraphics[width=0.4\textwidth]{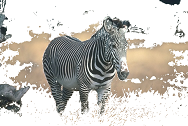}}
                \caption{GMM fitting in RGB (GrabCut without edges)}
        \end{subfigure} \\[1ex]
         \begin{subfigure}[b]{0.5\textwidth}
         \centering
                \setlength{\fboxsep}{1pt}\fbox{\includegraphics[width=0.355\textwidth]{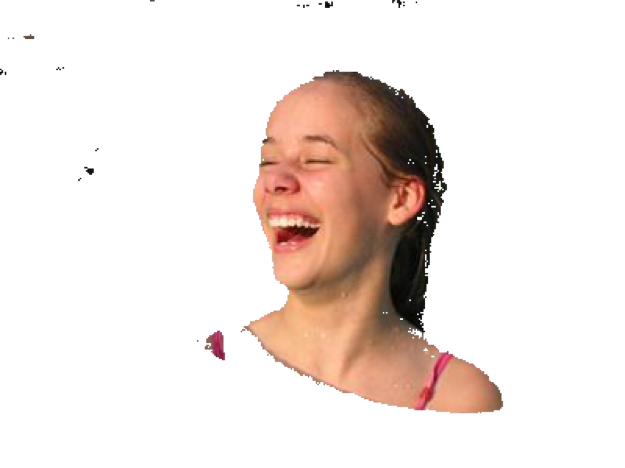}}~\setlength{\fboxsep}{1pt}\fbox{\includegraphics[width=0.4\textwidth]{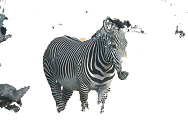}}
                \caption{Normalized Cut in RGB}
        \end{subfigure}
               \caption{Without edge alignment \eqref{eq:potts} iterative GMM fitting \cite{GrabCuts:SIGGRAPH04} shows stronger data over-fit compared to pairwise clustering \cite{Shi2000}.}
        \label{fig:firstexamples}
\end{figure}

A seemingly different way to generalize K-means
is to treat both means and covariance matrices for the normal distributions in \eqref{eq:Gmodels} as variables. 
This corresponds to the standard {\em elliptic K-means} \cite{Sung1995,Rousson2002,LabelCosts:IJCV12}.
In this case variable model parameters $\theta_k=\{m_k,\Sigma_k\}$ and data points $I_p$ are not in 
the same space. Yet, it is still possible to present elliptic K-means as distortion clustering \eqref{eq:distortion} 
with ``distortion'' between $I_p$ and $\theta_k$ defined by an operator $\|\cdot-\cdot\|_d$ corresponding to a likelihood function
$$\|I_p-\theta_k\|_d := -\log N (I_p|\theta_k).$$
Similar distortion measures can be defined for arbitrary probability distributions with any variable 
parameters $\theta_k$. Then, distortion clustering \eqref{eq:distortion} generalizes to ML model fitting objective 
\begin{equation} \label{eq:pKmeans}
\sum_k \sum_{p \in S^k} \|I_p-\theta_k\|_d  \;\;\equiv\;\; -\sum_k \sum_{p \in S^k} \log P (I_p|\theta_k)
\end{equation}
which is \eqref{eq:E_PS} with explicit model parameters $\theta_k$. This formulation
suggests {\em probabilistic K-means}\footnote{The name {\em probabilistic K-means} in 
the general clustering context 
was coined by \cite{UAI:97}. They formulated \eqref{eq:pKmeans} after representing distortion energy \eqref{eq:distortion} 
as ML fitting of Gibbs models $\frac{1}{Z_d}e^{-\|x-m\|_d}$ for arbitrary integrable metrics.}  
(\pKM/) as a good idiomatic name for ML model fitting or distortion clustering \eqref{eq:distortion}, 
even though the corresponding parameters $\theta_k$ are not ``means'', in general. 

Probabilistic K-means  \eqref{eq:pKmeans} is used in image segmentation with models such as 
elliptic Gaussians \cite{Sung1995,Rousson2002,LabelCosts:IJCV12}, gamma/exponential \cite{ismail:pami05}, 
or other generative models \cite{ismail:book10}. 
Zhu-Yuille \cite{Zhu:96} and GrabCut \cite{GrabCuts:SIGGRAPH04} use \pKM/ with 
highly descriptive probability models such as GMM or histograms. 
Information theoretic analysis in \cite{UAI:97} shows that in this case 
\pKM/ objective \eqref{eq:pKmeans} reduces to the standard {\em entropy criterion} for clustering
\begin{equation} \label{eq:entropy}
\sum_k \; |S^k|\cdot H(S^k)
\end{equation}
where $H(S^k)$ is the entropy of the distribution for $\{I_p|p\in S^k\}$.

Intuitively, minimization of the entropy criterion \eqref{eq:entropy} favors clusters with tight or ``peaked'' distributions.
This criterion is widely used in categorical clustering \cite{li:icml04} and decision trees \cite{breiman1996,Louppe2013}
where the entropy evaluates histograms over ``naturally'' discrete features. However, the entropy criterion
with either discrete histograms or continuous GMM densities has limitations in the context of 
{\em continuous} feature spaces, see Appendix C. Iterative fitting of descriptive models 
is highly sensitive to local minima \cite{onecut:iccv13,pbo:ECCV14} and 
easily over-fits even low dimentional features in $\mathcal{R}^2$ (Fig.\ref{fig:synexp}b,e) or  in
$\mathcal{R}^3$ (RGB colors, Fig.\ref{fig:firstexamples}b).
This may explain why this approach to clustering is not too common in the learning community.
As proposed in \eqref{eq:EA+MRF}, instead of entropy criterion we will combine MRF regularization with general pairwise 
clustering objectives $E_A$ widely used for balanced partitioning of 
arbitrary high-dimensional features \cite{Shi2000}.

\begin{savenotes}
\begin{table*}[ht]
\begin{center}
\begin{tabular}{|c||c|}
\hline
\multicolumn{2}{|c|}{A. {\bf basic K-means} (\KM/)
\hspace{3ex} (\eg \cite{Duda:01}) } \\ \hline \multicolumn{2}{|c|}{} \\[-0.5ex]
\multicolumn{2}{|c|}{\;\;\;
$\sum_k\sum_{p \in S^k}\|I_p-\mu_{S^k} \|^2 $ \hspace{9ex} {\bf Variance criterion}} \\[1ex]
\multicolumn{2}{|c|}{$ \;\;\; = \;\;\;\;\;\;\sum_k\frac{\sum_{pq \in S^k}\|I_p-I_q \|^2}{2|S^k|}  
\;\;\;\;\;\;\;\;\;\;\; \quad \sum_k\;|S^k|\cdot var(S^k)  \quad $} \\[1ex]
\multicolumn{2}{|c|}{$\utc \;\; -\sum_k\frac{\sum_{pq \in S^k}\langle I_p,I_q \rangle}{|S^k|} 
\quad\quad\quad\quad\quad\quad\quad\quad\quad\quad\quad\quad\quad\quad\;\;$} \\[1ex]
\multicolumn{2}{|c|}{$\utc \;\; -\sum_k\sum_{p \in S^k} \ln {\cal N} (I_p|\mu_{S^k}) 
\quad\quad\quad\quad\quad\quad\quad\quad\quad\quad\quad\;\;$} \\[1.5ex] \hline
\end{tabular}
\\[1ex] \includegraphics[height=4ex]{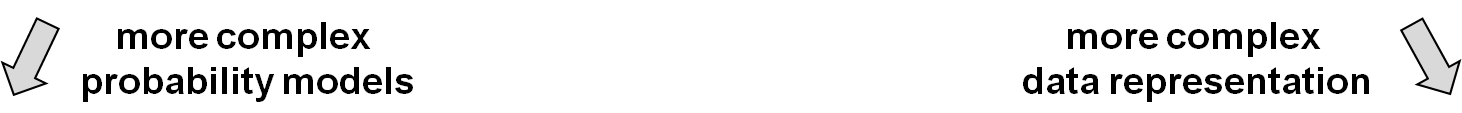} \\[1ex]
\begin{tabular}{|c||c|}
\hline
\hspace{14ex} B. {\bf probabilistic K-means} (\pKM/) \hspace{14ex} & 
\hspace{14ex} C. {\bf kernel K-means}  (\kKM/)  \hspace{14ex} \\ \hline
{\small {\em equivalent energy formulations:}}  & {\small {\em equivalent energy formulations:}} \\[2ex] 
\multirow{2}{*}{ $\quad \sum_k\sum_{p \in S^k}\|I_p - \theta_k \|_d \;\;\;\;=\;\; -\sum_k\sum_{p \in S^k} \ln \P (I_p|\theta_k) \quad$ }  
& $\quad \sum_k\sum_{p \in S^k}\|\phi (I_p)-\mu_{S^k} \|^2\;\;\;\;= \;\;\;\; \sum_k\frac{\sum_{pq \in S^k}||I_p-I_q||_k^2}{2|S^k|}  \quad$ \\[1ex] 
& \hspace{29ex} $\utc \; - \sum_k\frac{\sum_{pq \in S^k}k(I_p,I_q)}{|S^k|} $  \\[2ex] \hline
{\small {\em related examples:}} & 
{\small {\em related examples:}}  \\[1ex] 
elliptic K-means \cite{Sung1995,Rousson2002} & Average Association or Distortion \cite{buhmann:pami03}  \\[1ex]
geometric model fitting \cite{LabelCosts:IJCV12}  & Average Cut \cite{Shi2000} \\[1ex]
K-modes \cite{Ismail:tip10} or mean-shift \cite{dorin:pami02}\hspace{2ex} (weak \kKM/)  
& Normalized Cut  \cite{Shi2000,dhillon2004kernel} \hspace{2ex} (weighted \kKM/)\\[1ex]
Entropy criterion  $\;\; \sum_k \; |S^k|\cdot H(S^k)$ \hspace{1ex} \cite{Zhu:96,GrabCuts:SIGGRAPH04} &
Gini criterion  $\;\; \sum_k \; |S^k|\cdot G(S^k)$ \cite{breiman1996,kkm:iccv15} \\[-0.5ex]
{\tiny for highly descriptive models (GMMs, histograms)}
& {\tiny for small-width normalized kernels (see Sec.\ref{sec:extreme})} \\[0.5ex]
\hline
\end{tabular}
\end{center}\caption{{\em K-means and related clustering criteria:} Basic K-means (A) minimizes clusters variances. It works as 
Gaussian model fitting. Fitting more complex models like elliptic Gaussians \cite{Sung1995,Rousson2002,LabelCosts:IJCV12}, 
exponential distributions \cite{ismail:pami05}, GMM or histograms \cite{Zhu:96,GrabCuts:SIGGRAPH04} 
corresponds to {\em probabilistic K-means} \cite{UAI:97} in (B). Pairwise clustering via {\em kernel K-means} (C) 
using more complex data representation. \label{tab:kmeans}}
\end{table*}
\end{savenotes}

\subsubsection{Kernel K-means (\kKM/) and pairwise clustering} \label{sec:kKM}

This section reviews pairwise extensions of K-means \eqref{eq:kmeans3} such 
as {\em kernel K-means} (\kKM/) and related clustering criteria.
In machine learning, \kKM/ is a well established data clustering technique 
\cite{vapnik:98,Ksurvey:TNN01,Girolami2002,dhillon2004kernel,Chitta2011,Jayasumana2015}
that can identify non-linearly separable structures. 
In contrast to \pKM/ based on complex models, \kKM/ corresponds to complex (nonlinear) mappings 
$$\phi: \mathcal{R}^N\rightarrow\mathcal{H}$$
embedding data $\{I_p|p\in\Omega\}\subset \mathcal{R}^N$ 
as points $\phi_p \equiv \phi(I_p)$ in a higher-dimensional Hilbert space $\mathcal{H}$.
The original non-linear problem can often be solved by simple linear separators of the embedded 
points $\{\phi_p|p\in\Omega\}\subset\mathcal{H}$. Kernel K-means corresponds to 
the basic K-means \eqref{eq:mixed} in the embedding space
\begin{equation} \label{eq:kmixed}
F(S,m) \;\;=\;\; \sum_k \sum_{p \in S^k} \|\phi_p - m_k \|^2 . 
\end{equation}
Optimal segment centers $m_k$ corresponding to the means
\begin{equation} \label{eq:mu_phi}
\mu_{S^k} = \frac{\sum_{q \in S^k} \phi_q}{|S^k|}. 
\end{equation}
reduce \eqref{eq:kmixed}  to \kKM/ energy of the single variable $S$ similar to \eqref{eq:kmeans1}
\begin{equation}
\label{eq:kKmeans1}
F(S) \;\;=\;\; \sum_k \sum_{p \in S^k}\|\phi_p-\mu_{S^k} \|^2.
\end{equation} 

Similarly to \eqref{eq:kmeans2} and \eqref{eq:kmeans3} one can write
pairwise clustering criteria equivalent to \eqref{eq:kKmeans1} based on Euclidean distances 
$\|\phi(I_p) - \phi(I_q) \|$ or inner products $\langle \phi(I_p), \phi(I_q) \rangle$, which are
commonly represented via {\em kernel function} $k(x,y)$
\begin{equation} \label{eq:dotproduct} 
k(x,y) := \langle\phi(x),\phi(y)\rangle.
\end{equation}
The (non-linear) kernel function $k(x,y)$ corresponds to the inner product in $\mathcal{H}$. 
It also defines {\em Hilbertian metric}\footnote{Such metrics can be isometrically 
embedded into a Hilbert space \cite{Hein:PR04}.} 
\begin{eqnarray} 
 \|x-y\|^2_k  & := &   \|\phi(x)-\phi(y)\|^2   \nonumber  \\  
                             &    \equiv      & k(x,x)+k(y,y)-2k(x,y)    \label{eq:kernel_distance} 
\end{eqnarray}
isometric to the Euclidean metric in the embedding space.
Then, pairwise formulations \eqref{eq:kmeans2} and \eqref{eq:kmeans3} for K-means in 
the embedding space \eqref{eq:kKmeans1} can be written with respect to the original data points 
using isometric kernel distance $\|\|^2_k$ in  \eqref{eq:kernel_distance}
\begin{equation} \label{eq:kKmeans2}
F(S)\;\;\equiv\;\;\sum_k \frac{\sum_{pq \in S^k} \|I_p-I_q\|_k^2}{2|S^k|}
\end{equation}
or using kernel function $k$ in \eqref{eq:dotproduct} 
\begin{equation} \label{eq:kKmeans3}
F(S) \;\;\utc\;\;  - \sum_k \frac{\sum_{pq \in S^k}k(I_p,I_q)}{|S^k|}. 
\end{equation}

The definition of kernel $k$ in \eqref{eq:dotproduct} requires embedding $\phi$.
Since pairwise objectives \eqref{eq:kKmeans2} and \eqref{eq:kKmeans3} are defined 
for any kernel function in the original data space, it is possible to formulate \kKM/
by directly specifying an affinity function or kernel $k(x,y)$ rather than embedding $\phi(x)$.
This is typical for \kKM/ explaining why the method is called {\em kernel K-means} 
rather than {\em embedding K-means}\footnote{This could be a name for some clustering techniques constructing explicit embeddings
\cite{belkin2003laplacian,yu2015piecewise} instead of working with pairwise affinities/kernels.}\@.

Given embedding $\phi$,  kernel function $k$ defined by \eqref{eq:dotproduct} is {\em positive semi-definite} (p.s.d), 
that is $k(x,y)\geq 0$ for any $x,y$. Moreover, {\em Mercer's theorem} \cite{Muller2001} states that p.s.d. condition 
for any given kernel $k(x,y)$ is sufficient to guarantee that $k(x,y)$ is an inner product in some Hilbert space. That is,
it guarantees existence of some embedding $\phi(x)$ such that \eqref{eq:dotproduct} is satisfied. 
Therefore, \kKM/ objectives \eqref{eq:kKmeans1}, \eqref{eq:kKmeans2}, \eqref{eq:kKmeans3}  are equivalently defined 
either by embeddings $\phi$ or p.s.d. kernels $k$. Thus, kernels are commonly assumed p.s.d.
However, as discussed later, pairwise clustering objective \eqref{eq:kKmeans3} is also used with non p.s.d. affinities.

To optimize \kKM/ objectives \eqref{eq:kKmeans1}, \eqref{eq:kKmeans2}, \eqref{eq:kKmeans3} 
one can use the basic \KM/ procedure \eqref{eq:KMproc} iteratively minimizing mixed objective  \eqref{eq:kmixed} 
explicitly using embedding $\phi$
\begin{equation} \label{eq:exp_kKMproc}
\left(\parbox{12ex}{\centering\bf\small explicit kKM \\ procedure}\right)\quad\quad
S_p \;\;\;\leftarrow\;\;\; \arg\min_k \|  \phi_p - \mu_{S^k_t}  \| \quad\quad
\end{equation}
where $\mu_{S^k_t}$ is the mean \eqref{eq:mu_phi} for current segment $S^k_t$.
Equivalently, this procedure can use kernel $k$ instead of $\phi$. Indeed, as in Section~8.2.2 of \cite{TaylorCristianini:04},
the square of the objective in \eqref{eq:exp_kKMproc} is
$$ \cancel{\| \phi_p\|^2} - 2{\phi_p}'\mu_{S^k_t} + \| \mu_{S^k_t}\|^2\;\; = \;\;
-2\frac{{\phi_p}'\phi S^k_t}{|S^k_t|} + \frac{{S^k_t}'\phi'\phi S^k_t}{|S^k_t|^2}$$
where we use segment $S^k$ as an indicator vector, embedding $\phi$ as an {\em embedding matrix} $\phi:=[\phi_p]$
where points $\phi_p\equiv\phi(I_p)$ are columns, and $'$ denotes the transpose.
Since the crossed term is a constant at $p$, the right hand side gives an equivalent objective 
for computing $S_p$ in \eqref{eq:exp_kKMproc}. 
Using {\em kernel matrix} ${\cal K} := \phi'\phi$ and 
indicator vector ${\mathbf 1}_p$ for element $p$ we get
\begin{equation} \label{eq:imp_kKMproc}
\left(\parbox{9ex}{\centering\bf\small implicit kKM \\ procedure}\right)\;\;
S_p \;\leftarrow\; \arg\min_k \;
\frac{{S^k_t}'{\cal K}S^k_t}{|S^k_t|^2}  -  2\frac{{{\mathbf 1}_p}'{\cal K}S^k_t}{|S^k_t|}
\end{equation}
where the kernel matrix is directly determined by kernel $k$ 
$${\cal K}_{pq}\equiv \phi_p' \phi_q = \langle\phi_p,\phi_q\rangle = k(I_p,I_q).$$
Approach \eqref{eq:imp_kKMproc} has quadratic complexity ${\cal O}(|\Omega|^2)$ iterations. But,
it avoids explicit high-dimensional embeddings $\phi_p$ in \eqref{eq:exp_kKMproc}
replacing them by kernel $k$ in all computations, {\em a.k.a.} the {\em kernel trick}. 

Note that the implicit \kKM/ procedure \eqref{eq:imp_kKMproc} is guaranteed to decrease pairwise \kKM/ 
objectives \eqref{eq:kKmeans2} or \eqref{eq:kKmeans3}  only for p.s.d. kernels. Indeed, equation \eqref{eq:imp_kKMproc} is derived 
from the standard greedy K-means procedure in the embedding space \eqref{eq:exp_kKMproc} assuming kernel \eqref{eq:dotproduct}. 
The backward reduction of \eqref{eq:imp_kKMproc} to \eqref{eq:exp_kKMproc} can be done only for p.s.d. kernels $k$ 
when Mercer's theorem guarantees existence of some embedding $\phi$ such that  $k(I_p,I_q)=\langle\phi(I_p),\phi(I_q)\rangle$.  

Pairwise energy \eqref{eq:kKmeans2} helps to explain the positive result for \kKM/ with
common Gaussian kernel $k=\exp \frac{-(I_p-I_q)^2}{2\sigma^2}$ in Fig.\ref{fig:synexp}(h). 
Gaussian kernel distance (red plot below) 
\begin{equation} \label{eq:kmetric}
\|I_p-I_q\|_k^2 \propto 1- k(I_p,I_q) = 1 - \exp \frac{-(I_p-I_q)^2}{2\sigma^2}
\end{equation}
\begin{wrapfigure}{r}{30mm}
\includegraphics[width=30mm]{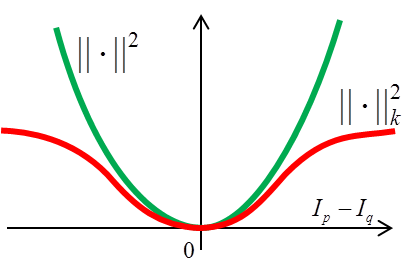}
\end{wrapfigure} 
is a ``robust'' version of Euclidean metric (green) in basic K-means \eqref{eq:kmeans2}.
Thus, Gaussian \kKM/ finds clusters with small local variances, Fig.\ref{fig:synexp}(h).
In contrast, basic K-means (c) tries to find good clusters with small global variances, 
which is impossible for non-compact clusters.

{\bf Average association (\AA/) or distortion (\AD/):} 
Equivalent pairwise objectives \eqref{eq:kKmeans2} and \eqref{eq:kKmeans3} suggest natural extensions of \kKM/. 
For example, one can replace Hilbertian metric $\|\|^2_k$ in  \eqref{eq:kKmeans2} by an arbitrary
zero-diagonal distortion matrix $D=[D_{pq}]$ generating {\em average distortion}  (\AD/) energy 
\begin{equation} \label{eq:AD}
E_{ad}(S) \;:=\; \sum_k\frac{\sum_{pq \in S^k} D_{pq}}{2|S^k|}
\end{equation}
reducing to \kKM/ energy \eqref{eq:kKmeans2} for $D_{pq}=\|I_p-I_q\|^2_k$.
Similarly, p.s.d. kernel $k$ in \eqref{eq:kKmeans3}  can be replaced by an arbitrary 
pairwise similarity or affinity matrix $A=[A_{pq}]$ defining standard {\em average association} (\AA/) energy
\begin{equation} \label{eq:AA}
E_{aa}(S) := -\sum_k \frac{\sum_{pq \in S^k} A_{pq}}{|S^k|} 
\end{equation}
reducing to \kKM/ objective \eqref{eq:kKmeans3} for $A_{pq}=k(I_p,I_q)$. 
We will also use association between any two segments $S^i$ and $S^j$ 
\begin{equation}\label{eq:assoc}
assoc(S^i,S^j) \;\;:= \sum_{p\in S^i, q\in S^j} \kp_{pq} \;\;\equiv \;\; {S^i}' A S^j 
\end{equation} 
allowing to rewrite \AA/ energy \eqref{eq:AA} as
\begin{equation} \label{eq:AA2}
E_{aa}(S) \;\equiv\; -\sum_k \frac{assoc(S^k,S^k)}{|S^k|} \;\equiv\;  -\sum_k \frac{{S^k}' A S^k }{{\bf 1}'S^k} 
\end{equation}
The matrix expressions in \eqref{eq:assoc} and \eqref{eq:AA2} represent segments 
$S^k$ as indicator vectors such that $S^k_p = 1$ iff $S_p=k$ and symbol~$'$ means a transpose.
Matrix notation as in  \eqref{eq:AA2}  will be used for all pairwise clustering objectives discussed in this paper.

\begin{figure*}
\centering
\includegraphics[width=5.0in]{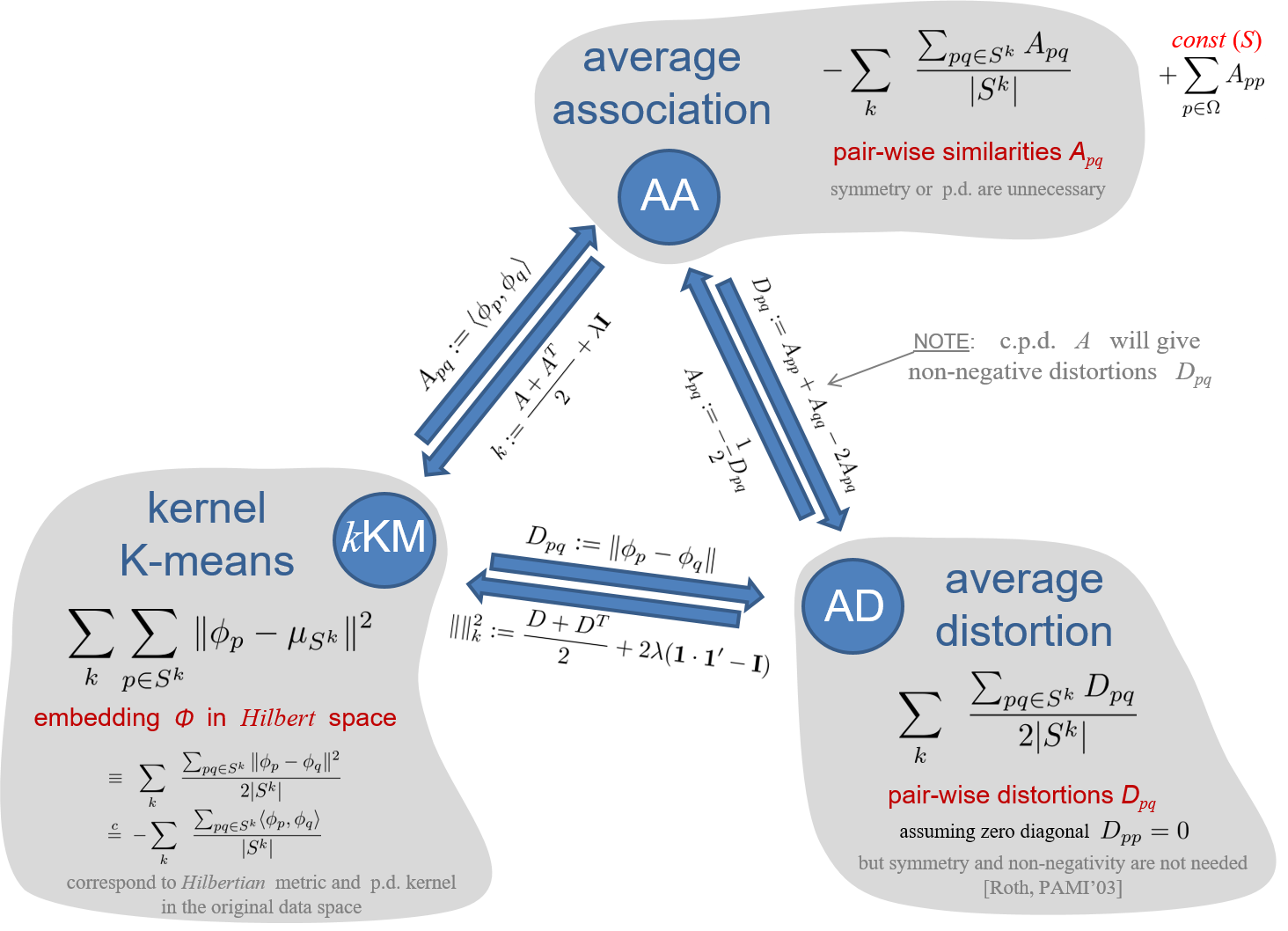} 
\centering
 \caption{Equivalence of pairwise clustering methods: 
{\em kernel K-means} (\kKM/), {\em average distortion} (\AD/), {\em average association} (\AA/) based on
Roth et al. \cite{buhmann:pami03}, see \eqref{eq:k_for_AA}, \eqref{eq:d_for_AD}.
Equivalence of these methods in the general {\em weighted} case is discussed in Appendix A (Fig.~\ref{fig:dclustering2}).
\label{fig:dclustering1}}
\end{figure*}

\kKM/ algorithm \eqref{eq:imp_kKMproc} is not guaranteed to decrease \eqref{eq:AA} for improper (non p.s.d.) kernel matrix 
${\cal K}=A$, but general \AA/ and \AD/ energies could be useful despite optimization issues.
However, \cite{buhmann:pami03} showed that dropping metric and proper kernel assumptions
are not essential; there exist p.s.d. kernels with  \kKM/ energies equivalent (up to constant) to 
\AD/ \eqref{eq:AD} and \AA/ \eqref{eq:AA} for arbitrary associations $A$ and 
zero-diagonal distortions $D$, see Fig.~\ref{fig:dclustering1}.

For example, for any given affinity $A$ in \eqref{eq:AA} the {\em diagonal shift} trick
of Roth et al. \cite{buhmann:pami03} generates the ``kernel matrix'' 
\begin{equation} \label{eq:k_for_AA}
{\cal K} = \frac{A+A'}{2} + \ds \cdot \Identity .
\end{equation}
For sufficiently large scalar $\ds$ matrix ${\cal K}$ is positive definite
yielding a proper discrete kernel $k(I_p,I_q) \equiv  {\cal K}_{pq}$
$$k(I_p,I_q): \chi\times\chi\rightarrow\mathcal{R}$$ for finite set $\chi=\{I_p|p\in\Omega\}$. 
It is easy to check that \kKM/ energy \eqref{eq:kKmeans3} with kernel $k\equiv{\cal K}$ in \eqref{eq:k_for_AA} 
is equivalent to \AA/ energy \eqref{eq:AA} with affinity $A$, up to a constant. Indeed, for any indicator $X\in\{0,1\}^{|\Omega|}$
we have $X'X={\bf 1}'X$ implying
$$ \frac{X'{\cal K} X}{{\bf 1}'X} =  \frac{X'AX}{2({\bf 1}'X)} + \frac{X'A'X}{2({\bf 1}'X)} + \ds \frac{X'X}{{\bf 1}'X} = \frac{X'AX}{{\bf 1}'X} + \ds.$$
Also, Section~\ref{sec:embedding} uses eigen decomposition of ${\cal K}$
to construct an explicit finite-dimensional Euclidean embedding\footnote{Mercer's theorem is
a similar eigen decomposition for continuous p.d. kernels $k(x,y)$ giving infinite-dimensional Hilbert embedding $\phi(x)$. 
Discrete kernel embedding  $\phi_p\equiv\phi(I_p)$ in Sec.~\ref{sec:embedding} \eqref{eq:phi_exact} has finite dimension $|\Omega|$, which is still much higher than the dimension 
of points $I_p$, \eg ${\mathcal R}^3$ for colors. Sec.~\ref{sec:embedding} also shows
lower dimensional embeddings $\tilde{\phi}_p$ approximating isometry  \eqref{eq:kernel_distance}.}  
$\phi_p\in{\mathcal R}^{|\Omega|}$ satisfying isometry \eqref{eq:kernel_distance} 
for any p.d. discrete kernel $k\equiv {\cal K}$. Minimizing \kKM/ energy \eqref{eq:kKmeans1} over such embedding
isometric to ${\cal K}$ in \eqref{eq:k_for_AA}  is equivalent to optimizing \eqref{eq:kKmeans3} and, therefore, \eqref{eq:AA}.

Since average distortion energy \eqref{eq:AD} for arbitrary $D$ is equivalent to average association for 
$A = -\frac{D}{2}$, it can also be converted to \kKM/ with a proper kernel  \cite{buhmann:pami03}. 
Using the corresponding kernel matrix \eqref{eq:k_for_AA} and \eqref{eq:kernel_distance} it is easy 
to derive Hilbertian distortion (metric) equivalent to original distortions $D$ 
\begin{equation} \label{eq:d_for_AD}
\|I_p-I_q\|^2_k  := \frac{D+D'}{2} + 2\ds ({\bf 1}\cdot{\bf 1}'-\Identity).
\end{equation}

For simplicity and without loss of generality, \underline{the rest of the} \underline{paper assumes symmetric affinities $A=A'$} since
non-symmetric ones can be equivalently replaced by $\frac{A+A'}{2}$. 
However, we do not assume positive definiteness and discuss diagonal shifts, if needed.

{\bf Weighted \kKM/ and weighted \AA/:} 
Weighted K-means \cite{Duda:01} is a common extension of \KM/ techniques
incorporating some given point weights $w=\{w_p|p\in\Omega\}$. 
In the context of embedded points $\phi_p$ it corresponds to {\em weighted} \kKM/ iteratively 
minimizing the weighted version of the mixed objective in \eqref{eq:kmixed}
\begin{equation}  \label{eq:wkmixed} 
F^w(S,m)  \; := \; \sum_k \sum_{p \in S^k} w_p\|\phi_p- m_k \|^2  .
\end{equation}
Optimal segment centers $m_k$ are now weighted means
\begin{equation} \label{eq:wmean}
\mu^w_{S^k} \;\; = \;\; \frac{\sum_{q \in S^k} w_q \phi_q}{\sum_{q \in S^k} w_q }\;\;\equiv \;\;  
                              \frac{  \phi\, W S^k}{w'S^k}
\end{equation}
where the matrix formulation has weights represented by column vector $w\in{\mathcal R}^{|\Omega|}$  
and diagonal matrix $W:=diag(w)$. Assuming a finite dimensional data embedding $\phi_p\in{\mathcal R}^{m}$
this formulation uses {\em embedding matrix} $\phi :=[\phi_p]$ with column vectors $\phi_p$.
This notation implies two simple identities used in \eqref{eq:wmean}
\begin{equation} \label{eq:identities}
\;\sum_{q\in S^k} w_q \equiv w' S^k \;\;\;\;\;\;\text{and}  \;\;\;\;\;\; \sum_{q\in S^k} w_q\phi_p\equiv\phi WS^k.
\end{equation}
Inserting weighted means \eqref{eq:wmean} into mixed objective \eqref{eq:wkmixed} produces 
a pairwise energy formulation for weighted \kKM/ similar to \eqref{eq:kKmeans3}
\begin{eqnarray} 
F^w(S)  & :=  &   \sum_k \sum_{p \in S^k} w_p\|\phi_p- \mu^w_{S^k} \|^2        \label{eq:wkKM0}   \\
               & \utc   &  -\sum_k \;\; \frac{\sum_{pq \in S^k} w_p w_q {\cal K}_{pq}}{\sum_{p \in S^k} w_p}  \label{eq:wkKM}   \\
              & \equiv & -\sum_k \;\; \frac{ {S^k}'W \,{\cal K}\, W S^k }{w'S^k}  \nonumber
\end{eqnarray}
where p.s.d kernel matrix ${\cal K} = \phi'\phi $ corresponds to the dot products 
in the embedding space, \ie ${\cal K}_{pq}\;=\; \phi'_p \phi_q$. 

Replacing the p.s.d. kernel with an arbitrary affinity matrix $A$ defines a weighted \AA/ objective generalizing \eqref{eq:AA} and \eqref{eq:AA2}
\begin{eqnarray} \label{eq:wAA}
E^w_{aa}(S) & :=  &  -\sum_k \frac{{S^k}' WAW S^k }{w' S^k}.  
\end{eqnarray}
Weighted \AD/ can also be defined. Equivalence of \kKM/, \AA/, and \AD/ in the general {\em weighted} case 
is discussed in Appendix A.

{\bf Other pairwise clusteing criteria:} Besides \AA/ there are many other standard 
pairwise clustering criteria defined by affinity matrices $A=[A_{pq}]$. For example, Average Cut (\AC/) 
\begin{eqnarray} 
E_{ac}(S)  & := &  \sum_k \frac{assoc(S^k,\bar{S}^k)}{|S^k|}   \;\equiv\;  \sum_k \frac{{S^k}' A ({\bf 1}-S^k)}{{\bf 1}'S^k}   \nonumber \\
                  & = &    \sum_k \frac{{S^k}' (D-A) S^k}{{\bf 1}'S^k}     \label{eq:AC}
\end{eqnarray}
where $D: = diag(d)$ is a {\em degree matrix} defined by node degrees vector $d:=A \bf 1$. 
The formulation on the last line \eqref{eq:AC} comes from the following identity valid for Boolean $X\in\{0,1\}^{|\Omega|}$
$$X'DX = X'd. $$

Normalized Cut (\NC/) \cite{Shi2000}  in \eqref{eq:NC} is another well-known pairwise clustering criterion. 
Both \AC/ and \NC/ can also be reduced to \kKM/ \cite{buhmann:pami03,bach:nips03,dhillon2004kernel}. 
However, {\em spectral relaxation} \cite{Shi2000} is more common optimization method for pairwise clustering objectives than 
iterative \kKM/ procedure \eqref{eq:imp_kKMproc}.
Due to popularity of  \NC/ and spectral relaxation we review them in a dedicated Section \ref{sec:NC_overview}.

\subsubsection{Pairwise vs. pointwise distortions} \label{sec:pKM-vs-kKM}

Equivalence of \kKM/ to pairwise distortion criteria \eqref{eq:AD} 
helps to juxtapose {\em kernel K-means} with {\em probabilistic K-means} (Sec.\ref{sec:pKM})
from one more point of view.
Both methods generalize the basic K-means \eqref{eq:mixed}, \eqref{eq:kmeans2} 
by replacing the Euclidean metric with a more general distortion measure $\|\|_d$. While \pKM/ uses 
``pointwise'' formulation \eqref{eq:distortion} where $\|\|_d$ measures distortion between a point
and a model, \kKM/ uses ``pairwise'' formulation \eqref{eq:kKmeans2} where $\|\|_d=\|\|^2_k$ 
measures distortion between pairs of points. 

These two different formulations are equivalent for Euclidean distortion (\ie basic K-means), but
the pairwise approach is strictly stronger than the pointwise version using
the same Hilbertian distortion $\|\|_d = \|\|^2_k$ in non-Euclidean cases (see Appendix B). 
The corresponding pointwise approach is often called {\em weak kernel K-means}. 
Interestingly, weak \kKM/ with standard Gaussian kernel can be seen as {\em K-modes} \cite{Ismail:tip10}, see Fig.~\ref{fig:synexp}(g). 
Appendix B also details relations between K-modes and popular {\em mean-shift} clustering 
\cite{dorin:pami02}. An extended version of Table \ref{tab:kmeans} including weighted \KM/ and weak \kKM/ techniques 
is given in Figure~\ref{fig:dclustering}.

\subsection{Overview of Normalized Cut and spectral clustering} \label{sec:NC_overview}

Section \ref{sec:kKM} has already discussed \kKM/ and many related {\em pairwise clustering} criteria based on specified
affinities $A=[A_{pq}]$.  This section is focused on a related pairwise clustering method, Normalized Cut (\NC/) \cite{Shi2000}.
Shi and Malik \cite{Shi2000} also popularized  pairwise clustering optimization via {\em spectral relaxation}, which is
different from iterative K-means algorithms \eqref{eq:exp_kKMproc} \eqref{eq:imp_kKMproc}. 
Note that there are many other popular optimization methods for different clustering energies using pairwise affinities
\cite{ratioregions:icpr96,jermyn:pami01,ratiocuts:PAMI03,parmaxflow:iccv07,Hochbaum2010},
which are outside the scope of this work. 

\subsubsection{\NC/ objective and its relation to \AA/ and \kKM/}   \label{sec:NC-to-wKKM}

Normalized Cut (\NC/) energy \cite{Shi2000} can be written as
\begin{equation} \label{eq:NC}
E_{nc}(S)  :=\; -\sum_k \frac{assoc(S^k,S^k)}{assoc(\Omega,S^k)} \;\equiv\; - \sum_k \frac{{S^k}' A S^k}{{\bf 1}'AS^k}
\end{equation} 
where {\em association} \eqref{eq:assoc} is defined by a given {\em affinity matrix} $A$.
The second matrix formulation above shows that the difference between \NC/ and
\AA/  \eqref{eq:AA2} is in the normalization. \AA/ objective uses denominator ${\bf 1}'S^k \equiv  |S^k|$, which is $k$-th segment's size.
\NC/ \eqref{eq:NC} normalizes by weighted size. Indeed, using $d := A' {\bf 1}$
 $$ {\bf 1}'AS^k  \; \equiv \;  d'S^k \; \equiv \; \sum_{p\in S^k} d_p$$ 
where weights $d=\{d_p|p\in\Omega\}$ are {\em node degrees}
\begin{equation} \label{eq:d}
d_p\;\;:=\;\;\sum_{q \in\Omega} A_{pq}.
\end{equation}
For convenience, \NC/ objective can be formatted similarly to \eqref{eq:AA2}
 \begin{equation} \label{eq:NC2}
E_{nc}(S) \;\;\equiv\;\;  - \sum_k \frac{{S^k}' A S^k}{d'S^k}.
\end{equation} 

For some common types of affinities where $d_p\approx const$, \eg $K$-nearest neighbor (\KNN/) graphs, \NC/ and \AA/ objectives \eqref{eq:NC2} and 
\eqref{eq:AA2} are equivalent. More generally, Bach \& Jordan \cite{bach:nips03}, Dhillon et al. \cite{dhillon2004kernel} showed 
that \NC/ objective can be reduced to weighted \AA/ or \kKM/ with specific weights and affinities.

Our matrix notation makes equivalence between \NC/ \eqref{eq:NC2}  and weighted \AA/  \eqref{eq:wAA} straightforward. Indeed,
objective \eqref{eq:NC2} with affinity $A$ coincides with \eqref{eq:wAA} for 
weights $w$ and affinity $\tilde{A}$
\begin{equation} \label{eq:Atilde}
w=d= A'{\mathbf 1} \;\;\;\;\;\;\; \text{and}   \;\;\;\;\;\;\;\; \tilde{A} = W^{-1} A W^{-1}.
\end{equation}  
The weighted version of \kKM/ procedure \eqref{eq:imp_kKMproc} detailed in Appendix A
minimizes weighted \AA/  \eqref{eq:wAA}  only for p.s.d. affinities, but
positive definiteness of $A$ is not critical. For example, an extension of the {\em diagonal shift}  \eqref{eq:k_for_AA} \cite{buhmann:pami03} 
can convert \NC/ \eqref{eq:NC2}  with arbitrary (symmetric) $A$ 
to an equivalent \NC/ objective with p.s.d. affinity
\begin{equation} \label{eq:dshift-NC}
{\cal K} =  A + \ds \cdot D
\end{equation}
using {\em degree matrix} $D:=diag(d)\equiv W$ and sufficiently large $\ds$. 
Indeed, for indicators $X\in\{0,1\}^{|\Omega|}$ we have $X'DX=d'X$ and
$$  \frac{X'{\cal K} X}{d'X} =  \frac{X'AX}{d'X} + \ds \frac{X'DX}{d'X} = \frac{X'AX}{d'X} + \ds.$$
Positive definite ${\cal K}$ \eqref{eq:dshift-NC} implies p.d. affinity \eqref{eq:Atilde} of weighted \AA/
\begin{equation} \label{eq:dshift-NC2} 
\tilde{\cal K}=D^{-1}{\cal K}D^{-1} \;=\; D^{-1}AD^{-1} + \ds D^{-1}. 
\end{equation} 
The weighted version of \kKM/ procedure \eqref{eq:imp_kKMproc} for this p.d. kernel \cite{Dhillon:PAMI07} 
greedily optimizes \NC/ objective \eqref{eq:NC2} for any (symmetric) $A$.

\subsubsection{Spectral relaxation and other optimization methods}

There are many methods for approximate optimization of NP-hard pairwise clustering energies 
besides greedy K-mean procedures. In particular, Shi, Malik, and Yu \cite{Shi2000,Yu:iccv03} 
popularized {\em spectral relaxation} methods in the context of normalized cuts. 
Such methods also apply to \AA/ and other problems \cite{Shi2000}. For example, 
similarly to \cite{Yu:iccv03} one can rewrite \AA/ energy \eqref{eq:AA} as 
$$E_{aa} (S) = -\operatorname{tr}(Z' AZ)  \;\;\;\;\mbox{for}\;\;\; Z:=
\left[\dots,\frac{S^k}{\sqrt{|S^k|}},\dots\right]$$
where $Z$ is a $|\Omega|\times K$ matrix of normalized indicator vectors $S^k$. 
Orthogonality $(S^i)' S^j =0$ implies $Z' Z = I_K$ where $I_K$ is an identity matrix of size $K\times K$. 
Minimization of the {\em trace} energy above with relaxed $Z$ constrained to a ``unit sphere'' $Z' Z = I_K$ 
is a simple representative example of {\em spectral relaxation} in the context of AA.
This relaxed trace optimization is a generalization of {\em Rayleigh quotient} problem that has 
an exact closed form solution in terms of $K$ largest eigenvectors for matrix $A$. 
This approach extends to general multi-label weighted \AA/ and related graph clustering problems, 
\eg \AC/ and \NC/ \cite{Shi2000,Yu:iccv03}.
The main computational difficulties for spectral relaxation methods are explicit eigen decomposition 
for large matrices and integrality gap - there is a final heuristics-based discretization step 
for extracting an integer solution for the original combinatorial problem from an optimal relaxed solution.
For example, one basic discretization heuristic is to run K-means over the row-vectors of the 
optimal relaxed $Z$.

Other optimization techniques are also used for pairwise clustering.
Buhmann et al. \cite{hofmann:pami97,buhmann:pami03} address the general \AD/ and \AA/ 
energies via mean field approximation and {\em deterministic annealing}. 
It can be seen as a {\em fuzzy} version\footnote{{\em Fuzzy} version of K-means in 
Duda et al. \cite{duda:2001} generalizes to \kKM/. } of the kernel K-means algorithm. 
In the context of normalized cuts Dhillon et al.  \cite{dhillon2004kernel,Dhillon:PAMI07} 
use spectral relaxation to initialize \kKM/ algorithm. 

In computer vision it is common to combine various constraints or energy terms into one objective function.
Similar efforts were made in the context of pairwise clustering as well.
For example, to combine \kKM/ or \NC/ objectives with Potts regularization \cite{Kulis:ML09} normalizes 
the corresponding pairwise constraints by cluster sizes. This alters the Potts model to fit the problem 
to a standard trace-based formulation. 

Adding non-homogeneous linear constraints into spectral relaxation techniques also requires 
approximations \cite{stella:pami04} or model modifications \cite{xu:cvpr09}. 
Exact optimization for the relaxed quadratic ratios (including \NC/) with arbitrary linear equality constraints 
is possible by solving a sequence of spectral problems \cite{eriksson:JMIV11}.

Our bound optimization approach allows to combine many standard pairwise clustering objectives with any regularization
terms with existing solvers. In our framework such pairwise clustering objectives are interpreted as
high-order energy terms approximated via linear upper bounds during optimization.

\subsection{Our approach summary}

We propose to combine standard pairwise clustering criteria such as \AA/, \AC/, or \NC/ with standard
regularization constraints such as geometric or MRF priors. To achieve this goal,
we propose unary (linear) bounds for the pairwise clustering objectives that are easy to integrate into
any standard regularization algorithms. Below we summarize our motivation and our main technical contributions.

\subsubsection{Motivation and Related work}

Due to significant differences in existing optimization methods, pairwise clustering (\eg \NC/) and Markov Random Fields (MRF) techniques
are used separately in many applications of vision and learning. They have complementary strengths and weaknesses. 

For example, \NC/ can find a balanced partitioning of data points from pairwise affinities
for high-dimensional features \cite{Shi2000,malik:ijcv01,arbelaez2011contour}. 
In contrast, discrete MRF as well as continuous regularization methods
commonly use {\em probabilistic K-means} \cite{UAI:97} or {\em model fitting} to partition image features 
\cite{Chan-Vese-01b,Zhu:96,GrabCuts:SIGGRAPH04,LabelCosts:IJCV12}.
Clustering data by fitting simple parametric models seems intuitively justified when data supports such simple models, 
\eg Gaussians \cite{Chan-Vese-01b} or lines/planes \cite{LabelCosts:IJCV12}. 
But, clustering arbitrary data by fitting complex models like GMM or histograms \cite{Zhu:96,GrabCuts:SIGGRAPH04} 
seems like an idea bound to over-fit the data. Indeed, over-fitting happens even for low dimensional 
color features \cite{onecut:iccv13}, see also Fig.\ref{fig:synexp}(b,e) and Fig.\ref{fig:firstexamples}(b). 
Our energy \eqref{eq:EA+MRF} allows a general MRF framework to benefit from standard
pairwise clustering criteria, \eg \NC/, widely used for complex data. 
In general, kernel-based clustering methods are a prevalent choice 
in the learning community as model fitting (\eg EM) becomes intractable for high dimensions. 
We show potent segmentation results for basic formulations of energy \eqref{eq:EA+MRF} with higher-dimensional 
features like RGBXY, RGBD, RGBM where standard MRF methods using model-fitting fail. 

On the other hand, standard \NC/ applications can also benefit from additional constraints 
\cite{stella:pami04,eriksson:JMIV11,Chew:ICCV15}. We show how to add a wide class of standard MRF potentials.
For example, standard \NC/ segmentation has weak alignment to contrast edges \cite{arbelaez2011contour}. 
While this can be addressed by post-processing, inclusion of the standard pair-wise Potts term \cite{BVZ:PAMI01,BJ:01} 
offers a principled solution. We show benefits from combining \NC/ with lower and higher-order constraints, 
such as sparsity or label costs \cite{LabelCosts:IJCV12}.
In the context of a general graph clustering,  higher-order consistency terms 
based on a $P^n$-Potts model \cite{kohli2009robust} also give significant improvements.

The synergy of the joint energy \eqref {eq:EA+MRF} is illustrated by juxtaposing the use of the pixel location information (XY) 
in standard \NC/ and MRF techniques.
The basic pairwise Potts model typically works on the nearest-neighbor pixel grids $\N_4$ or $\N_8$
where XY information helps contrast/edge alignment and enforces ``smooth'' segment boundaries.  
Wider connectivity Potts leads to denser graphs with slower optimization 
and poorer edge localization. In contrast, common \NC/ methods \cite{Shi2000} augment pixel color features with location
using relatively wide bandwidth for the XY dimension. This encourages segments with spatially ``compact'' regions. 
Narrower XY kernels improve edge alignment \cite{arbelaez2011contour},
but weaken regional color/feature consistency. On the other hand, very large XY kernels produce 
color-only clustering with spatially incoherent segments. Combining regional color consistency 
with spatial coherence in a single \NC/ energy requires a compromise when selecting XY bandwidth. 
Our general energy \eqref{eq:EA+MRF} can separate the regional consistency (\eg \NC/ clustering term)
from the boundary smoothness or edge alignment (\eg Potts potential). Interestingly, it may still be useful to augment colors 
with XY in the \NC/ term in \eqref{eq:EA+MRF} since XY kernel bandwidth $\sigma$ allows to separate similar-appearance objects 
at distances larger than $\sigma$, see Sec.\ref{sec:expxy}.




Adding high-order term $E_A$ to MRF potentials in \eqref{eq:EA+MRF} differs from fully-connected CRF \cite{koltun:NIPS11}.
Like many MRF/CRF models the method in \cite{koltun:NIPS11} lacks balancing. 
It is a denser version of the pairwise Potts model giving a trivial solution without a data term. 
In contrast, ratio-based objectives $E_A$ are designed for balanced clustering. 
In fact, our unary bounds for such objectives allow to combine them with fully connected CRF or any other MRF/CRF model 
with known solvers,  \eg mean field approximation \cite{koltun:NIPS11}.

Some earlier work also motivates a combination of pairwise clustering (\eg \NC/) with MRF potentials like the Potts model \cite{Kulis:ML09}.
They alter the Potts model to fit it to the standard trace-based formulation of \NC/. 
Our general bound approach can combine many pairwise clustering methods with any solvable discrete or continuous
regularization potentials. 

\subsubsection{Main contributions} \label{sec:contributions}

We propose a new energy model \eqref{eq:EA+MRF} for multi-label image segmentation and clustering,
efficient bound optimization algorithms, and demonstrate many useful applications. Our preliminary results
appear in \cite{kkm:iccv15} and \cite{NC-MRF:eccv16}. Our main contributions are:
\begin{itemize}
\item We propose a general multi-label segmentation or clustering energy  \eqref{eq:EA+MRF} combining 
pairwise clustering (\eg \NC/) with standard second or higher-order MRF regularization potentials. 
A pairwise clustering term can enforce balanced partitioning of observed image features and MRF 
terms can enforce many standard regularization constraints.
\item We obtain {\em kernel} (exact) and {\em spectral} (approximate) bounds for common pairwise clustering criteria
providing two {\em auxiliary functions} for joint energy \eqref{eq:EA+MRF}.
In the context of standard MRF potentials (\eg Potts, robust $P^n$-Potts, label cost)
we propose move-making algorithms for energy \eqref{eq:EA+MRF} generalizing 
$\alpha$-expansion and $\alpha\beta$-swap moves\footnote{Our
bounds for pairwise criteria can be also integrated into auxiliary functions with other standard 
regularization potentials (truncated, cardinality, TV) addressed by discrete (\eg message passing, relaxations, mean-field approximations) 
or continuous (\eg convex, primal-dual) algorithms.}\@.
\item For our {\em spectral bound}\footnote{Our term {\em spectral bound} means {\em ``spectral'' auxiliary function} 
in the context of bound optimization, not to be confused with bounds on eigenvalues.} 
we derive a new low-dimensional data embedding $\phi_p$ minimizing isometry errors w.r.t. affinities. 
These optimal embeddings complete the gap between standard \KM/ discretization heuristics in spectral relaxation \cite{Shi2000} and 
the exact \kKM/ formulations  \cite{bach:nips03,dhillon2004kernel}. 
\item Our experiments demonstrate that typical \NC/ applications benefit from extra MRF constraints, 
as well as, MRF segmentation benefit from the high-order \NC/ term encouraging balanced partitioning of image features.
In particular, our NC+MRF framework works for higher-dimensional features (\eg RGBXY, RGBD, RGBM) 
where standard model-fitting clustering \cite{Zhu:96,GrabCuts:SIGGRAPH04,LabelCosts:IJCV12} fails.
\end{itemize}

The rest of the paper is organized as follows. Sections \ref{sec:KBmain} and \ref{sec:SBmain} present our kernel and spectral bounds 
for \eqref{eq:EA+MRF} and detail combinatorial move making graph cut algorithms for its optimization.
Section \ref{sec:other_optimization} discusses a number of extensions for our bound optimization approach.
Sections \ref{sec:analysis} analyses kernel and bandwidth selection strategies.
Section \ref{sec:experiments} presents many experiments
where either pairwise clustering criteria benefit from the additional MRF constraints or common MRF formulations 
benefit from an additional clustering term for various high-dimensional features.

%
%

\begin{table*}
\centering
\begin{tabular}{|c|c|c|c|c|} \hline &&&& \\ 
{\bf objective} & {\bf matrix formulation}  & {\bf concave relaxation} &  {\bf ${\cal K}$ and $w$} & {\bf kernel bound}  \\ $E_A(S)$ & 
$e(X)$ in $\sum_k e(S^k)$  &    {\bf  $\e(X)$ \eqref{eq:e}}  &  
{\bf  in Lemma~\ref{lm:concave}}  & for $E_A(S)$ at $S_t$  \\[2ex] \hline &&&& \\
\AA/ \eqref{eq:AA2} &    $- \frac{X' A X }{{\bf 1}'X}  $  &   $- \frac{X' (\ds \Identity + A) X }{{\bf 1}'X}  $  & 
${\cal K} = \ds \Identity + A$, ~~~ $w={\bf 1}$  & \\[3ex]  \cline{1-4}  &&&& \\
\AC/ \eqref{eq:AC} &   $\frac{X' (D-A) X}{{\bf 1}'X}$  &  $ - \frac{X' (\ds \Identity +A-D) X}{{\bf 1}'X}$
& ${\cal K} = \ds \Identity + A - D$, ~~~ $w={\bf 1}$  &   $ \;\;\;\;\;\sum_k\; \nabla \e (S^k_t)'\, S^k \;\;+\;\;const\;\;\;\;\;$  \\[3ex]  \cline{1-4} &&&&
for $\nabla\e$ in \eqref{eq:ne} \\
\NC/ \eqref{eq:NC2} &  $-\frac{X' A X}{d'X}$ &      $-\frac{X' ( \ds D + A ) X}{d'X}$  &
${\cal K} = \ds D + A$, ~~~ $w=d$   &       \\[3ex] \hline
\end{tabular}
\centering
\caption{{\em Kernel bounds} for different pairwise clustering objectives $E_A(S)$. 
The second column shows formulations of these objectives $E_A(S)\equiv\sum_k e(S^k)$ 
using functions $e$ over segment indicator vectors $S^k\in\{0,1\}^{|\Omega|}$.  
The last column gives a unary (linear) upper bound for $E_A(S)$ at $S_t$ based on the first-order Taylor approximation 
of {\em concave relaxation} function $\e: {\mathcal R}^{\Omega}\rightarrow {\mathcal R}^1$ \eqref{eq:e}.
\label{tab:kernel_bounds}}
\end{table*}

%

\section{Kernel Bounds} \label{sec:KBmain}

First, we review the general bound optimization principle and present basic K-means as an example.
Section \ref{sec:KB} derives kernel bounds for standard pairwise clustering objectives. 
Without loss of generality, we assume symmetric affinities $A=A'$ since
non-symmetric ones can be equivalently replaced by $\frac{A+A'}{2}$, \eg see \eqref{eq:k_for_AA} in Sec.\ref{sec:kKM}.
Positive definiteness of $A$ is not assumed and {\em diagonal shifts} are discussed when needed.
Move-making bound optimization for energy \eqref{eq:EA+MRF} is discussed in Section \ref{sec:move-making}.

\subsection{Bound optimization and K-means}  \label{sec:KM_and_BO}

\begin{figure}
\centering
\includegraphics[width = 0.29\textwidth,page=2]{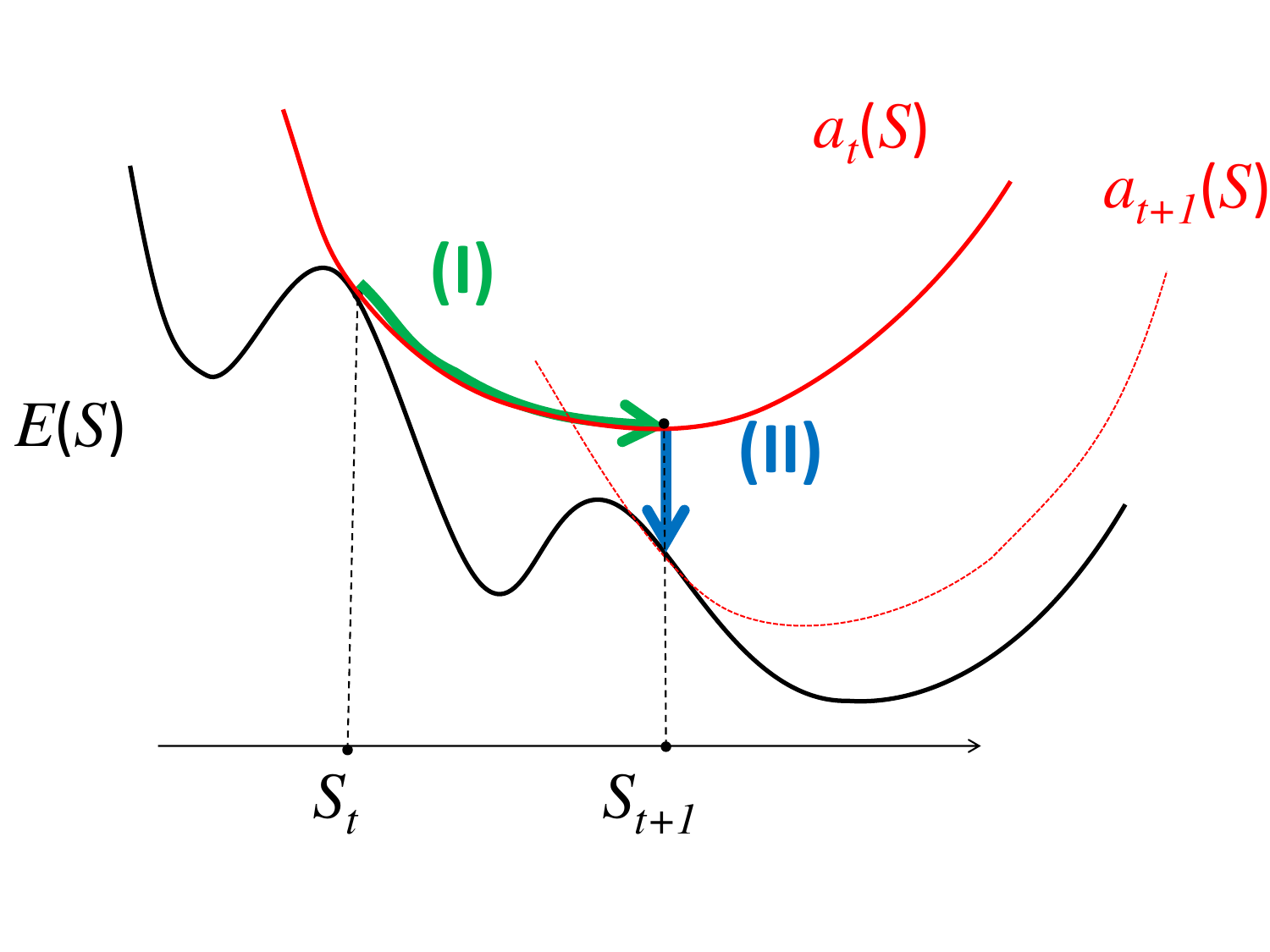}
 \caption{Iteration $t$ of the general bound optimization procedure for function $E(S)$ using auxiliary functions (bounds) $a_t(S)$.
 {\em Step I} minimizes $a_t(S)$. {\em Step II} computes the next bound $a_{t+1}(S)$.
\label{fig:BO}}
\end{figure}

In general, bound optimizers are iterative algorithms that optimize {\em auxiliary functions} (upper bounds) for a given energy $E(S)$ 
assuming that these auxiliary functions are more tractable than the original difficult optimization 
problem \cite{Lange2000,bilmes:UAI05,AuxCut:cvpr13,pbo:ECCV14}. Let $t$ be a current iteration index. Then
$a_t(S) $ is an auxiliary function of $E(S)$ at current solution $S_t$ if
\begin{subequations}
\begin{align}
E(S) \, &\leq \, a_t(S) \quad \forall S \label{general-second} \\
E(S_t) &= a_t(S_t). \label{general-third} 
\end{align}
\label{Eq:Auxiliary_function_conditions}
\end{subequations}
The auxiliary function is minimized at each iteration $t$ (Fig.~\ref{fig:BO}) 
\begin{equation} \label{eq:BOiteration}
S_{t+1} = \arg \min_{S} a_t(S).
\end{equation}
This procedure iteratively decreases original function $E(S)$ since $$E(S_{t+1}) \leq  a_t(S_{t+1}) \leq a_t(S_t) = E(S_t).$$ 
\begin{figure}
\centering
\includegraphics[width = 0.29\textwidth,page=3]{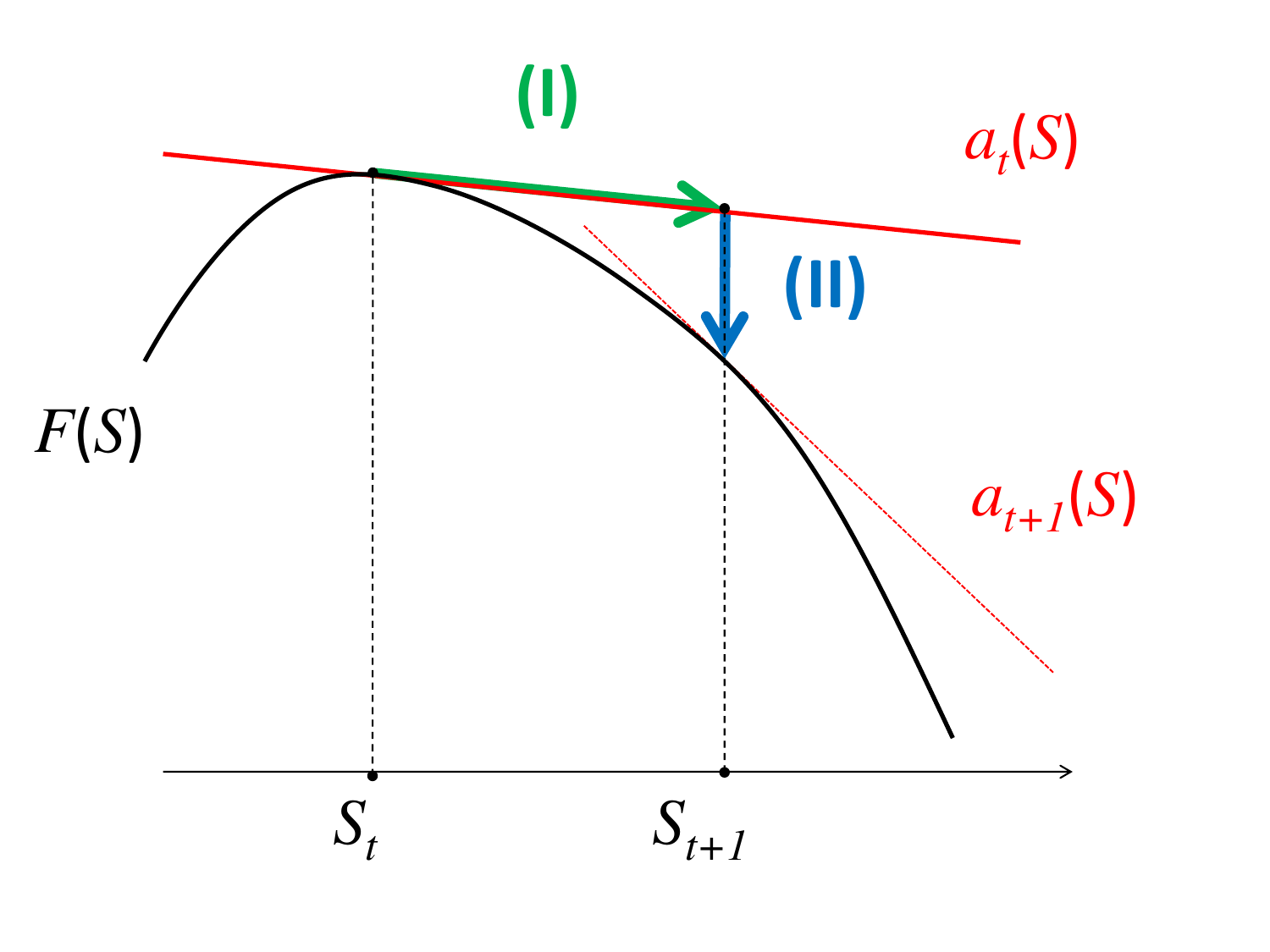}
 \caption{{\em K-means as linear bound optimization:} As obvious from formula \eqref{eq:kmixed0}, 
 the bound $a_t(S)$ 
in Theorem \ref{th:kKM_bound1} is a unary function of $S$. That is, \KM/ procedures (\ref{eq:exp_kKMproc},\ref{eq:imp_kKMproc}) 
correspond to optimization of linear auxiliary functions $a_t(S)$ for \KM/ objectives. 
Optimum $S_{t+1}$ is finite since optimization is over Boolean $S^k\in\{0,1\}^{|\Omega|}$. 
\label{fig:linearBO}}
\end{figure}

We show that standard \KM/ procedures \eqref{eq:exp_kKMproc}, \eqref{eq:imp_kKMproc}
correspond to bound optimization for K-means objective \eqref{eq:kKmeans1}. 
Note that variables $m_k$ in mixed objective $F(S,m)$ \eqref{eq:kmixed}
can be seen as relaxations of segment means $\mu_{S^k}$ \eqref{eq:mu_phi} in 
single-variable \KM/ objective $F(S)$ \eqref{eq:kKmeans1} since
\begin{eqnarray} 
\mu_{S^k} & = & \arg\min_{m_k} \sum_{p \in S^k} \|\phi_p- m_k \|^2     \nonumber  \\ 
\text{and}  \;\;\;\;F(S)  &  = & \min_m F(S,m).   \label{eq:various}
\end{eqnarray}

\begin{theor}[{\bf bound for \KM/}] \label{th:kKM_bound1}
Standard iterative K-means procedures (\ref{eq:exp_kKMproc},\ref{eq:imp_kKMproc}) can be seen as  bound optimization 
methods for K-means objectives $F(S)$ (\ref{eq:kKmeans1},\ref{eq:kKmeans3}) using auxiliary function
\begin{equation} \label{eq:kKM_bound1}
a_t(S) \;\;=\;\;  F(S,\mu_t)
\end{equation}
at any current segmentation $S_t = \{S^k_t\}$ with means $\mu_t = \{\mu_{S^k_t}\}$.
\end{theor}
\begin{proof}
Equation \eqref{eq:various} implies $a_t(S) \geq F(S)$. Since $a_t(S_t) =F(S_t)$ then $a_t(S)$ is an auxiliary function
for $F(S)$. Re-segmentation step \eqref{eq:exp_kKMproc} gives optimal segments $S_{t+1}$ 
minimizing the bound $a_t(S)$. The re-centering step minimizing $F(S_{t+1},m)$ 
for fixed segments gives means $\mu_{t+1}$ defining bound $a_{t+1}(S)$ for the next iteration. 
These re-segmentation (I) and re-centering (II) steps are also illustrated in Fig.~\ref{fig:BO} and Fig.~\ref{fig:linearBO}.  
\end{proof}

Theorem~\ref{th:kKM_bound1} could be generalized to {\em probabilistic K-means} \cite{UAI:97}
by stating that block-coordinate descent for distortion clustering or ML model fitting
\eqref{eq:pKmeans} is a bound optimization \cite{pbo:ECCV14,kkm:iccv15}. 
Theorem~\ref{th:kKM_bound1} can also be extended to pairwise and weighted versions of \KM/.
For example, one straightforward extension  is to
show that $F^w(S,\mu^w_t)$ \eqref{eq:wkmixed} with weighted means $\mu^w_t = \{\mu^w_{S^k_t}\}$ \eqref{eq:wmean} 
is a bound for weighted \KM/ objective $F^w(S)$ \eqref{eq:wkKM0}, \eg see Th.~\ref{th:wKM_bound} in Appendix A. 
Then, some bound for pairwise \wkKM/ energy \eqref{eq:wkKM} 
can also be derived (Cor.~\ref{cor:wKKM_bound}, Appendix A). It follows that bounds can be deduced for many pairwise clustering 
criteria using their reductions to various forms of \kKM/ reviewed in Sec.\ref{sec:kKM} or \ref{sec:NC-to-wKKM}.

Alternatively, the next Section~\ref{sec:KB} follows a more direct and intuitive approach to deriving pairwise clustering bounds 
motivated by the following simple observation. Note that function $a_t(S)$ in Theorem~\ref{th:kKM_bound1} is {\em unary} 
with respect to $S$. Indeed, functions $F(S,m)$ \eqref{eq:kmixed} or $F^w(S,m)$ \eqref{eq:wkmixed} can be written in the form
\begin{equation}  \label{eq:kmixed0} 
F(S,m) \;\;  \equiv \;\;  \sum_{k }\sum_{p }\;\;\;\;\;\;   \|\phi_p - m_k \|^2 \; S^k_p
\end{equation}
\begin{equation}  \label{eq:wkmixed0} 
F^w(S,m) \;\;  \equiv \;\;  \sum_{k }\sum_{p }\;  w_p   \|\phi_p - m_k \|^2\;  S^k_p
\end{equation}
highlighting the sum of unary terms for variables $S^k_p$. 
Thus, bounds for \KM/ or weighted \KM/ objectives are {\em modular} (linear) function of $S$. 
This simple technical fact has several useful implications that were previously overlooked. For example,
\begin{itemize}
\item in the context of bound optimization, \KM/ can be integrated with many regularization potentials 
whose existing solvers can easily work with extra unary (linear) terms
\item in the context of real-valued relaxation of indicators $S^k$, linearity of upper bound $a_t(S)$ \eqref{eq:kKM_bound1}
implies that the bounded function $F(S)\in{\mathcal C}^1$ \eqref{eq:kKmeans3} is {\em concave}, see Fig.\ref{fig:linearBO}.
\end{itemize}
In Section~\ref{sec:KB} we confirm that many standard pairwise clustering objectives in Sections \ref{sec:kKM} and \ref{sec:NC-to-wKKM}
have {\em concave relaxations}. Thus, their linear upper bounds easily follow from the corresponding 
first-order Taylor expansions, see Figure~\ref{fig:linearBO} and Table~\ref{tab:kernel_bounds}.

\subsection{Kernel Bounds for \AA/, \AC/, and \NC/} \label{sec:KB}

\begin{figure}
\centering
\includegraphics[width = 0.4\textwidth]{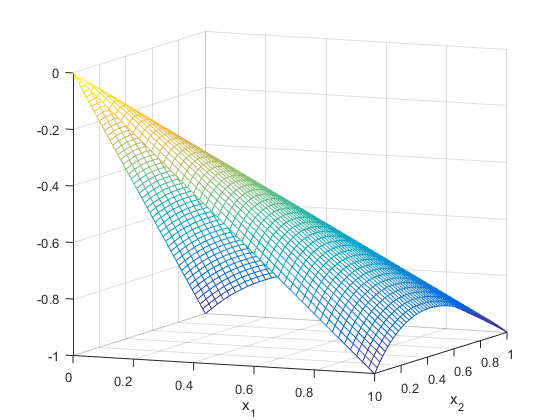}
\caption{Example: concave function $\e(X)=-\frac{X'X}{{\bf 1}'X}$ for $X\in[0,1]^2$. 
Note that convexity/concavity of similar rational functions with quadratic enumerator and linear denominator is known 
in other optimization areas, \eg \cite[p.72]{boyd:04} states convexity of $\frac{x^2}{y}$ for $y>0$ and 
\cite[exercise 3.14]{bazaraa:06} states convexity of  $\frac{(v'X)^2}{w'X}$  for $w'X>0$.
\label{fig:shift}}
\end{figure}

The next lemma is needed to state a bound for our joint energy \eqref{eq:EA+MRF} in Theorem \ref{th:kernel-bound}
for clustering terms \AA/, \AC/, or \NC/ in Table \ref{tab:kernel_bounds}.
\begin{lem}[{\bf concave relaxation}] \label{lm:concave}
Consider function $\e : {\mathcal R}^{\Omega}\rightarrow {\mathcal R}^1$ 
defined by matrix ${\cal K}$ and vector $w$ as 
\begin{equation} \label{eq:e}
\e(X)\;:=\;-\frac{X' {\cal K} X}{w'X}.
\end{equation}
Function $\e(X)$ above is concave over region $w'X > 0$ assuming that (symmetric) matrix ${\cal K}$ is 
positive semi-definite. (See Fig.~\ref{fig:shift})
\end{lem}
\begin{proof}
Lemma \ref{lm:concave} follows from the following expression for the Hessian of function $\e$ for symmetric ${\cal K}$
\begin{eqnarray}
\frac{\nabla\nabla \e}{2}  &  = & -\frac{\cal K}{w'X} + \frac{{\cal K} Xw' + wX'{\cal K}}{(w'X)^2} - \frac{wX'{\cal K}Xw'}{(w'X)^3} \nonumber  \\  
&   \equiv &  -\;\;\frac{1}{w'X}\left(\Identity-\frac{Xw'}{w'X}\right)'{\cal K}\left(\Identity-\frac{Xw'}{w'X}\right)  \nonumber
\end{eqnarray}
which is negative semi-definite for $w'X>0$ for p.s.d. ${\cal K}$. 
\end{proof}
The first-order Taylor expansion at current solution $X_t$  
$$T_t(X)\;:=\;\e(X_t)\;+\;\nabla \e(X_t)'\,(X-X_t)$$
is a bound for the concave function $\e(X)$ \eqref{eq:e}. 
Its gradient\footnote{Function $\e$ and gradient $\nabla \e$ are defined only at non-zero indicators $X_t$ 
where $w' X_t >0$. We can formally extend $\e$ to $X={\bf 0}$ and make the bound $T_t$ work for $\e$ at $X_t={\bf 0}$ 
with some {\em supergradient}. However, $X_t=0$ is not a problem in practice since it corresponds to an empty segment.} 
\begin{equation} \label{eq:ne}
\nabla \e(X_t) \;=\; w \;\frac{{X_t}'{\cal K}X_t}{(w'X_t)^2} \;-\; {\cal K}X_t \frac{2}{w'X_t}
\end{equation}
implies linear bound $T_t(X)$ for concave function $\e(X)$ at $X_t$
\begin{equation} \label{eq:Tbound}
T_t(X) \;\;\equiv\;\; \nabla \e (X_t)'\,X. 
\end{equation}

As shown in the second column of Table~\ref{tab:kernel_bounds}, common pairwise clustering objectives defined by affinity matrix 
$A$ such as \AA/ \eqref{eq:AA2}, \AC/ \eqref{eq:AC}, and \NC/ \eqref{eq:NC2} have the form
$$E_A(S)\;\;=\;\; \sum_k e(S^k)$$
with function $e(X)$ as in \eqref{eq:e} from Lemma~\ref{lm:concave}. However, 
arbitrary affinity $A$ may not correspond to a positive semi-definite ${\cal K}$ in \eqref{eq:e} and $e(X)$ may not be concave
for $X\in{\mathcal R}^{|\Omega|}$. However, the {\em diagonal shift} trick \cite{buhmann:pami03} 
in \eqref{eq:k_for_AA} works here too. The third column in Table~\ref{tab:kernel_bounds} 
shows concave function $\e(X)$ that equals $e(X)$ for any non-zero Boolean $X\in\{0,1\}^{|\Omega|}$, up to a constant. 
Indeed, for $\AA/$
$$  \e(X)   \;=  -\frac{X' (\ds \Identity + A) X }{{\bf 1}'X}   \;= - \frac{X' A X }{{\bf 1}'X} - \ds  \;\utc\;   e(X) $$
since $X'X = {\bf 1}'X$ for Boolean $X$. Clearly, $\ds \Identity + A$ is p.s.d. for sufficiently large $\ds$ and 
Lemma~\ref{lm:concave} implies that the first-order Taylor expansion $T_t(X)$ \eqref{eq:Tbound} is a linear bound for 
concave function $\e(X)$. Equivalence between $e$ and $\e$ over Booleans allows to use $T_t(X)$
as a bound for $e$ when optimizing over indicators $X$. Function $\e:{\mathcal R}^{|\Omega|}\rightarrow{\mathcal R}^1$ 
can be described as a {\em concave relaxation} of the high-order pseudo-boolean function $e:\{0,1\}^{|\Omega|}\rightarrow{\mathcal R}^1$.

Concave relaxation $\e$ for $\AC/$ in Table~\ref{tab:kernel_bounds} follows from the same diagonal shift $\ds \Identity$ as above. 
But \NC/ requires diagonal shift $\ds D$ with degree matrix $D=diag(d)$ as in \eqref{eq:dshift-NC}.  Indeed, 
\begin{equation} \label{eq:e-e}
 \e(X)   \;=  -\frac{X' (\ds D + A) X }{d'X}   \;= - \frac{X' A X }{d'X} - \ds  \;\utc\;   e(X)
 \end{equation}
since $X'DX\equiv X'diag(d)X = d'X$ for any Boolean $X$. Clearly, $\ds D + A$ is p.s.d. for sufficiently large $\ds$
assuming $d_p>0$ for all $p\in\Omega$. Concave relaxations and the corresponding Taylor-based bounds for $E_A(S)$
in Table~\ref{tab:kernel_bounds} imply the following theorem.
\begin{theor}[{\bf kernel bound}] \label{th:kernel-bound}
For any (symmetric) affinity matrix $A$ and any current solution $S_t$ 
the following is an auxiliary function for energy \eqref{eq:EA+MRF} with any clustering term $E_A(S)$
from Tab.\ref{tab:kernel_bounds}
\begin{equation}  \label{Kernel-bound}
 a_t(S)\;\;\; =\;\;\; \sum_k   \nabla \e(S^k_t)' \, S^k 
                      \;\;+\;\;\rp\;  \sum_{c\in\F}  E_{c} (S_{c})
\end{equation}
where $\e$ and $\nabla \e$ are defined in \eqref{eq:e}, \eqref{eq:ne} and
$\ds$ is large enough so that the corresponding $\cal K$ in Table \ref{tab:kernel_bounds} is positive semi-definite.
\end{theor}

\subsection{Move-making algorithms} \label{sec:move-making}

Combination \eqref{Kernel-bound} of regularization potentials with a unary/linear bound $\sum_k   \nabla \e(S^k_t)' \, S^k$ 
for high-order term $E_A(S)$ can be optimized with many standard discrete or continuous multi-label methods including 
graph cuts \cite{BVZ:PAMI01,Ishikawa:PAMI03}, message passing \cite{Kolmogorov:PAMI06}, LP relaxations \cite{werner2007linear}, 
or well-known continuous convex formulations \cite{chambolle2004algorithm,chambolle2011first,cremers2007review}. 
We focus on MRF regularizers (see Sec.\ref{sec:MRF_overview}) commonly addressed by graph cuts \cite{BVZ:PAMI01}. 
We discuss some details of  kernel bound optimization technique using such methods.

Step I of the bound optimization algorithm (Fig.\ref{fig:BO}) using auxiliary function $a_t(S)$  \eqref{Kernel-bound} 
for energy $E(S)$ \eqref{eq:EA+MRF} with regularization potentials 
reviewed in Sec.\ref{sec:MRF_overview} can be done via move-making methods \cite{BVZ:PAMI01,kohli2009robust,LabelCosts:IJCV12}.
Step II requires re-evaluation of the first term in \eqref{Kernel-bound}, \ie the kernel bound for $E_A$. 
Estimation of gradients $\nabla \e(S^k_t)$ in \eqref{eq:ne}  has complexity $O(K|\Omega|^2)$.

Even though the global optimum of $a_t$  at step I  (Fig.\ref{fig:BO}) is not guaranteed for general potentials $E_c$, 
it suffices to decrease the bound in order to decrease the energy, \ie \eqref{general-second} and \eqref{general-third} imply 
$$a_t(S_{t+1})\leq a_t(S_t)\;\;\;\;\Rightarrow\;\;\;\; E(S_{t+1})\leq E(S_t).$$ 
For example, Algorithm \ref{alg:kernelcut} shows a version of our kernel cut algorithm using $\alpha$-expansion \cite{BVZ:PAMI01}
for decreasing bound $a_t(S)$ in \eqref{Kernel-bound}. Other moves are also possible, for example  $\alpha\beta$-swap.

\begin{algorithm}[t]
\scriptsize
\SetAlgoLined
\SetKwInOut{Input}{Input}\SetKwInOut{Output}{Output}
 \Input{Affinity Matrix $\mathcal A$ of size $|\Omega|\times|\Omega|$;  initial labeling $S^1_0,...,S^K_0$}
  \Output{$S^1,...,S^K$: partition of the set $\Omega$}
 Find p.s.d. matrix ${\cal K}$ as in Table~\ref{tab:kernel_bounds}. Set $t:=0$\;
 \While{not converged}{
 Set $a_t(S)$ to be kernel bound \eqref{Kernel-bound} at current partition $S_t$\;
 \For{each label $\alpha\in\mathcal{L}=\{1,...,K\}$}{
  Find  $S_t := \arg\min{a_t(S)}$ within one $\alpha$ expansion of $S_t$;  
 }
 Set $t:=t+1$;
 }
 \caption{$\alpha$-Expansion for Kernel Cut}
 \label{alg:kernelcut}
\end{algorithm}

\begin{figure}
\begin{subfigure}[b]{0.45\textwidth}
	\includegraphics[width=\textwidth,trim=0 5mm 0 5mm]{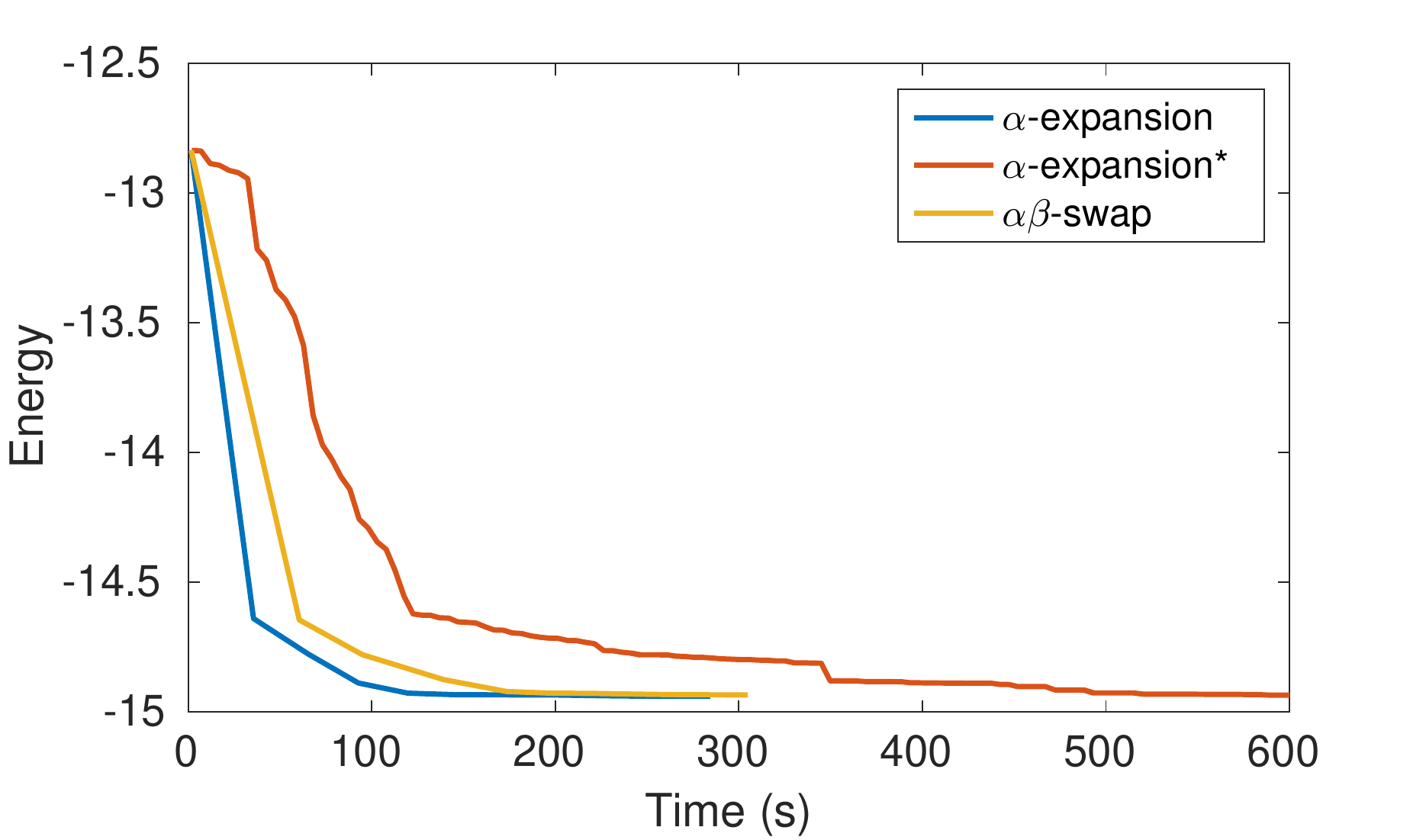}
	\subcaption{Versions of Kernel Cut}
	\vspace{2ex}
\end{subfigure}
\begin{subfigure}[b]{0.45\textwidth}
	\begin{center}
	\begin{tabular}{|m{12mm}|m{13mm}|c|c|}
		\hline
		Compare & against & \# of wins & p-value${}^\dag$\\
		\hline
		$\alpha$-expan-sion & $\alpha\beta$-swap & 135/200 & $10^{-6}$ \\
		\hline
		$\alpha$-expan-sion & $\alpha$-expan-sion${}^*$ & 182/200${}^\ddag$ & $10^{-34}{}^\ddag$  \\
		\hline
	\end{tabular}
	\end{center}
	{\scriptsize ${}^\dag$ The probability to exceed the given number of wins by random chance. \\[-1ex]
	${}^\ddag$ The algorithm stopped due to time limit (may cause incorrect number of wins).}
	\subcaption{BSDS500 training dataset}
\end{subfigure}
\caption{Typical energy evolution wrt different moves and frequency of bound updates. {\em $\alpha$-expansion} updates the bound after a round of expansions, {\em $\alpha$-expansion*} updates the bound after each expansion move. Initialization is a regular 5$\times$5 grid of patches.}
\label{fig:kc_versions}
\end{figure}

In general, \textit{tighter} bounds work better. Thus, we do not run iterative move-making algorithms for 
bound $a_t$ until convergence before re-estimating $a_{t+1}$. 
Instead, one can reestimate the bound either after each move or after a certain number of moves. 
One should decide the order of iterative move making and bound evaluation. In the case of $\alpha$-expansion, there are at least three options: updating the bound after a single expansion step, or after a single expansion loop, or after the convergence of $\alpha$-expansion. More frequent bound recalculation slows down the algorithm, but makes the bound tighter.  The particular choice generally depends on the trade-off between the speed and solution quality. However, in our experiments more frequent update does not always improve the energy, see Fig.\ref{fig:kc_versions}. We recommend updating the bound after a single loop of expansions, see Alg.\ref{alg:kernelcut}. We also evaluated 
a swap move version of our kernel cut method with bound re-estimation after a complete $\alpha\beta$-swaps loop, see Fig.\ref{fig:kc_versions}.

\section{Data Embeddings and Spectral Bounds}    \label{sec:SBmain}

This section shows a different bound optimization approach to pairwise clustering and joint regularization energy 
\eqref{eq:EA+MRF}. In contrast to the bounds explicitly using affinity $A$ or kernel matrices ${\cal K}$ in Sec.\ref{sec:KB}, 
the new approach is based on explicit use of isometric data embeddings $\phi$. 
While the general Mercer theorem guarantees existence of such possibly infinite dimensional Hilbert space embedding, 
we show finite dimensional Euclidean embedding
$$\phi:=[\phi_p] \;\;\;\;\quad \text{where} \quad \{\phi_p|p\in\Omega\}\;\;\subset\;\;{\mathcal R}^{|\Omega|}$$ 
with exact isometry (\ref{eq:dotproduct},\ref{eq:kernel_distance})  
to kernels ${\cal K}$ in Table~\ref{tab:kernel_bounds} and lower dimensional embeddings
$$\tilde{\phi}:=[\tilde{\phi}_p] \;\;\quad\text{where}\quad  \{\tilde{\phi}_p|p\in\Omega\}\;\;\subset\;\;{\mathcal R}^m \;\;\;\;\text{for} \;\;\;m\leq|\Omega|$$ 
that can approximate the same isometry with any accuracy level. The embeddings use eigen decompositions 
of the kernels.

Explicit embeddings allow to formulate exact or approximate {\em spectral bounds} for standard pairwise clustering 
objectives like \AA/, \AC/, \NC/. This approach is very closely related to spectral relaxation, 
see Sec.~\ref{sec:relation_to_spectral_clustering}. For example, optimization of our approximate spectral bounds for $m=K$
is similar to standard discretization heuristics using K-means over eigenvectors \cite{Shi2000}. Thus,
our bound optimization framework provides justification for such heuristics. More importantly, our spectral bounds 
allow to optimize joint energy \eqref{eq:EA+MRF} combing pairwise clustering objectives with common regularization terms.

Spectral bound is a useful alternative to kernel bounds in Sec.~\ref{sec:KB} with different complexity and other numerical properties.
In particular, spectral bound optimization using lower dimensional Euclidean embeddings $\tilde{\phi}$ for $m\ll|\Omega|$ is
often less sensitive to local minima. This may lead to better solutions, even though such $\tilde{\phi}$ 
only approximately isometric to given pairwise affinities. For $m=|\Omega|$ the spectral bound is mathematically 
equivalent to the kernel bound, but numerical representation is different.

\subsection{Exact and approximate embeddings $\phi$ for \kKM/}   \label{sec:embedding}

This section uses some standard methodology \cite{cox2000mds} to build the finite-dimensional embedding $\phi_p\equiv\phi(I_p)$ 
with exact or approximate isometry  (\ref{eq:dotproduct},\ref{eq:kernel_distance})  to any given positive definite kernel $k$ over 
finite data set $\{I_p|p\in\Omega\}$. 
As discussed in Sec.~\ref{sec:kKM}, \kKM/ and other pairwise clustering methods are typically defined by affinities/kernels $k$ 
and energy \eqref{eq:kKmeans3} rather than by high-dimensional embeddings $\phi$ with basic \KM/ formulation \eqref{eq:kKmeans1}. 
Nevertheless, data embeddings $\phi_p$ could be useful and some clustering techniques explicitly construct them
\cite{Shi2000,weiss:nips02,buhmann:pami03,belkin2003laplacian,bach:nips03,yu2015piecewise}. 
In particular, if dimensionality of the embedding space is relatively low then the basic iterative \KM/ procedure  \eqref{eq:exp_kKMproc} 
minimizing \eqref{eq:kKmeans1} could be more efficient than its kernel variant \eqref{eq:imp_kKMproc} for quadratic formulation \eqref{eq:kKmeans3}.
Even when working with a given kernel $k$ it may be algorithmically beneficial to build the corresponding isometric embedding $\phi$.
Below we discuss finite-dimensional Euclidean embeddings in ${\mathcal R}^m$ ($m\leq|\Omega|$) allowing to approximate
standard pairwise clustering via basic \KM/. 

\begin{figure}
\begin{center}
\begin{tabular}{c}
\includegraphics[width=0.9\columnwidth]{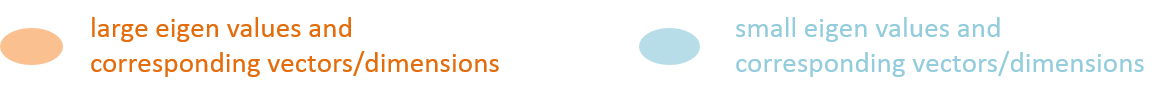} \\[3ex]
(a) decomposition ${\cal K} = V'\Lambda V$ \\[1ex]
\includegraphics[width=0.95\columnwidth]{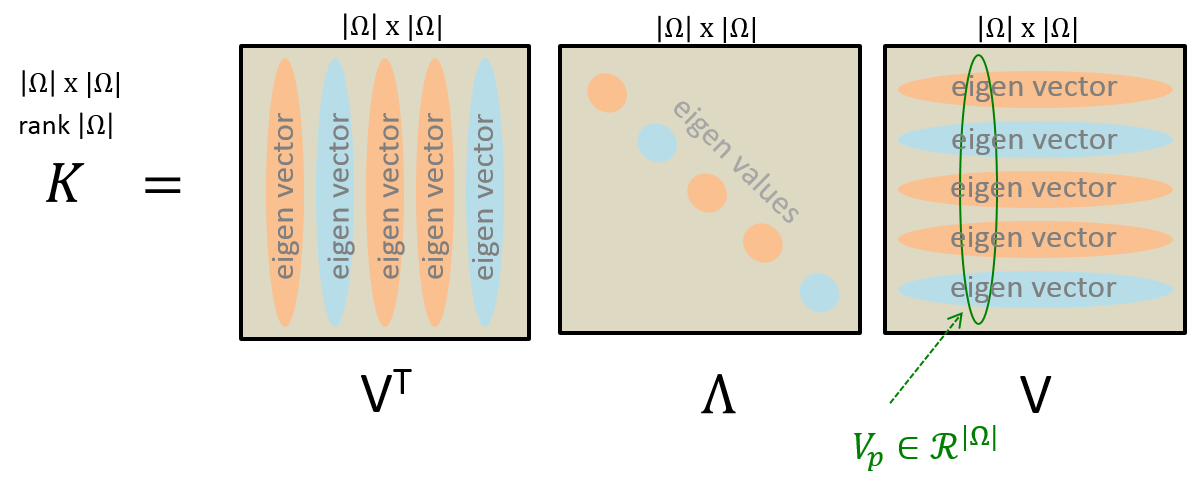} \\[3ex]
(b) decomposition $\tilde{\cal K}= (V^m)'\Lambda V^m$ for $m<|\Omega|$ \\[1ex]
\includegraphics[width=0.95\columnwidth]{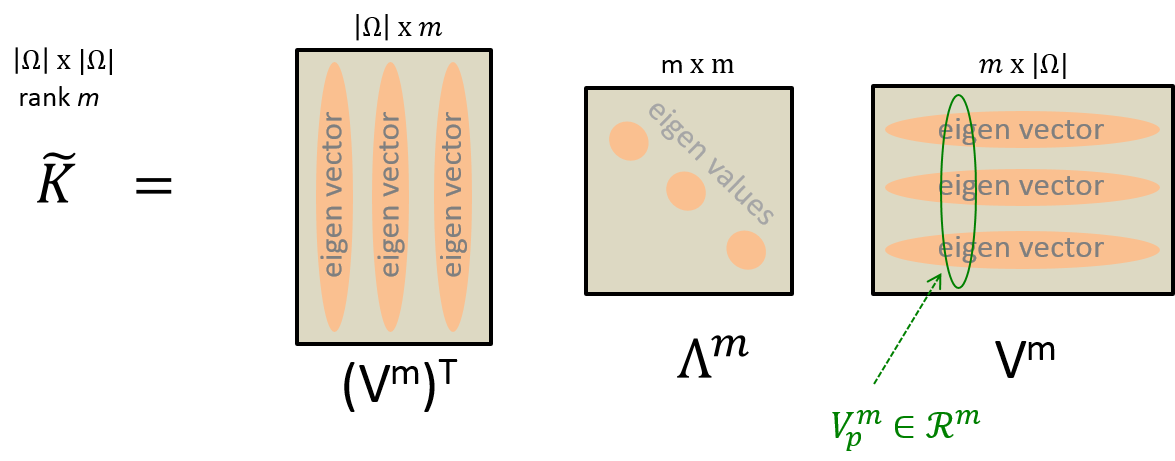} 
\end{tabular}
\end{center}
\caption{Eigen decompositions for kernel matrix ${\cal K}$ (a) and its rank $m$ approximation $\tilde{\cal K}$ (b)
minimizing Frobenius errors \eqref{eq:Fro_norm} \cite{cox2000mds}.
Decompositions (a,b) allow to build explicit embeddings (\ref{eq:phi_exact},\ref{eq:phi_appox}) 
isometric to the kernels, as implied by the Mercer theorem.}
        \label{fig:eigen_dec}
\end{figure}

First, we show an exact Euclidean embedding isometric to a given kernel. 
Any finite data set $\{I_p|p\in\Omega\}$ and any given kernel $k$ define a positive definite {\em kernel matrix}\footnote{If $k$ is
given as a continuous kernel $k(x,y):{\mathcal R}^N\times{\mathcal R}^N \rightarrow {\mathcal R}$ 
matrix ${\cal K}$ can be seen as its {\em restriction} to finite data set $\{I_p|p\in\Omega\} \subset{\mathcal R}^N $.}
$${\cal K}_{pq} = k(I_p,I_q)$$
of size $|\Omega|\times|\Omega|$.  The {\em eigen decomposition} of this matrix $${\cal K}=V'\Lambda V$$ involves 
diagonal matrix $\Lambda$ with non-negative eigenvalues and orthogonal matrix $V$ whose rows are eigenvectors, 
see Fig.\ref{fig:eigen_dec}(a). 
Non-negativity of the eigenvalues is important for obtaining decomposition $\Lambda = \sqrt{\Lambda}\cdot\sqrt{\Lambda}$
allowing to define the following Euclidean space embedding
\begin{equation} \label{eq:phi_exact}
\phi_p:=\sqrt{\Lambda}V_p \;\;\;\;\;\in {\mathcal R}^{|\Omega|}
\end{equation}
where $V_p$ are column of $V$, see Fig.\ref{fig:eigen_dec}(a). 
This embedding satisfies isometry (\ref{eq:dotproduct},\ref{eq:kernel_distance}) since
$$\langle\phi_p,\phi_q\rangle = (\sqrt{\Lambda} V_p)' (\sqrt{\Lambda} V_q) = {\cal K}_{pq} = k(I_p,I_q).$$

Note that \eqref{eq:phi_exact} defines a simple finite dimensional embedding $\phi_p\equiv\phi(I_p)$ only for subset of points 
$\{I_p|p\in\Omega\}$ in ${\mathcal R}^{N}$ based on a discrete kernel, \ie matrix ${\cal K}_{pq}$.
In contrast, Mercer's theorem should produce a more general infinite dimensional Hilbert embedding $\phi(x)$ 
for any $x\in{\mathcal R}^N$ by extending the eigen decomposition to continuous kernels $k(x,y)$.
In either case, however, the embedding space dimensionality is much higher than the original data space. For example,
$\phi_p$ in \eqref{eq:phi_exact}  has dimension $|\Omega|$, which is much  larger than the dimension 
of data $I_p$, \eg $3$ for RGB colors.

Embedding \eqref{eq:phi_exact} satisfying isometry (\ref{eq:dotproduct},\ref{eq:kernel_distance}) is not unique. 
For example, any decomposition ${\cal K}=G' G$, \eg Cholesky \cite{golub2012matrix}, defines 
a mapping $\phi^G_p:=G_p$ with desired properties. 
Also, rotational matrices $R$ generate a class of isometric embeddings $\phi^R_p:=R\phi_p$. 

\begin{figure}
\begin{center}
\begin{tabular}{cc}
\includegraphics[width=0.38\columnwidth]{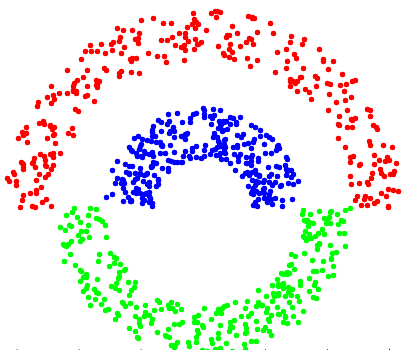} &
\includegraphics[width=0.32\columnwidth]{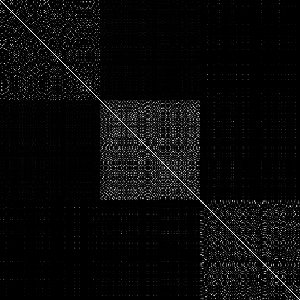}\\
(a) data $\{I_p|p\in\Omega\}$ & (b) Gaussian kernel matrix ${\cal K}$ \\[2ex]
\includegraphics[width=0.31\columnwidth]{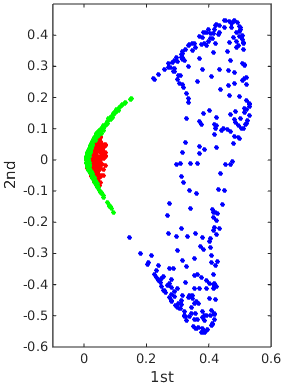} & 
\includegraphics[width=0.45\columnwidth]{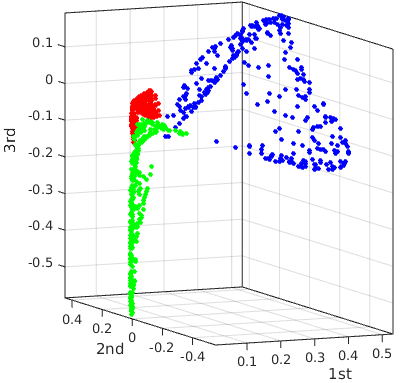} \\
(c) 2D embedding $\tilde{\phi}(I_p)$ & (d) 3D embedding $\tilde{\phi}(I_p)$
\end{tabular}
\end{center}
\caption{Low-dimensional Euclidean embeddings \eqref{eq:phi_appox} for $m=2$ and $m=3$ in (c,d) 
are approximately isometric to a given affinity matrix (b) over the data points in (a). The approximation error \eqref{eq:Fro_error}
decreases for larger $m$. While generated by standard MDS methodology \cite{cox2000mds}, 
it is intuitive to call embeddings $\phi$ in \eqref{eq:phi_exact} and \eqref{eq:phi_appox} as (exact or approximate) 
{\em isometry eigenmap} or {\em eigen isomap}. }
        \label{fig:embedding}
\end{figure}

It is easy to build lower dimensional embeddings by weakening the exact isometry requirements
(\ref{eq:dotproduct},\ref{eq:kernel_distance}) following the standard {\em multi-dimensional scaling} (MDS) 
methodology \cite{cox2000mds}, as detailed below. Consider a given rank $m<|\Omega|$ 
approximation ${\cal \tilde{K}}$ for kernel matrix ${\cal K}$ minimizing Frobenius norm errors \cite{cox2000mds}
\begin{equation} \label{eq:Fro_norm}
||{\cal K}-{\cal \tilde{K}}||_F \;\;:=\;\; \sum_{pq\in\Omega} ( {\cal K}_{pq} - {\cal \tilde{K}}_{pq})^2.
\end{equation} 
It is well known \cite{cox2000mds,golub2012matrix} that the minimum Frobenius error is achieved by
$${\cal \tilde{K}} = (V^m)'\Lambda^m V^m$$
where $V^m$ is a submatrix of $V$ including $m$ rows corresponding to the largest $m$ eignenvalues of ${\cal K}$
and $\Lambda^m$ is the diagonal matrix of these eigenvalues, see Fig.\ref{fig:eigen_dec}(b). 
The corresponding minimum Frobenius error is given by the norm of zeroed out eigenvalues
\begin{equation} \label{eq:Fro_error}
||{\cal K} - {\cal \tilde{K}}||_F \;\;=\;\;\sqrt{\lambda_{m+1}^2+\cdot\cdot\cdot+\lambda_{|\Omega|}^2 }. 
\end{equation}
It is easy to check that lower dimensional embedding
\begin{equation} \label{eq:phi_appox}
\tilde{\phi}_p:=\sqrt{\Lambda^m}V^m_p \;\;\;\;\;\in {\mathcal R}^{m}
\end{equation}
is isometric with respect to approximating kernel ${\cal \tilde{K}}$, that is
\begin{equation} \label{eq:approx_isometry}
\langle \tilde{\phi}_p,\tilde{\phi}_q\rangle \;=\; {\cal \tilde{K}}_{pq} \;\approx\; {\cal K}_{pq}.
\end{equation}
Fig.~\ref{fig:embedding} shows examples of low-dimensional approximate isometry embeddings \eqref{eq:phi_appox}
for a Gaussian kernel. 
Note that $\tilde{\phi}_p\in \mathcal{R}^m$ \eqref{eq:phi_appox} 
can be obtained from $\phi_p\in \mathcal{R}^{|\Omega|}$ \eqref{eq:phi_exact} 
by selecting coordinates corresponding to dimensions of the largest $m$ eigenvalues.

According to \eqref{eq:Fro_error} lower dimensional embedding $\tilde{\phi}_p$ in \eqref{eq:phi_appox}
is nearly-isometric to kernel matrix ${\cal K}$ if the ignored dimensions have sufficiently small eigenvalues.
Then \eqref{eq:phi_appox} may allow efficient approximation of kernel K-means.
For example, if sufficiently many eigenvalues are close to zero then a small rank $m$ approximation $\hat{\cal K}$ 
will be sufficiently accurate. In this case, we can use a basic iterative K-means procedure directly in ${\mathcal R}^m$
with $O(|\Omega|m)$ complexity of each iteration. In contrast, each iteration of the standard kernel K-means \eqref{eq:kKmeans3}
is $O(|\Omega|^2)$ in general\footnote{Without \KNN/ or other special kernel accelerations.}\@.

\begin{table*}
\centering
\begin{tabular}{|c|c|c|c|l|c|} \hline &&&&& \\ 
{\bf objective} & {\bf matrix formulation} & {\bf equivalent} \kKM/ (\ref{eq:kKmeans3},\ref{eq:wkKM})  & {\bf eigen decomposition}  &  
 {\bf embedding in} ${\mathcal R}^m,\;\;m\leq|\Omega|$ & 
 {\bf spectral bound}  \\ $E_A(S)$ & $e(X)$ in $\sum_k e(S^k)$ & as in \cite{buhmann:pami03,Dhillon:PAMI07} & $V' \Lambda V\;=\;\dots$  &  
with approx. isometry  \eqref{eq:approx_isometry}    & for $E_A(S)$ at $S_t$  \\[2ex] \hline &&&&& \\
\AA/ \eqref{eq:AA2}&  $- \frac{X'AX}{{\bf 1}'X}$ &  ${\cal K}=\;\ds  \Identity + A$ & $A$  & 
$\tilde{\phi}_p =\sqrt{\ds \Identity^m+\Lambda^m}V^m_p$  \;\eqref{eq:phi_AA}& 
  \\[3ex]  \cline{1-5}  &&&&&  $F(S,\mu_t)\;$ \eqref{eq:kmixed0}  \\
\AC/ \eqref{eq:AC} &  $\frac{X'(D-A)X}{{\bf 1}'X}$   & ${\cal K}=\;\ds \Identity+A-D$  & $D-A$  & 
$\tilde{\phi}_p =\sqrt{\ds \Identity^m-\Lambda^m}V^m_p$ \; \eqref{eq:phi_AC} &   for points $\tilde{\phi}_p$ 
 \\[3ex]  \hline &&&&& \\
\NC/ \eqref{eq:NC2} & $- \frac{X'AX}{d'X}$ & ${\cal K}=\;\ds D^{-1} + D^{-1}AD^{-1}$  & 
$D^{-\frac{1}{2}} A D^{-\frac{1}{2}}$  & 
$\tilde{\phi}_p = \sqrt{\frac{\ds  \Identity^m + \Lambda^m}{d_p}} V^m_p$ \; \eqref{eq:phi_NC} &  
$F^w(S,\mu^w_t)\;$ \eqref{eq:wkmixed0}   \\
&& weighted, $w_p=d_p$ &&&   for points $\tilde{\phi}_p$  \\ \hline
\end{tabular}
\centering
\caption{ {\em Spectral bounds} for objectives $E_A(S)$. 
The third column shows p.d. kernel matrices ${\cal K}$ for the equivalent \kKM/ energy \eqref{eq:kKmeans3}. 
Eigen decomposition for matrices in the forth column defines our Euclidean embedding $\tilde{\phi}_p\in{\mathcal R}^m$ 
(fifth column) isometric to ${\cal K}$ \eqref{eq:approx_isometry}.
Thus, K-means over $\tilde{\phi}_p$ approximates \kKM/ \eqref{eq:kKmeans3}.
Bounds for \KM/ (last column) follow from Th.~\ref{th:kKM_bound1} \& \ref{th:wKM_bound} where
$\mu_t = \{\mu_{S^k_t}\}$ are means \eqref{eq:mu_phi} and $\mu^w_t = \{\mu^w_{S^k_t}\}$
are weighted means \eqref{eq:wmean}. Functions $F(S,m)$ and $F^w(S,m)$ are modular (linear) w.r.t. $S$, 
see (\ref{eq:kmixed0},\ref{eq:wkmixed0}). 
\label{tab:spectral_bounds}}
\end{table*}

\begin{table*}
\centering
\begin{tabular}{|r|c|lcl|}
\hline
  & {\bf spectral relaxation \cite{Shi2000}}  
  & \multicolumn{3}{|c|}{{\bf common discretization heuristic \cite{von:tutorial2007}} ~~~(embedding \& K-means) } \\ \hline
\AA/ &  $A \mathbf{u}=\lambda \mathbf{u}$ &  
             $\tilde{\phi}_p:=U_p^K \;\;\;\;\equiv\;\;\;\; V^K_p $  & $\Leftarrow$ &   $V' \Lambda V = A$      \\ \hline
\AC/ & $(D-A)\mathbf{u} = \lambda \mathbf{u}$ &
            $\tilde{\phi}_p:=U_p^K \;\;\;\;\equiv\;\;\;\; V^K_p $    & $\Leftarrow$ &   $V' \Lambda V = D-A$              \\ \hline
\NC/ & $(D-A)\mathbf{u}=\lambda D \mathbf{u}$ & 
           $\tilde{\phi}_p:=U_p^K \;\;\;\;\equiv\;\;\;\;   [V^K D^{-\frac{1}{2}}]_p^{rn} $   & $\Leftarrow$ &   
           $V' \Lambda V = D^{-\frac{1}{2}} A D^{-\frac{1}{2}}$       \\ \hline
\end{tabular}
\caption{{\em Spectral relaxation} and discretization heuristics for objectives for pairwise clustering objectives $E_A(S)$ for affinity $A$. 
The corresponding {\em degree} matrix $D$ is diagonal with elements $d_p := \sum_q A_{pq}$. 
To extract integer labeling from the relaxed solutions produced by the eigen systems (second column), 
spectral methods often apply basic \KM/ to some ad hoc data embedding $\tilde{\phi}$ (last column)
based on the first $K$ unit eigenvectors $\mathbf{u}$, the rows of matrix $U^K$.  
While our main text discusses some variants, the most basic idea \cite{Shi2000,von:tutorial2007} is to use the columns of $U^K$ 
as embedding $\tilde{\phi}_p$.
For easier comparison, the last column also shows equivalent representations of this embedding based on 
the same eigen decompositions $V' \Lambda V$ as those used for our isometry eigenmaps in Tab.~\ref{tab:spectral_bounds}. 
In contrast, our embeddings are derived from justified approximations of the original non-relaxed 
\AA/, \AC/, or \NC/ objectives. Note that \NC/ corresponds to a {\em weighted} case of K-means 
with data point weights $w_p=d_p$ \cite{bach:nips03,dhillon2004kernel}, see \eqref{eq:Atilde} in Section \ref{sec:NC-to-wKKM}.} 
\label{tab:spectralsolution}
\end{table*}


There is a different way to justify approximate low-dimensional embedding $\tilde{\phi}_p$  ignoring small eigenvalue 
dimensions in $\phi_p$. The objective in \eqref{eq:kKmeans3} for exact kernel ${\cal K}$ is equivalent to the basic K-means \eqref{eq:kmixed} 
over points $\phi_p$ \eqref{eq:phi_exact}. The latter can be shown to be
equivalent to (probabilistic) K-means  \eqref{eq:distortion} over columns $V_p$ in orthonormal matrix $V$ using
weighted distortion measure $$||V_p - \mu||_{\Lambda}^2 := \sum_{i=1}^{|\Omega|} \lambda_i (V_p[i] - \mu[i])^2 = ||\phi_p - \sqrt{\Lambda}\mu||^2$$
where index $[i]$ specifies coordinates of the column vectors. Thus, a good approximation is achieved when ignoring 
coordinates for small enough eigenvalues contributing low weight in the distortion above. 
This is equivalent to K-means \eqref{eq:kmixed}  over points \eqref{eq:phi_appox}.

\subsection{Spectral Bounds for \AA/, \AC/, and \NC/ }

The last Section showed that \kKM/ clustering with given p.s.d. kernel ${\cal K}$ can be approximated by basic \KM/ 
over low-dimensional Euclidean embedding $\tilde{\phi}\in{\mathcal R}^m$ \eqref{eq:phi_appox} with approximate 
isometry to ${\cal K}$ \eqref{eq:approx_isometry}. Below we use equivalence of standard pairwise clustering criteria to
\kKM/, as discussed in Sections~\ref{sec:kKM} and \ref{sec:NC-to-wKKM}, to derive the corresponding low-dimensional 
embeddings for \AA/, \AC/, \NC/.
Then, equivalence of \KM/ to bound optimization (Theorem~\ref{th:kKM_bound1}) allows to formulate our 
approximate {\em spectral bounds} for the pairwise clustering and joint energy \eqref{eq:EA+MRF}.
The results of this Section are summarized in Table \ref{tab:spectral_bounds}.
For simplicity, assume symmetric affinity matrix $A$. If not, equivalently replace $A$ by $\frac{A+A'}{2}$.

{\bf Average association (\AA/):} 
Diagonal shift ${\cal K} = \ds \Identity +A$ in \eqref{eq:k_for_AA} converts $\AA/$ \eqref{eq:AA2} with $A$ 
to equivalent \kKM/ \eqref{eq:kKmeans3} with p.d. kernel ${\cal K}$. 
We seek rank-$m$ approximation ${\cal \tilde{K}}$ minimizing Frobenius error $||{\cal K}-{\cal \tilde{K}}||_{F}$. 
Provided eigen decomposition $A=V' \Lambda V$,
equation \eqref{eq:phi_appox} gives low-dimensional embedding  (also in Tab.~\ref{tab:spectral_bounds})
\begin{equation} \label{eq:phi_AA}
\tilde{\phi}_p \;\;=\;\;\sqrt{\ds \Identity^m +\Lambda^m}V^m_p
\end{equation}
corresponding to optimal approximation kernel ${\cal \tilde{K}}$. It follows that \KM/ \eqref{eq:exp_kKMproc} over this embedding 
approximates \AA/ objective \eqref{eq:kKmeans3}. 
Note that  the eigenvectors (rows of matrix $V$, Fig.~\ref{fig:eigen_dec}) 
also solve the spectral relaxation for \AA/ in Tab.~\ref{tab:spectralsolution}. However, ad hoc discretization by
\KM/ over points $V^K_p$ may differ from the result for points \eqref{eq:phi_AA}.    

{\bf Average cut (\AC/):} As follows from objective \eqref{eq:AC} and diagonal shift \eqref{eq:k_for_AA} \cite{buhmann:pami03}, 
average cut clustering for affinity $A$ is equivalent to minimizing \kKM/ objective with kernel ${\cal K}=\ds \Identity+A-D$
where $D$ is a diagonal matrix of node degrees $d_p = \sum_q A_{pq}$. 
Diagonal shift $\ds \Identity$ is needed to guarantee positive definiteness of the kernel. 
Eigen decomposition  for $D-A=V' \Lambda V$ implies ${\cal K}=V' (\ds \Identity-\Lambda) V$. 
Then, \eqref{eq:phi_appox} implies rank-$m$ approximate isometry embedding (also in Tab.~\ref{tab:spectral_bounds})
\begin{equation} \label{eq:phi_AC}
\tilde{\phi}_p \;\;=\;\; \sqrt{\ds \Identity^m-\Lambda^m}V^m_p
\end{equation}
using the same eigenvectors (rows of $V$) that solve \AC/'s spectral relaxation in Tab.~\ref{tab:spectralsolution}.
However, standard discretization heuristic using \KM/ over $\tilde{\phi}_p=V^K_p$ may differ from the results 
for our approximate isometry embedding $\tilde{\phi}_p$ \eqref{eq:phi_AC} due to different weighting.

{\bf Normalized cut (\NC/):} According to \cite{dhillon2004kernel} and a simple derivation in Sec.\ref{sec:NC-to-wKKM}
normalized cut for affinity $A$ is equivalent to weighted \kKM/ with 
kernel ${\cal K}=\ds D^{-1} + D^{-1}AD^{-1}$ \eqref{eq:dshift-NC2} 
and node weights $w_p=d_p$ based on their degree. Weighted \kKM/  \eqref{eq:wkKM} can be
interpreted as \KM/ in the embedding space with weights $w_p$ for each point $\phi_p$ as in (\ref{eq:wkmixed},\ref{eq:wmean}).
The only issue is computing $m$-dimensional embeddings approximately isometric to ${\cal K}$. 
Note that previously discussed solution $\tilde{\phi}$ in \eqref{eq:phi_appox} uses eigen decomposition 
of matrix ${\cal K}$ to minimize the sum of quadratic errors between ${\cal K}_{pq}$ and approximating kernel
$\tilde{\cal K}_{pq}=\langle\tilde{\phi}_p,\tilde{\phi}_q\rangle$.
This solution may still be acceptable, but in the context of weighted points it seems natural to 
minimize an alternative approximation measure taking $w_p$ into account. For example,
we can find rank-$m$ approximate affinity matrix $\tilde{\cal K}$ minimizing the sum of weighted squared errors
\begin{equation}  \label{eq:wFro_error}
\sum_{pq\in\Omega} w_p w_q ({\cal K}_{pq}-\tilde{\cal{K}}_{pq})^2 = 
||D^{\frac{1}{2}} ( {\cal{K}} - \tilde{\cal{K}} ) D^{\frac{1}{2}}  ||_F .
\end{equation}
Substituting ${\cal K}=\ds D^{-1} + D^{-1}AD^{-1}$ gives an equivalent objective 
\begin{equation*}
 ||D^{-\frac{1}{2}} (\ds D + A) D^{-\frac{1}{2}} -  D^{\frac{1}{2}} \tilde{\cal{K}} D^{\frac{1}{2}} ||_F .
\end{equation*}
Consider rank-$m$ matrix $\tilde{M} := D^{\frac{1}{2}} \tilde{\cal{K}} D^{\frac{1}{2}}$ as a new
minimization variable.  Its optimal value $(V^m)'(\ds  \Identity^m + \Lambda^m) V^m$ follows from 
$D^{-\frac{1}{2}} (\ds D + A) D^{-\frac{1}{2}} = V'(\ds \Identity +  \Lambda)V $
for eigen decomposition 
\begin{equation}  \label{eq:eigen_dec1}
D^{-\frac{1}{2}} A D^{-\frac{1}{2}} \equiv V'\Lambda V.
\end{equation} 
Thus, optimal rank-$m$ approximation kernel $\tilde{\cal{K}}$ is 
\begin{equation} \label{eq:tilde K for NC}
\tilde{\cal{K}} = D^{-\frac{1}{2}} (V^m)'(\ds  \Identity^m + \Lambda^m) V^m D^{-\frac{1}{2}}. 
\end{equation} 
It is easy to check that $m$-dimensional embedding (also in Tab.~\ref{tab:spectral_bounds})
\begin{equation} \label{eq:phi_NC}
\tilde{\phi}_p\;\;=\;\;\sqrt{\frac{\ds  \Identity^m + \Lambda^m}{d_p}} V^m_p
\end{equation}
is isometric to kernel $\tilde{\cal K}$, that is 
$\langle\tilde{\phi}_p,\tilde{\phi}_q\rangle = \tilde{\cal K}_{pq}$. Therefore, 
weighted \KM/ \eqref{eq:wkmixed} over low-dimensional embedding $\tilde{\phi}_p$  \eqref{eq:phi_NC}  with weights $w_p=d_p$
approximates \NC/ objective \eqref{eq:NC2}.

{\bf Summary:} The ideas above can be summarized as follows.
Assume \AA/, \AC/, or \NC/ objectives $E_A(S)$ with (symmetric) $A$. 
The third column in Table~\ref{tab:spectral_bounds} shows
kernels ${\cal K}$ for equivalent \kKM/ objectives $F(S)$ (\ref{eq:kKmeans3},\ref{eq:wkKM}).  
Following {\em eigenmap} approach (Fig.\ref{fig:eigen_dec}), 
we find rank-$m$ approximate kernel $\tilde{\cal K}\approx{\cal K}$ minimizing  Frobenius error 
$\|\tilde{\cal K}-{\cal K}\|_F$ \eqref{eq:Fro_norm} or its weighted version \eqref{eq:wFro_error}
and deduce embeddings $\tilde{\phi}_p\in{\mathcal R}^m$ \eqref{eq:phi_AA}, \eqref{eq:phi_AC}, \eqref{eq:phi_NC} 
satisfying isometry $$\tilde{\phi}'_p \tilde{\phi}_q = \tilde{\cal K}_{pq} \approx {\cal K}_{pq}.$$ 
Basic K-means objective $\tilde{F}(S,m)$ (\ref{eq:kmixed},\ref{eq:wkmixed}) for $\{\tilde{\phi}_p\}$ is equivalent to 
\kKM/ energy $\tilde{F}(S)$ (\ref{eq:kKmeans3},\ref{eq:wkKM}) for kernel $\tilde{\cal K}\approx{\cal K}$ 
and, therefore, approximates the original pairwise clustering objective 
$$ \tilde{F}(S,\mu_S)  \;\;\utc\;\;  \tilde{F}(S) \;\;\approx\;\;  F(S)\;\;\utc\;\; E_A(S).$$
Theorem \ref{th:kKM_bound1} gives unary (linear) bound $\tilde{F}(S,\mu_t)$ (\ref{eq:kmixed0},\ref{eq:wkmixed0}) for 
objective $\tilde{F}(S)$ (\ref{eq:kmixed},\ref{eq:wkmixed}). We refer to $\tilde{F}(S,\mu_t)$ as a {\em  spectral auxiliary function}
for approximate optimization of $E_A(S)$ (last column in Table~\ref{tab:spectral_bounds}). 
We will also simply call $\tilde{F}(S,\mu_t)$ a {\em  spectral bound}, not to be confused with 
a similar term used for matrix eigenvalues.

Similarly to {\em kernel bound} in Section~\ref{sec:KB}, {\em spectral bound} is useful for optimizing \eqref{eq:EA+MRF}.
In fact, for $m=|\Omega|$ the spectral bounds (Tab.\ref{tab:spectral_bounds}) 
are algebraically equivalent to the kernel bounds (Tab.\ref{tab:kernel_bounds}) since
$\tilde{\cal K}={\cal K}$, see \eqref{eq:Fro_error}. For $m<|\Omega|$ we have $\tilde{\cal K}\approx{\cal K}$ and $\tilde{F} \;\;\approx\;\;  F$.
Therefore, we can iteratively minimize energy $E(S)$ \eqref{eq:EA+MRF} by applying bound optimization to its {\em spectral approximation}
\begin{equation} \label{eq:approxEA+MRF}
\tilde{E}(S)\;\;=\;\; \tilde{F}(S) +\;\;\rp\;\; \sum_{c\in\F}  E_{c} (S_{c}) 
\end{equation}
or its {\em weighted spectral approximation}
\begin{equation} \label{eq:wapproxEA+MRF}
\tilde{E}(S)\;\;=\;\; \tilde{F}^w(S) +\;\;\rp\;\; \sum_{c\in\F}  E_{c} (S_{c}) .
\end{equation}

\begin{theor}[{\bf spectral bound}] \label{th:spectral-bound}
For any (symmetric) affinity matrix $A$ assume sufficiently large diagonal shift $\ds$ generating p.s.d. kernel ${\cal K}$ 
as in Table~\ref{tab:spectral_bounds}. Then, auxiliary function
\begin{equation}  \label{Spectral-bound}
 \tilde{a}_t(S)\;\;\; =\;\;\; \tilde{F}(S,\mu_t)  \;\;+\;\;\rp\;  \sum_{c\in\F}  E_{c} (S_{c})
\end{equation}
using $\tilde{F}(S,m)$ (\ref{eq:kmixed0},\ref{eq:wkmixed0})
with embedding $\{\tilde{\phi}_p\}\subset{\mathcal R}^m$ in Tab.~\ref{tab:spectral_bounds}
is a bound for joint energy (\ref{eq:approxEA+MRF},\ref{eq:wapproxEA+MRF}) approximating \eqref{eq:EA+MRF} 
as $m\rightarrow|\Omega|$.
\end{theor}

\begin{figure}[t]
\begin{center}
\includegraphics[width=0.45\columnwidth]{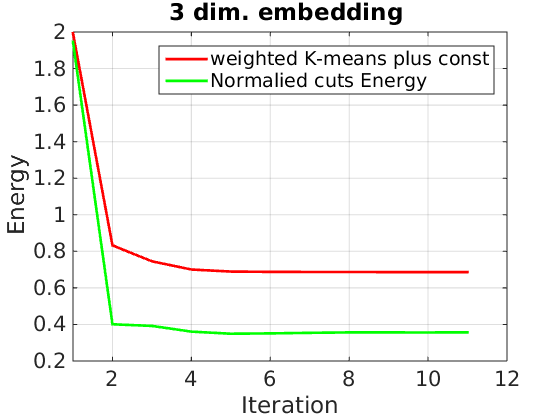} 
\includegraphics[width=0.45\columnwidth]{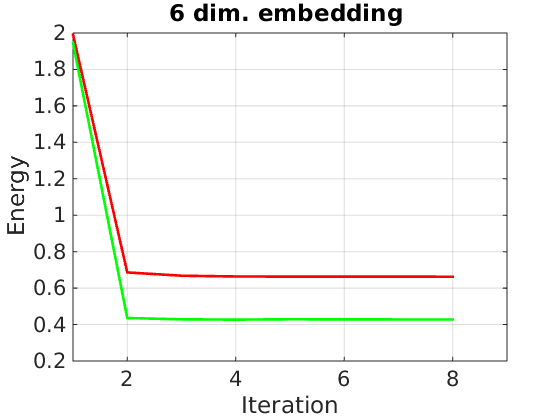}\\
\includegraphics[width=0.45\columnwidth]{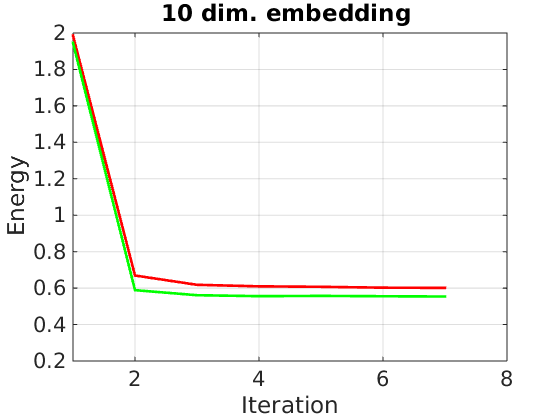}
\includegraphics[width=0.45\columnwidth]{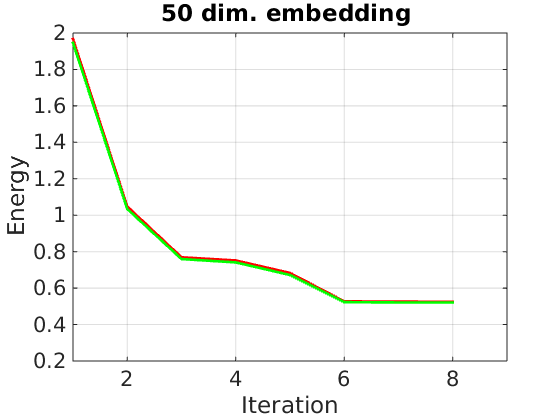}
\end{center}
\caption{For data and affinity matrix in Fig. \ref{fig:embedding}, we run weighted K-means with our approximate embedding. The approximation errors $||{\cal K}-{\cal \tilde{K}}||_F^2/||{\cal K}||_F^2$ for 3, 6, 10 and 50 dim. embedding are 58\%, 41\%, 27\% and 3\% respectively. We compute weighted K-means energy (up to a const) and normalized cuts energy for solution obtained at each iteration. We observed that normalized cuts energy indeed tends to decrease during iterations of K-means. Even 10 dim. embedding gives good alignment between K-means energy and normalized cuts energy. Higher dimensional embedding gives better energy approximation, but not necessarily better solution with lower energy.}
        \label{fig:kmeansonembedding}
\end{figure}

Approximation quality \eqref{eq:Fro_error} depends on omitted eigenvalues $\lambda_i$ for $i>m$.
Representative examples in Fig.\ref{fig:eigenvalues} show that relatively few
eigenvalues may dominate the others. Thus, practically good approximation with
small $m$ is possible. Larger $m$ are computationally expensive since more eigenvalues/vectors are needed.
Interestingly, smaller $m$ may give better optimization since K-means in higher-dimensional spaces 
may be more sensitive to local minima. Thus, spectral bound optimization for smaller $m$ may find solutions with 
lower energy, see Fig.\ref{fig:kmeansonembedding}, even though the quality of approximation is better for larger $m$. 

Similarly to the kernel bound algorithms discussed in Section~\ref{sec:move-making}
one can optimize the approximate spectral bound \eqref{Spectral-bound} for energy \eqref{eq:EA+MRF}
using standard algorithms for regularization.
This follows from the fact that the first term in \eqref{Spectral-bound} is unary (linear). 
Algorithms~\ref{alg:spectralcut} shows a representative (approximate) bound optimization
technique for \eqref{eq:EA+MRF} using move-making algorithms \cite{boykov2001fast}.
Note that for $\rp=0$ (no regularization terms) our bound optimization Algorithm~\ref{alg:spectralcut} reduces to
basic K-means over approximate isometry embeddings $\{\tilde{\phi}_p\}\subset{\mathcal R}^m$ similar but not identical to common
discretization heuristics in spectral relaxation methods.
\begin{algorithm}[t]
\scriptsize
\SetAlgoLined
\SetKwInOut{Input}{Input}\SetKwInOut{Output}{Output}
 \Input{Affinity Matrix $\mathcal A$ of size $|\Omega|\times|\Omega|$;  initial labeling $S^1_0,...,S^K_0$}
  \Output{$S^1,...,S^K$: partition of the set $\Omega$}
 Find top $m$ eigenvalues/vectors $\Lambda^m,V^m$ for a matrix in the $4^{th}$ col. of Tab.~\ref{tab:spectral_bounds} \;
 Compute embedding $\{\tilde{\phi}_p\}\subset{\mathcal R}^m$ for some $\ds$ and set $t:=0$\;
 \While{not converged}{
 Set $\tilde{a}_t(S)$ to be the spectral bound \eqref{Spectral-bound} at current partition $S_t$\;
 \For{each label $\alpha\in\mathcal{L}=\{1,...,K\}$}{
 Find $S_t:=\arg\min{\tilde{a}_t(S)}$ within one $\alpha$ expansion of $S_t$;
 }
 Set $t:=t+1$;
 }
 \caption{$\alpha$-Expansion for Spectral Cut}
 \label{alg:spectralcut}
\end{algorithm}

\subsection{Relation to spectral clustering} \label{sec:relation_to_spectral_clustering}

Our approximation of pairwise clustering such as \NC/ via basic \KM/ over low dimensional embeddings $\tilde{\phi}_p$
is closely related to popular spectral clustering algorithms \cite{Shi2000,weiss:nips02,belkin2003laplacian}
using eigen decomposition for various combinations of kernel, affinity, distortion, laplacian, or other matrices. 
Other methods also build low-dimensional Euclidean embeddings \cite{weiss:nips02,belkin2003laplacian,yu2015piecewise} 
for basic \KM/ using motivation different from isometry and approximation errors with respect to given affinities. 
We are mainly interested in discussing relations to spectral methods approximately optimizing
pairwise clustering criteria such as \AA/, \AC/, and \NC/ \cite{Shi2000}. 

Many spectral relaxation methods also use various eigen decompositions to build explicit data embeddings 
followed by basic K-means. In particular, the smallest or largest eigenvectors for the (generalized) eigenvalue problems 
in Table~\ref{tab:spectralsolution} give well-known exact solutions for the relaxed problems. 
In contrast to our approach, however, the final K-means stage in spectral methods is often presented 
without justification \cite{Shi2000,von:tutorial2007,arbelaez2011contour} as a heuristic for quantizing
the relaxed continuous solutions into a discrete labeling. It is commonly understood that 
\begin{quote} 
{\em ``...~~there is nothing principled about using the K-means algorithm in this step''} 
~~~~~~~~~(Sec.~8.4 in \cite{von:tutorial2007})
\end{quote} 
or that 
\begin{quote} 
{\em ``...~~K-means introduces additional unwarranted assumptions.''}
\hspace{2.4cm} (Sec.~4 in \cite{Yu:iccv03})
\end{quote}
Also, typical spectral methods use $K$ eigenvectors
solving the relaxed $K$-cluster problems followed by \KM/ quantization. In contrast, we choose the number 
of eigenvectors $m$ based on Frobenius error for isometry approximation \eqref{eq:Fro_error}.
Thus, the number $m$ is independent from the predefined number of clusters.

Below we juxtapose our approximate isometry low-dimensional embeddings in Table~\ref{tab:spectral_bounds} with 
embeddings used for ad-hoc discretization by the standard spectral relaxation methods in Table~\ref{tab:spectralsolution}.
While such  embeddings are similar, they are not identical. 
Thus, our Frobenius error argument offers a justification and minor corrections for \KM/ heuristics in spectral methods, 
even though the corresponding methodologies are unrelated. More importantly, our bound formulation allows integration of
pairwise clustering with additional regularization constraints \eqref{eq:EA+MRF}.

{\bf Embeddings in spectral methods for \NC/:}
Despite similarity, there are differences between our low-dimensional embedding \eqref{eq:phi_NC} 
provably approximating kernel ${\cal K}=\ds D^{-1} +D^{-1}AD^{-1}$ for the \kKM/ formulation of \NC/ \cite{bach:nips03,dhillon2004kernel}
and common ad-hoc embeddings used for \KM/ discretization step in the spectral relaxation methods.
For example, one such discretization heuristic \cite{Shi2000,von:tutorial2007} uses embedding $\tilde{\phi}_p$ 
(right column in Tab.~\ref{tab:spectralsolution})  defined by the columns of matrix $U^K$ 
whose rows are the $K$ top (unit) eigenvectors of the standard eigen system (left column). 
It is easy to verify that the rows of matrix $VD^{-\frac{1}{2}}$ 
are non-unit eigenvectors for the generalized eigen system for \NC/. The following  relationship 
$$ \tilde{\phi}_p \;\;=\;\; U^K \;\;\equiv\;\; [V^K D^{-\frac{1}{2}}]^{rn} $$
where operator $[\cdot]^{rn}$ normalizes matrix rows, demonstrates certain differences between 
ad hoc embeddings used by many spectral relaxation methods in their heuristic K-means discretization step and 
justified approximation embedding \eqref{eq:phi_NC} in Tab.~\ref{tab:spectral_bounds}. Note that 
our formulation scales each embedding dimension, \ie rows in matrix $V^K D^{-\frac{1}{2}}$, according to eigenvalues 
instead of normalizing these rows to unit length.

There are other common variants of embeddings for the K-means discretization step in
spectral relaxation approaches to the normalized cut. 
For example, \cite{BM_mass_spring:eccv98,malik:ijcv01,arbelaez2011contour} use 
$$\tilde{\phi}_p\;\;=\;\;[\Lambda^{-\frac{1}{2}} U]_p^K$$ for discretization of the relaxed \NC/ solution.
The motivation comes from the physics-based {\em mass-spring system} interpretation  \cite{BM_mass_spring:eccv98} 
of the generalized eigenvalue system. 

Some spectral relaxation methods motivate their discretization procedure differently. 
For example, \cite{Yu:iccv03,bach:nips03} find the closest integer solution to a {\em subspace} of equivalent 
solutions for their particular very similar relaxations of \NC/ based on the same eigen decomposition  \eqref{eq:eigen_dec1} that we used above. 
Yu and Shi \cite{Yu:iccv03} represent the subspace via matrix $$X'\;\;\equiv\;\; [\sqrt{\Lambda^m} V^m D^{-\frac{1}{2}}]^{cn}$$ 
where columns differ from our embedding $\tilde{\phi}(I_p)$ in \eqref{eq:phi_NC} only by normalization.
Theorem 1 by Bach and Jordan \cite{bach:nips03} equivalently reformulates the distance between the subspace and integer labelings
via a {\em weighted} K-means objective for embedding
\begin{equation} \label{eq:bach_emb1} 
\tilde{\phi}_p\;\;=\;\;\sqrt{\frac{1}{d_p}} V^m_p
\end{equation}
and weights $w_p=d_p$. This embedding is different from \eqref{eq:phi_NC} only by eigenvalue scaling.

Interestingly, a footnote in \cite{bach:nips03} states that \NC/ objective \eqref{eq:NC2} is equivalent to 
weighted \KM/ objective \eqref{eq:wkmixed} for exact isometry embedding
\begin{equation} \label{eq:bach_emb2} 
\phi_p \;\;=\;\; \frac{1}{d_p}G_p \;\;\;\;\;\;\;\in {\mathcal R}^{|\Omega|}
\end{equation}
based on any decomposition $A\equiv G' G$. For example, our exact isometry map \eqref{eq:phi_NC} 
for $m=|\Omega|$ and $G=\sqrt{\Lambda}VD^{\frac{1}{2}}$ is a special case. 
While \cite{bach:nips03} reduce \NC/ to K-means\footnote{ \KM/ procedure \eqref{eq:exp_kKMproc} (weighted version) 
is not practical for objective \eqref{eq:wkmixed} for points $\phi_p$ in ${\mathcal R}^{|\Omega|}$.
Instead, Dhillon et al. \cite{dhillon2004kernel} later suggested {\em pairwise} \KM/ procedure \eqref{eq:imp_kKMproc} 
(weighted version) using kernel ${\cal K}_{pq}\equiv \langle\phi_p,\phi_q\rangle$.},
their low-dimensional embedding $\tilde{\phi}$ \eqref{eq:bach_emb1}  is derived to approximate the subspace of relaxed \NC/ solutions.
In contrast, low-dimensional embedding \eqref{eq:phi_NC} approximates the  exact esometry map
$\phi$ ignoring relaxed solutions. It is not obvious if decomposition $A\equiv G' G$ for the exact embedding
\eqref{eq:bach_emb2} can be used to find any approximate lower-dimensional embeddings like \eqref{eq:phi_NC}.


\section{Other optimization ideas} \label{sec:other_optimization}

This Section outlines several extensions for optimization methods presented in 
Sec.\ref{sec:KBmain} and \ref{sec:SBmain}. For example, we vary diagonal shifts $\ds$ to
reduce Frobenius approximation error $\|\tilde{\cal K}-{\cal K}\|_F$ \eqref{eq:Fro_error}
between the optimal rank-$m$ matrix $\tilde{\cal K}$ and kernels ${\cal K}=A+\delta I$, see Sec.\ref{sec:lowdim}.
We also discuss {\em pseudo-bound} \cite{pbo:ECCV14} and {\em trust region} \cite{FTR:cvpr13} 
as alternative approximate optimization concepts that may improve
our bound optimization framework for high-order energy \eqref{eq:EA+MRF}, see Sec.\ref{sec:pbo}.

\begin{figure}
\begin{center}
\begin{tabular}{cc}
\includegraphics[width=0.45\columnwidth]{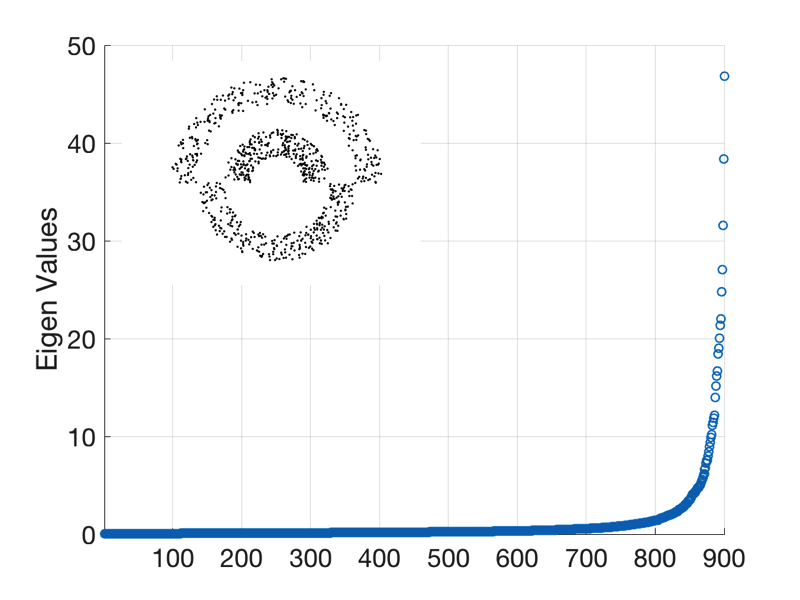} &
\includegraphics[width=0.45\columnwidth]{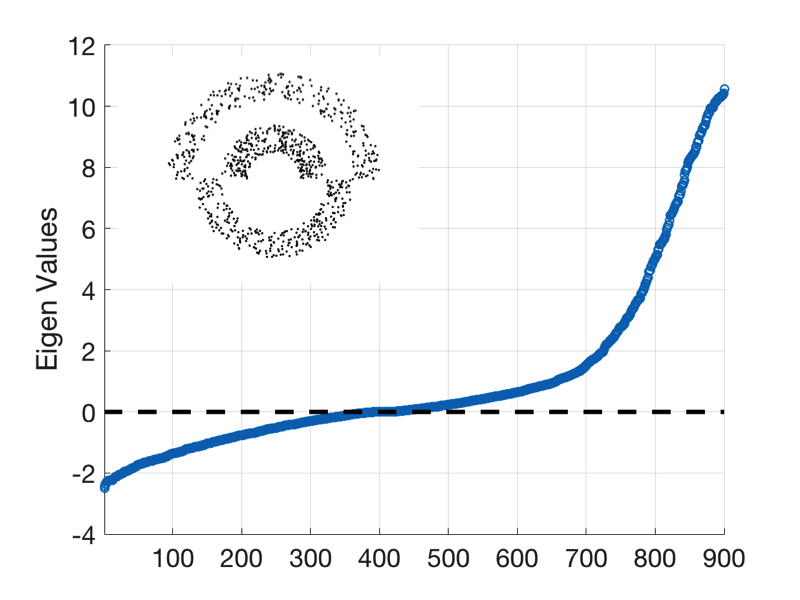}\\
(a) Gaussian for data in Fig. \ref{fig:embedding}& (b) KNN for data in Fig. \ref{fig:embedding} \\
\includegraphics[width=0.45\columnwidth]{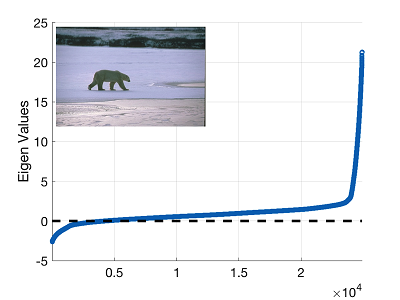} &
\includegraphics[width=0.45\columnwidth]{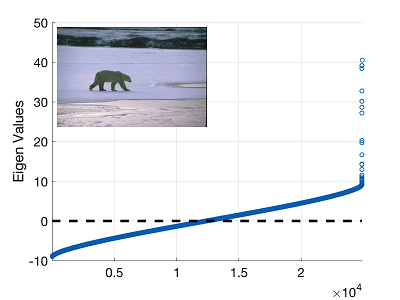}\\
(c) mPb kernel \cite{arbelaez2011contour} for image & (d) KNN kernel for image
\end{tabular}
\end{center}
\caption{Spectrum of eigenvalues of typical kernel matrices for synthetic data (top row) or real image color (bottom row). 
This helps us to select approximate embedding so as to have small approximation error \eqref{eq:Fro_error}. 
For example, with fixed width gaussian kernel in (a), it suffices to select a few top eigenvectors since 
the remaining eigenvalues are negligible. 
Note that the spectrum elevates with increasing diagonal shift $\ds$ in \eqref{eq:phi_AA}. 
In principle, we can find the optimal shift for a given number of dimensions $m$ to minimize approximation error.}
\label{fig:eigenvalues}
\end{figure}

\subsection{Extensions for spectral bound optimization} \label{sec:lowdim}

We derive low-dimensional embeddings \eqref{eq:phi_AA},\eqref{eq:phi_AC}, \eqref{eq:phi_NC} in a principled way 
allowing to estimate approximation errors with respect to the exact pairwise clustering criteria. 
Frobenius errors \eqref{eq:Fro_error} depend on embedding dimensionality $m$ and diagonal shift $\ds$ that can increase or decrease 
all eigenvalues $\lambda_i$, see Fig.\ref{fig:eigenvalues}. Unlike typical discretization heuristics in spectral relaxation methods, 
our choice of $m$ is not restricted to the fixed number of clusters $K$. 
Moreover, for each fixed $m$ we can find an optimal shift $\ds$ minimizing the sum of squared norms of omitted eigenvalues as follows. 

For simplicity we focus on \AA/ with symmetric affinity \mbox{$A=V' \Lambda V$} even though our technique can be easily extended to \AC/ and \NC/. Diagonal shift $\ds$ ensures positive semi-definiteness of the kernel matrix ${\cal K}(\ds)=\ds\mathbf I+A$ for an equivalent \kKM/. For approximate low dimensional embedding~\eqref{eq:phi_AA} it is enough to guarantee positive semi-definiteness of rank-$m$ approximating kernel $\cal \tilde K(\ds)$ 
$${\cal \tilde{K}}(\ds)\;\; =\;\; (V^m)'(\ds\mathbf I^m + \Lambda^m) V^m.$$  Thus one should use $\ds$ such that all eigenvalues $\lambda_i$ for $i\leq m$ are non-negative. Given this restriction one can choose $\ds$ to minimize Frobenius approximation error~\eqref{eq:Fro_error}:
\begin{equation}
\label{eq:optimal shift}
\min_{\ds} ||{\cal K(\ds) - \tilde K(\ds)}||_F, \;\;\; \text{\bf s.t.} \;\;\;  \ds \ge -\min_{i\leq m}\lambda_i.
\end{equation}
Making use of \eqref{eq:Fro_error}, the objective in~\eqref{eq:optimal shift} is $\sum_{i=m+1}^{|\Omega|}(\lambda_i+\ds)^2$ giving  optimum $$\ds^* \;\; = \;\; -\frac{\sum_{i=m+1}^{|\Omega|} \lambda_i}{|\Omega|-m} \;\; = \;\; -\frac{\operatorname{tr}({\cal K}(0)-{\cal \tilde K}(0))}{|\Omega|-m}$$
automatically satisfying the constraint in~\eqref{eq:optimal shift}. Indeed, $-\ds^*$ is the mean value of the discarded eigenvalues $\lambda_i$, so the shifted discarded eigenvalues $(\lambda_i+\ds^*)$ 
must contain both positive and negative values. Since $\cal \tilde K(\ds^*)$ and $\Lambda^m$ use 
the largest eigenvalues $\{\lambda_i|i\le m\}$ we have 
$$\min_{i\le m} (\lambda_i+\ds^* ) \;\;\ge\;\; \max_{i>m} (\lambda_i + \ds^*) \;\;\ge\;\;0$$  
and the constraint in \eqref{eq:optimal shift} is satisfied.
In practice the effect of the diagonal shift mainly depends on the whole spectrum of eigenvalues for a particular affinity matrix, see Figure~\ref{fig:eigenvalues}.

\subsection{Pseudo-bound and trust region methods}  \label{sec:pbo}

Our bound optimization approach to energy \eqref{eq:EA+MRF} can be extended in several ways.
For example, following {\em pseudo-bound} optimization ideas \cite{pbo:ECCV14}, parametric methods \cite{parmaxflow:iccv07} 
can be used to optimize our bounds with additional {\em perturbation terms} reducing sensitivity to local minima. 
It also makes sense to use our (spectral) bounds as an approximating functional for \eqref{eq:EA+MRF}
in the context of trust region approach to optimization \cite{FTR:cvpr13}. We also note that
parametric methods can explore all diagonal shifts $\ds$ alleviating the need for expensive eigen decomposition 
of large matrices. 

{\bf Pseudo-bound optimization:}
Consider the following definition introduced in \cite{pbo:ECCV14}.
\begin{defn} {\bf (pseudo-bound)} Given energy $E(S)$ and parameter $\pbp \in \mathcal{R}$, 
functional $B_t(S, \pbp)$ is a pseudo-bound for energy $E(S)$ at current solution $S_t$ if there exists at least one $\pbp'$ 
such that $B_t(S, \pbp')$ is an auxiliary function of $E(S)$ at $S_t$.       
\end{defn}
Instead of using an auxiliary function,  one can optimize a family of pseudo-bounds that includes at least one proper bound. 
This guarantees that original functional $E(S)$ decreases when the best solution is selected among the global minima for the whole 
family \cite{pbo:ECCV14}. In the meanwhile, such pseudo-bounds may approximate $E(S)$ better than a single auxiliary function, 
even though they come from the same class of sub-modular (globally optimizable) functionals.
The experiments of \cite{pbo:ECCV14} confirmed that pseudo-bounds  significantly outperform the optimization
quality obtained by a single auxiliary function in the context of several high-order segmentation functionals, 
e.g., entropy \cite{vicente:iccv09},  
Bhattacharyya measure \cite{BenAyed2010} and KL divergence \cite{AuxCut:cvpr13}\footnote{The segmentation 
functionals considered in \cite{pbo:ECCV14} are completely different from the pairwise clustering energies 
we consider in this work}. If the pseudo-bounds are monotone w.r.t. parameter $\pbp$, 
we can find all global minima for the whole family in polynomial time via parametric max-flow algorithm \cite{parmaxflow:iccv07}. 
This practical consideration is important when building a pseudo-bound family. For instance, \cite{pbo:ECCV14} built 
pseudo-bounds by simply adding monotone unary terms $\pbp |S^k|$ to the auxiliary functions. 
We can use a similar {\em perturbation term} in the context of auxiliary functions in Table~\ref{tab:kernel_bounds} 
and \ref{tab:spectral_bounds}.

\begin{figure}[t]
        \centering
       \begin{tabular}{cc}
       \includegraphics[width=0.18\textwidth]{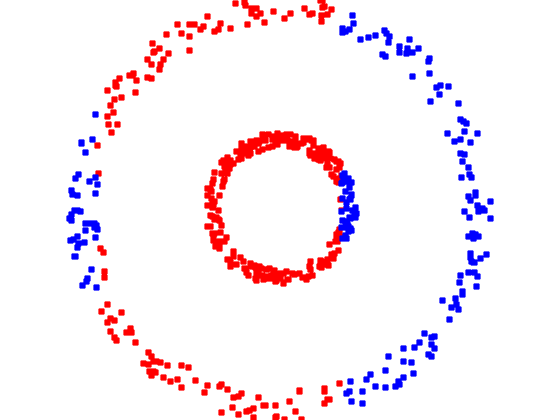} & \includegraphics[width=0.18\textwidth]{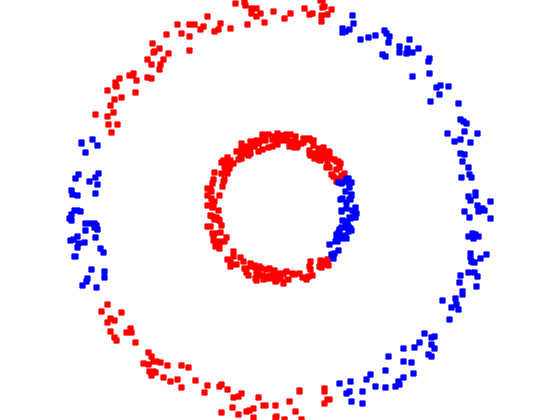} \\
     (a) Initialization  & (b) Bound optimization \\
      (energy: 0.404)  & (energy: 0.350) \\
	\\

        \includegraphics[width=0.18\textwidth]{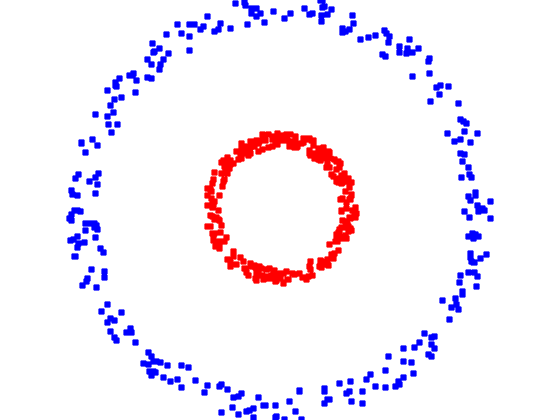} & \includegraphics[width=0.18\textwidth]{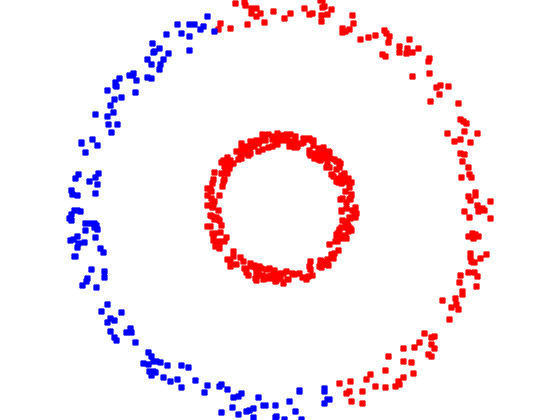} \\
       (c) Pseudo-bound opti- & (d) Spectral method \\
        mization (energy: 0.094) &  (energy: 0.115)
        \end{tabular}
        \caption{Normalized cut (\NC/) objective  \eqref{eq:NC2}. This proof-of-concept example shows that 
our {\em pseudo-bound} optimization (c) is less sensitive to local minima compared to the standard \kKM/ algorithm (bound optimizer) 
in (b). Pseudo-bound approach can also compete with spectral relaxation methods \cite{Shi2000} in (d).}
        \label{fig:pbotoy}
\end{figure}

As a proof-of-the-concept, we included Figure \ref{fig:pbotoy} demonstrating one synthetic binary clustering 
example. It uses standard \NC/ objective \eqref{eq:NC2} with p.d. Gaussian kernel (requiring no diagonal shift) 
and no additional regularization terms. Basic kernel bound optimization converges to a weak local minimum, 
while pseudo-bound optimization  over all parameters $\pbp$ in perturbation term $\pbp |S^k|$  
achieves a much better solution with lower energy. This toy example suggests that pseudo-bound approach 
may compete with standard spectral relaxation methods \cite{Shi2000}. 

A different perturbation term is motivated by using diagonal shift as a perturbation parameter $\pbp=\ds$.
Besides improving optimization, efficient search over $\ds$ may make it unnecessary to compute expensive 
eigen decompositions when estimating diagonal shift for proper p.s.d. kernels/affinities 
in the third columns of Table~\ref{tab:kernel_bounds} and \ref{tab:spectral_bounds}. 
Consider function $e(X)$ defined by symmetric affinities $A$ and weights $w$ 
$$e(X) \;\;:=\;\; -\frac{X'AX}{w'X}$$
which is similar to $\e(X)$ in \eqref{eq:e} except that $A$ is not necessarily positive definite.
Thus, $e(X)$ may not be concave as a function of relaxed continuous argument $X$, see Lemma~\ref{lm:concave}.
Pseudo-bound for clustering objectives in the general form $E_A(S)=\sum_k e(S^k)$ can be derived from diagonal shift $\ds W + A$ 
for $W:=diag(w)$ resulting in equivalent objectives.
\begin{theor}[{\bf pseudo-bound}] \label{th:pseudo_bound}
Consider pairwise clustering objectives in form $E_A(S):=\sum_k e(S^k)$. The following is a pseudo-bound of $E_A(S)$  at $S_t$
\begin{equation} \label{eq:pseudo-bound}
B_t(S,\ds) \;\;\utc\;\; \sum_k\nabla e(S^k_t)' S^k \;+\;\ds \sum_k \frac{({\bf 1}-2S^k_t)'WS^k}{w' S^k_t}
\end{equation}
becoming a proper auxiliary function for $\ds\geq -\lambda_0(W^{\frac{-1}{2}}AW^{\frac{-1}{2}})$
where $\lambda_0$ denotes the smallest eigenvalue of the matrix.
The corresponding pseudo-bound for the joint energy combining high-order clustering term $E_A$ 
with regularization terms in \eqref{eq:EA+MRF}  is
\begin{equation} \label{eq:pseudo-bound-joint}
B_t(S,\ds)   \;\;+\;\;\rp\;  \sum_{c\in\F}  E_{c} (S_{c}).
\end{equation}
\end{theor}
\begin{proof}
Implied by Lemma~\ref{lm:pseudo_bound} (Appendix D) after omitting $\ds K$, which is a constant w.r.t. $S$ not affecting
optimization.
\end{proof}

Theorem \ref{th:pseudo_bound} provides pseudo-bound \eqref{eq:pseudo-bound-joint} for joint energy \eqref{eq:EA+MRF}
combining \AA/, \AC/, \NC/ objectives in Table~\ref{tab:kernel_bounds} with regularization potentials. 
When the number of segments is $K=2$ parametric max-flow methods \cite{parmaxflow:iccv07} could be used to explore 
all distinct optimal solutions $S(\ds)$ for pseudo-bound \eqref{eq:pseudo-bound-joint} for all $\ds$. 
In particular, this search covers $\ds$ where \eqref{eq:pseudo-bound-joint} is a proper 
bound removing the need for explicitly evaluating $\lambda_0(D^{\frac{-1}{2}}AD^{\frac{-1}{2}})$ via expensive eigen decomposition.
However, in contrast to the common perturbation term $\pbp |S^k|$ \cite{pbo:ECCV14}, it is easy to check that 
the second term in $B_t(S, \ds)$ \eqref{eq:pseudo-bound} is not {\em monotonic} \cite{parmaxflow:iccv07} with respect to $\ds$. 
Thus, there is no guarantee that optimal object segments $S^k(\ds)$ form a nested sequence w.r.t. paprameter $\ds$ 
and that the number of such distinct discrete solutions is bounded by $1+|\Omega|$. 

Note that a parametric search could also be implemented for $K>2$ 
in the context of move making algorithms, see Section~\ref{sec:move-making}, using parametric max-flow \cite{parmaxflow:iccv07} to
explore all $\ds$ at each expansion or other move  \cite{BVZ:PAMI01}. 
If the number of distinct solutions for all $\ds$ gets too large at each expansion due to lack of monotonicity, one can use
simple practical heuristics. For example, restricted expansion moves can be limited to ``monotone'' subsets of pixels 
with either positive or negative unary potential  with respect to $\ds$.

Pseudo-bound  $B_t(S, \ds)$ \eqref{eq:pseudo-bound} could be useful for pairwise clustering without regularization
when monotonicity is not needed to guarantee efficiency. 
Indeed, for $K=2$ a unary potential for each pixel in \eqref{eq:pseudo-bound} changes its sign only once as parameter $\ds$ 
increases. It is enough to sort all critical values of $\ds$ changing at least one pixel in order to traverse all 
(at most $1+|\Omega|$) distinct solutions for pseudo bound \eqref{eq:pseudo-bound} in a linear time. 
For $K>2$ it is easy to check that the optimal label for
each pixel changes at most $K-1$ times as $\ds$ increases\footnote{The optimal value of a unary potential for each pixels
is the lower envelope of $K$ linear functions of $\ds$, which has at most $K-1$ breakpoints.}. 
Critical values of $\ds$ for all pixels can be sorted so that all (at most $1+(K-1)|\Omega|$) distinct solutions 
for pseudo bound \eqref{eq:pseudo-bound} can be explored in a linear time.

{\bf Trust region optimization:} Non-monotonicity of the second (perturbation) term in \eqref{eq:pseudo-bound}
follows from the factor $({\bf 1}-2S^k_t)$ assigning unary potentials of different signs to 
pixels inside and outside the current segment  $S^k_t$. This term can be represented as
$$ \ds \sum_k \frac{({\bf 1}-2S^k_t)'WS^k}{w' S^k_t}   \;\;\;\; \equiv \;\;\;\; \ds \sum_k \|S^k_t - S^k\|_w$$
where $\|S^k_t - S^k\|_w$ is a weighted Hamming distance between $S^k$ and the current segment $S^k_t$.
Thus, instead of bounds, functions \eqref{eq:pseudo-bound} and \eqref{eq:pseudo-bound-joint} can be treated as Lagrangians for 
the constrained optimization of the unary or higher-order approximations of the corresponding objectives 
over {\em trust region}  \cite{FTR:cvpr13}
$$ \|S^k_t - S^k\|_w \;\;\leq\;\; \tau.$$
In this case, parameter $\ds$ is a Lagrange multiplier that can be adaptively  changed from iteration to iteration \cite{FTR:cvpr13},
rather than exhaustively explored at each iteration. Note that it may still be useful to add a common monotone perturbation term
as in \cite{pbo:ECCV14} $$\pbp|S^k| \;\;\equiv\;\;\pbp {\bf 1}' S^k $$
that can be efficiently explored at each iteration. This corresponds to a combination of pseudo-bound and trust region
techniques.


\section{Parzen Analysis \& Bandwidth Selection} \label{sec:analysis}

This section discusses connections of \kKM/ clustering to Parzen densities providing probabilistic 
interpretations for many equivalent pairwise clustering objectives discussed in our work. 
In particular, this section gives insights on practical selection of kernels or their bandwidth. 
We discuss extreme cases and analyze adaptive strategies. For simplicity,
we mainly focus on Gaussian kernels, even though the analysis applies to 
other types of positive normalized kernels.

Note that standard Parzen density estimate for the distribution of data points within segment $S^k$ 
can be  expressed using normalized Gaussian kernels \cite{Bishop06,Girolami2002}
\begin{equation} \label{eq:Parzen_estimate}
\P_{\sigma} (I_p|S^k) \;\;=\;\; \frac{\sum_{q \in S^k}k(I_p,I_q)}{|S^k|} .
\end{equation}  
It is easy to see that \kKM/ energy \eqref{eq:kKmeans3} is exactly the following high-order Parzen density energy
\begin{equation} \label{eq:Parzen_energy}
F(S) \;\utc\; -\sum_k \sum_{p \in S^k} \P_{\sigma} (I_p|S^k).
\end{equation} 

This probabilistic interpretation of \kKM/ gives an additional point of view for comparing
it with  \pKM/ clustering with log-likelihood energy \eqref{eq:pKmeans}. 
Instead of parametric ML models \kKM/ uses Parzen density estimates. Another difference is
absence of the $\log$ in \eqref{eq:Parzen_energy}. Omitting the $\log$ reduces the weight
of low probability points, that is, outliers. In contrast, log-likelihoods in \eqref{eq:pKmeans}
are highly sensitive to outliers. To address this problem, \pKM/ methods often use heuristics like mixing 
the desired probability model $\P$ with a uniform distribution, \eg $\tilde{\P}(\cdot|\theta) := \epsilon+(1-\epsilon)\P(\cdot|\theta)$.

\subsection{Extreme bandwidth cases} \label{sec:extreme}

Parzen energy  \eqref{eq:Parzen_energy} is also useful for analyzing two extreme cases of kernel bandwidth:
large kernels approaching the data range and small kernels approaching the data resolution.
This section analyses these two extreme cases.

{\bf Large bandwidth and basic K-means:} 
Consider Gaussian kernels of large bandwidth $\sigma$
approaching the data range. In this case Gaussian kernels $k$ in \eqref{eq:Parzen_estimate}
can be approximated (up to a scalar) by Taylor expansion $1-\frac{\|I_p-I_q\|^2}{2\sigma^2}$.
Then, Parzen density energy \eqref{eq:Parzen_energy} becomes (up to a constant) 
$$  \sum_k \frac{\sum_{pq \in S^k} \|I_p-I_q\|^2}{2\sigma^2 |S^k|}$$
which is proportional to the pairwise formulation for the basic K-means or {\em variance criteria} 
in Tab.\ref{tab:kmeans} with Euclidean metric $\|\|$. 
That is, \kKM/ for large bandwidth Gaussian kernels reduces to the basic K-means 
in the original data space instead of the high-dimensional embedding space.

In particular, this proves that as the bandwidth gets too large \kKM/ looses its ability to
find non-linear separation of the clusters. This also emphasizes the well-known bias of basic K-means 
to equal size clusters \cite{UAI:97,volbias:ICCV15}.

{\bf Small bandwidth and Gini criterion:}
Very different properties could be shown for the opposite extreme case of small bandwidth
approaching data resolution. It is easy to approximate Parzen formulation of \kKM/ energy \eqref{eq:Parzen_energy} as
\begin{equation} \label{eq:parzen_criterion}
F(S) \;\approxutc \; - \sum_k |S^k| \cdot \langle \P_{\sigma}(S^k), \dSk \rangle \; 
\end{equation}
where $\P_{\sigma}(S^k)$ is kernel-based density \eqref{eq:Parzen_estimate} and $\dSk$ is
a ``true'' continuous density for the sample of intensities $\{I_p|p\in S^k\}$ in segment $S^k$. 
Approximation \eqref{eq:parzen_criterion} follows directly from the same Monte-Carlo estimation argument 
in Appendix C with the only difference being 
$f =  - \P_\sigma (S^k)$  instead of $ - \log \P (\theta_{\S})$. 

If kernels have small bandwidth
optimal for accurate Parzen density estimation\footnote{Bandwidth near inter-point distances avoids density oversmoothing.} 
we get $\P_{\sigma}(S^k) \approx \dSk$ further reducing \eqref{eq:parzen_criterion} to approximation
\begin{equation*} 
\approxutc \;\;\;\; - \sum_k  |S^k|\cdot \langle \dSk, \dSk \rangle 
\end{equation*}
that proves the following property.
\begin{proper}
\label{property:gini_approx}
Assume small bandwidth Gaussian kernels optimal for accurate Parzen density estimation. Then
kernel K-means energy \eqref{eq:Parzen_energy} can be approximated by the standard {\em Gini criterion}  
for clustering \cite{breiman1996}:
\begin{equation} \label{eq:gini_criterion}
E_G(S)\;:= \;\; \sum_k|S^k|\cdot G(S^k) 
\end{equation}
where $G(S^k)$ is the {\em Gini impurity} for the data points in $S^k$
\begin{equation}
\label{eq:gini_impurity}
G(S^k)\;:=\;1- \langle \dSk, \dSk \rangle \;\equiv\;\; 1-\int_x \dSk (x)^2 dx.
\end{equation} 
\end{proper}

Similarly to entropy, {\em Gini impurity} $G(S^k)$  can be viewed as a measure of sparsity or ``peakedness''
for continuous or discrete distributions. Both Gini and entropy clustering criteria 
are widely used for decision trees \cite{breiman1996,Louppe2013}.  
In this discrete context Breiman \cite{breiman1996} analyzed theoretical 
\begin{wrapfigure}{l}{32mm}
\includegraphics[width=30mm]{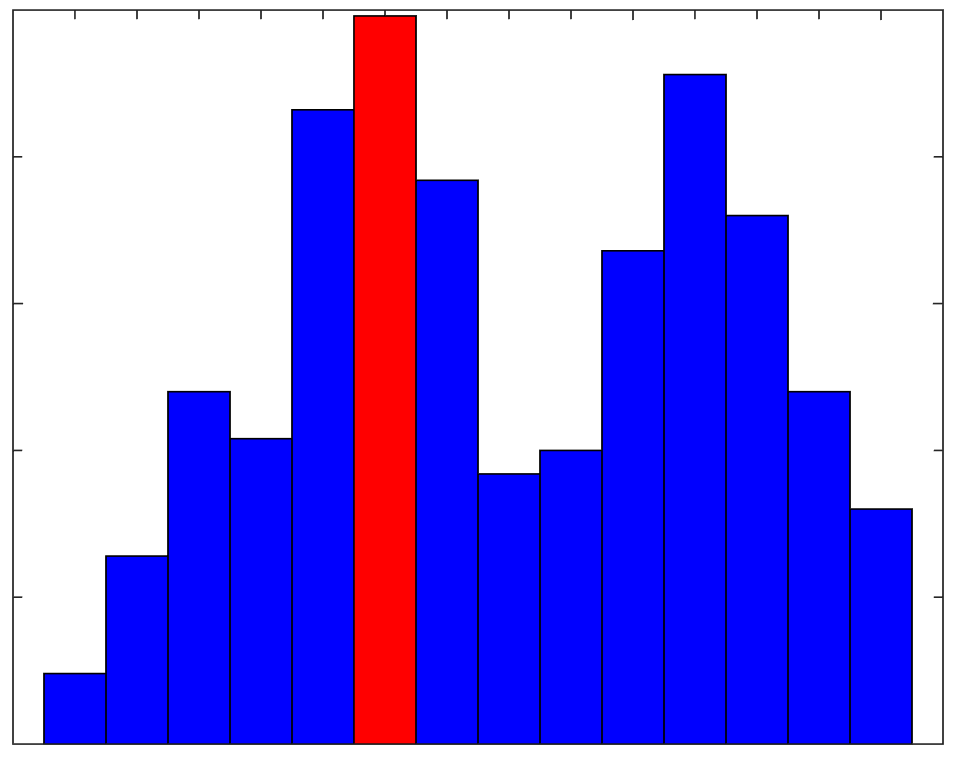}
\end{wrapfigure} 
properties of Gini criterion \eqref{eq:gini_criterion}
for the case of histograms $\P_h$ where $G(S^k)=1-\sum_x \P_h (x|S^k)^2$. He proved that  for $K=2$ the minimum of the Gini criterion 
is achieved by sending all data points within the highest-probability bin to one cluster and the remaining data points 
to the other cluster,  see the color encoded illustration above. We extend Brieman's result to the continuous
Gini criterion \eqref{eq:gini_criterion}-\eqref{eq:gini_impurity}.
\begin{theor} [{\bf Gini Bias}, $K=2$] \label{th:gini_bias}
Let $d_{\Omega}$  be a continuous probability density function over domain $\Omega\subseteq{\mathcal R}^n$
defining conditional density $\dSk(x):=d_{\Omega}(x|x\in S^k)$ for any non-empty subset $S^k\subset\Omega$.
Then, continuous version of Gini clustering criterion \eqref{eq:gini_criterion} achieves its optimal value at the partitioning of 
$\Omega$ into regions $S^1$ and $S^0=\Omega\setminus S^1$ such that $$S^1=\arg\max_x d_{\Omega}(x).$$
\end{theor}
\begin{proof}
See Appendix E and Proposition \ref{prop:appendixD}.
\end{proof}

The bias to small dense clusters is practically noticeable for small bandwidth kernels, see Fig.\ref{fig:nash}(d). 
Similar empirical bias to tight clusters was also observed in the context of average association \eqref{eq:kKmeans3} in \cite{Shi2000}. 
As kernel gets wider the continuous Parzen density \eqref{eq:Parzen_estimate} no longer approximates the true distribution
$\dS$ and Gini criterion \eqref{eq:gini_criterion} is no longer valid as an approximation for \kKM/ energy \eqref{eq:Parzen_energy}. 
In pratice, {\em Gini bias} gradually disappears as bandwidth gets wider.
This also agrees with the observations for wider kernel in average association \cite{Shi2000}.
As discussed earlier, in the opposite extreme case when bandwidth get very large (approaching data range) 
\kKM/ converges to basic K-means or {\em variance criterion}, which has very different properties. 
Thus, kernel K-means properties strongly depend on the bandwidth.

The extreme cases for kernel K-means, \ie Gini and variance criteria, are useful to know when selecting kernels. 
Variance criteria for clustering has bias to equal cardinality segments \cite{UAI:97,volbias:ICCV15}. 
In contrast, Gini criteria has bias to small dense clusters (Theorem \ref{th:gini_bias}). 
To avoid these biases kernel K-means should use kernels of width that is neither too small nor too large. 
Our experiments compare different strategies with fixed and adaptive-width kernels (Sec.\ref{sec:Nash}). 
Equivalence of kernel-K-means to many standard clustering criteria such as {\em average distortion}, 
{\em average association}, {\em normalized cuts}, \etc (see Sec.\ref{sec:kKM}) also suggest kernel selection strategies
based on practices in the prior literature.

\subsection{Adaptive kernels via Nash embedding and \KNN/} \label{sec:Nash}

\begin{figure}[t!]
\centering
\begin{tabular}{ cc }
{\small space of points $I$}  & {\small transformed points  $I'$}   \\
{\small with Riemannian metric}  & {\small with Euclidian metric} \\
\includegraphics[width=0.45\columnwidth]{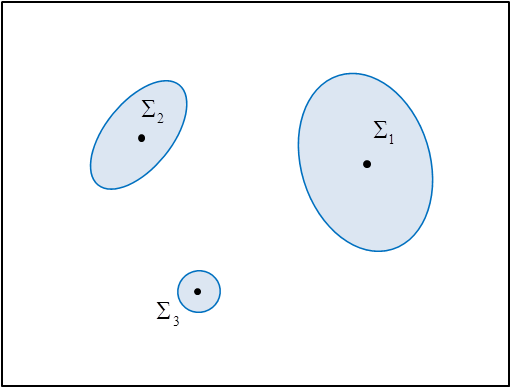} &
\includegraphics[width=0.45\columnwidth]{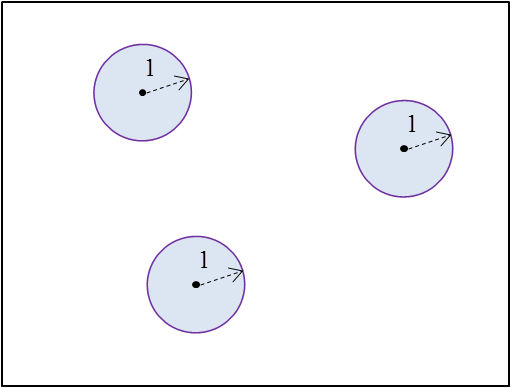}  \\[-1ex]
{\tiny unit balls in Riemannian metric defined by tensor $\Sigma$} &
{\tiny unit balls in Euclidean metric} 
\end{tabular}
\centering
 \caption{{\em Nash embedding:} adaptive non-normalized Gaussian kernels define
isometric transformation of the color space modifying density. Ellipsoids are mapped to balls. \label{fig:ne}}
\end{figure}

As discussed in Sec.\ref{sec:extreme}, kernel width should neither be too small nor too large.   
We propose adaptive kernels designed to equalize the density in highly dense regions in the color space. 
The following equation interprets adaptive Gaussian kernels via
Riemannian distances in the color space (left picture in Fig.\ref{fig:ne})
\begin{eqnarray*}
k_{p}(I_p,I_q) & = & e^{\frac{-\|I_p-I_q\|^2}{2\sigma_{p}^2}}  \\ &=& e^{\frac{-(I_p-I_q)'\Sigma^{-1}_{p}(I_p-I_q)}{2}}.
\end{eqnarray*}
According to Nash embedding theorem \cite{Nash1956}, this Riemannian color space can be
isometrically embedded into a Euclidean space, so that the last expression above is equal to
\begin{equation*}
\hspace{12ex} \;\; = \;\;  e^{\frac{-\|I'_p-I'_q\|^2}{2}} \;\; = \;\;  k(I'_p,I'_q)
\end{equation*}
where $k$ is a fixed-width Gaussian kernel in the new transformed space (right picture in Fig.\ref{fig:ne}).
Thus, non-normalized Gaussian kernels of adaptive width $\sigma_{p}$ (or covariance matrix $\Sigma_{p}$, 
in general) define some color space transformation, {\em Nash embedding}, that locally stretches or shrinks the space.  
After this transformation, clustering is done using a fixed (homogeneous) Gaussian kernel of constant width.

Figure \ref{fig:ne} helps to illustrate how Nash embedding changes the color space density.
The number of points in a unit (Euclidean) ball neighborhood in the transformed space is equal to the number of
points in the corresponding unit (Riemannian) ball in the original space: 
$$K = d'\cdot V_1 = d\cdot V_{\sigma}$$
where $d$ and $d'$ are local densities in the original and transformed spaces. Thus, kernel width $\sigma_{p}$ 
can be selected adaptively based on any desired transformation of density $d'(d)$ according to formula 
\begin{equation} \label{eq:adaptive_sigma}
\sigma_{p} \;\; \sim \;\;\; \sqrt[n]{\frac{d'(d_p)}{d_p}}
\end{equation}
where $d_p := d(I_p)$ is an observed local density for points in the color space.
This local density can be evaluated using any common estimator, \eg Parzen approach gives 
\begin{equation} \label{eq:dens}
d(I_p)\;\;\sim\;\; \sum_q\;\; \frac{1}{\Delta_q^n}\cdot e^{\frac{-\|I_p-I_q\|^2}{2\Delta_q^2}}
\end{equation}
where $\Delta_q$ could be adaptive or fixed $\Delta_q=const$, according to any standard technique 
for density estimation \cite{Bishop06}.
\begin{figure}[t]
\centering
        \includegraphics[width=\linewidth]{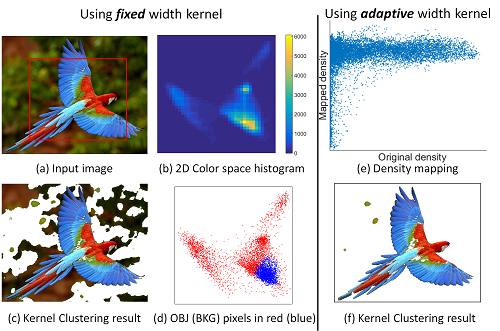}
\centering
 \caption{(a)-(d): Breiman bias for fixed (small) kernel. (e) Empirical density transform
for adaptive kernels \eqref{eq:adaptive_sigma} with $d'(d)=const$. Color space density equalization 
counters Breiman bias. (f) Segmentation for adaptive kernels.}
\label{fig:nash}
\end{figure}

To address Breiman bias one can use density equalizing transforms $d'(d)=const$ or 
$d'=\frac{1}{\alpha}\log(1 + \alpha d)$, which even up the highly dense parts of the color space. 
Formula \eqref{eq:adaptive_sigma} works for any target transform $d'(d)$. 
Once adaptive kernels $\sigma_p$ are chosen, Nash theory also allows 
to obtain empirical scatter plots $d'(d)_{\Omega} := \{ (d'(I'_p),d(I_p))|p\in\Omega\}$, for example, to
compare it with the selected ``theoretical'' plot  $d'(d)$. 
Estimates $d(I_p)$ are provided in \eqref{eq:dens} and the density estimate for Nash embedding are
\begin{equation} \label{eq:dprime}
d'(I'_p)\;\;\sim\;\;  \sum_q e^{\frac{-\|I'_p-I'_q\|^2}{2}} = \sum_q e^{\frac{-\|I_p-I_q\|^2}{2\sigma_{q}^2}}.
\end{equation}
Note the difference between empirical density estimates for $d'$ in \eqref{eq:dprime}
and $d$ in \eqref{eq:dens}: the former uses the sum of {\em non-normalized} kernels of selected 
adaptive width $\sigma_q$ in \eqref{eq:adaptive_sigma} and the latter is the sum of {\em normalized}
kernels of width $\Delta_q$ based on chosen density estimator. While parameter $\sigma_q$ directly controls
the density transformation, $\Delta$ plays a fairly minor role concerning the quality of 
estimating density $d$. 

Figure \ref{fig:nash}(e) illustrates the empirical density mapping $d'(d)_{\Omega}$ induced by 
adaptive kernels \eqref{eq:adaptive_sigma}  for $d'(d)=const$. 
Notice a density-equalization effect within high density areas in the color space addressing the Breiman bias. 

The const density mapping can be approximated using \KNN/ graph. To be specific, the symmetric \KNN/ kernel in this paper is defined as follows:
\begin{eqnarray}
k_{pq} \;\;=\;\; k(f_p,f_q) &=&[f_p \in \KNN/(f_q)] \nonumber \\ &+& [f_q \in \KNN/(f_p)]
\label{eq:knndefinition}
\end{eqnarray}
where $\KNN/(f_p)$ is a set of K nearest neighbors of $f_p$. The affinity between $f_p$ and $f_q$ achieves maximum value of 2 if they are mutually each other's \KNN/s.


\section{Experiments}
\label{sec:experiments}

This section is divided into two parts. The first part (Sec.\ref{exp:ncmeetmrf}) shows the benefits of extra MRF regularization for kernel \& spectral clustering, e.g. normalized cut. We consider pairwise Potts, label cost and robust bin consistency term, as discussed in
Sec.\ref{sec:MRF_overview}. We compare to spectral clustering \cite{Shi2000,malik:ijcv01} and kernel K-means \cite{dhillon2004kernel}, which can be seen as degenerated versions for spectral and kernel cuts (respectively) without MRF terms. 
We show that MRF helps kernel \& spectral clustering in segmentation and image clustering.
In the second part (Sec.\ref{exp:mrfmeetnc}) we replace the log-likelihoods 
in model-fitting methods, \eg GrabCut \cite{GrabCuts:SIGGRAPH04}, by pairwise clustering term, e.g. \AA/ and \NC/.
This is particularly advantageous for high dimension features (location, depth, motion).

\textbf{Implementation details:} For segmentation, our kernel cut uses \KNN/ kernel \eqref{eq:knndefinition} for pixel features $I_p$, which can be concatenation of LAB (color), XY (location) and M (motion or optical flow) \cite{Bro11a}. We choose 400 neighbors and {randomly} sample 50 neighbors for each pixel. Sampling does not degrade our segmentation but expedites bound evaluation.  We also experiment with popular \textit{mPb} contour based affinities \cite{arbelaez2011contour} for segmentation. The window radius is set to 5 pixels. 

For contrast-sensitive regularization, we use standard penalty $w_{pq}=\frac{1}{d_{pq}}e^{-0.5\|I_p-I_q\|_2^2/\eta}$ for \eqref{eq:potts}, where $\eta$ is 
the average of $\|I_p-I_q\|^2$ over a 8-connected neighborhood and $d_{pq}$ is the distance between pixels $p$ and $q$ in the image plane. We set $w_{pq}=\frac{1}{d_{pq}}$ for length regularization.

For GrabCut, we used histogram-based probability model, as is common in the literature \cite{vicente:iccv09,victor:eccv08}. We tried various bin size for spatial and depth channels. 

With fixed width Gaussian kernel used in Fig. \ref{fig:nash}, the time complexity of the naive implementation 
of kernel bound evaluation in \eqref{Kernel-bound} is $O(|\Omega|^2)$. The bottleneck is the evaluation of $\mathcal{K}{X_t}$ and ${X_t}'\mathcal{K}X_t$ in derivative $\nabla \e(X_t) $ \eqref{eq:ne}. In this case, we resort to fast approximate dense filtering method in \cite{paris2006fast}, which takes $O(|\Omega|)$ time. Also notice that the time complexity of the approach in \cite{paris2006fast} grows exponentially with data dimension $N$. A better approach for high-dimensional dense filtering is proposed in \cite{adams2010fast}, which is of time $O(|\Omega|\times N)$. We stick to \cite{paris2006fast} for low-dimensional color space in experiments in Fig. \ref{fig:nash}.


\begin{figure*}[t]
\centering
\begin{tabular} {c@{\hspace{0.4ex}}c@{\hspace{0.4ex}}c@{\hspace{0.4ex}}c@{\hspace{0.4ex}}c@{\hspace{0.4ex}}c}
\includegraphics[width=0.33\columnwidth]{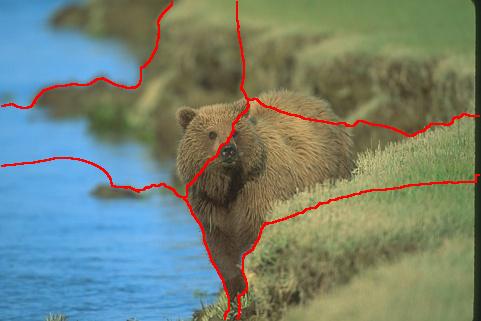} &
\includegraphics[width=0.33\columnwidth]{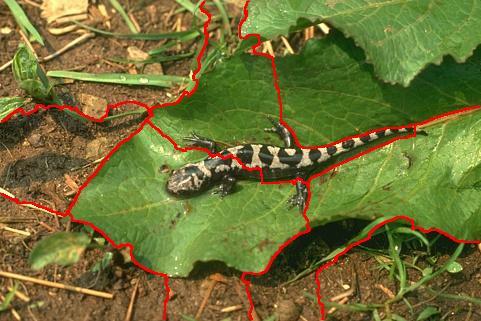} &
\includegraphics[width=0.33\columnwidth]{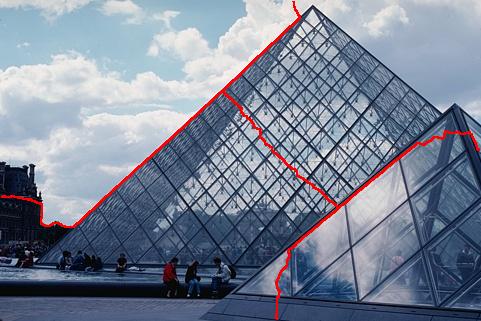} &
\includegraphics[width=0.33\columnwidth]{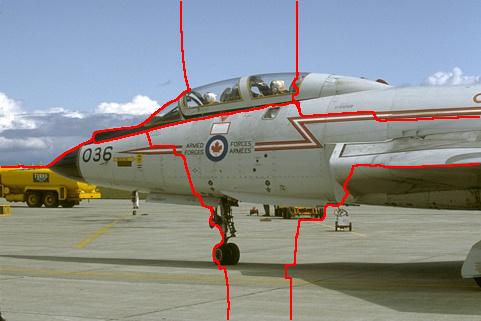} &
\includegraphics[width=0.33\columnwidth]{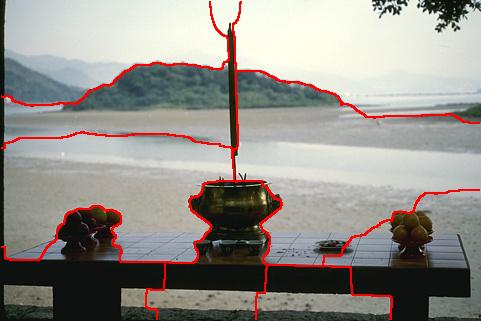} &
\includegraphics[width=0.33\columnwidth]{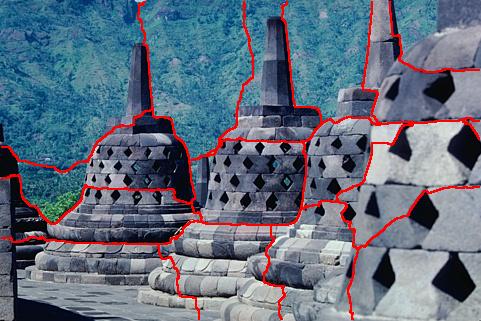} \\[-.05ex]
\includegraphics[width=0.33\columnwidth]{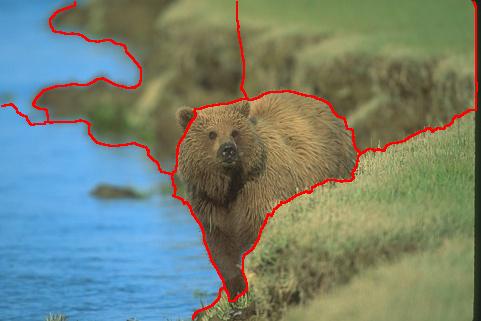}&
\includegraphics[width=0.33\columnwidth]{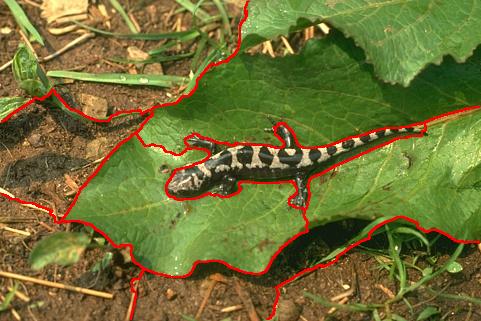}&
\includegraphics[width=0.33\columnwidth]{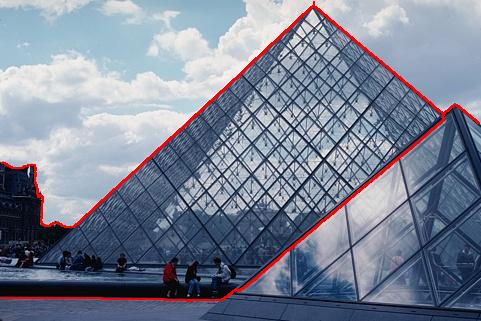}&
\includegraphics[width=0.33\columnwidth]{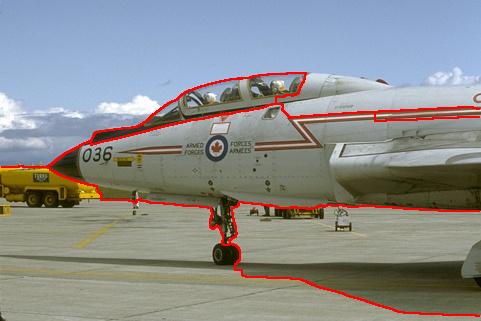}&
\includegraphics[width=0.33\columnwidth]{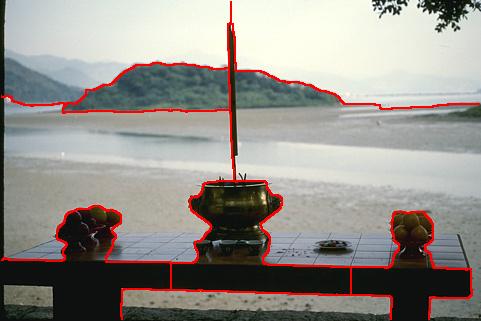}&
\includegraphics[width=0.33\columnwidth]{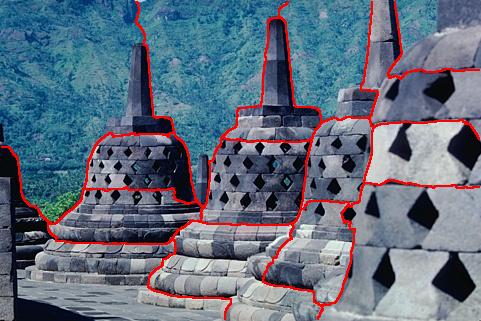} \\[-.05ex]
\includegraphics[width=0.33\columnwidth]{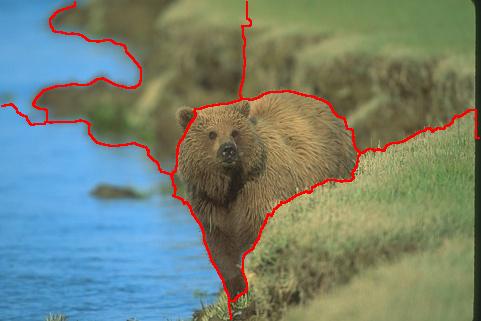}&
\includegraphics[width=0.33\columnwidth]{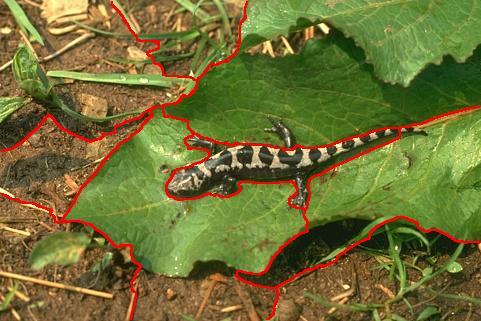}&
\includegraphics[width=0.33\columnwidth]{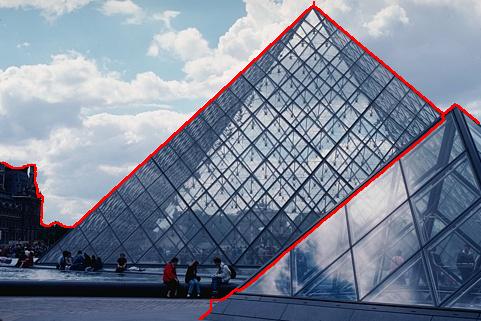}&
\includegraphics[width=0.33\columnwidth]{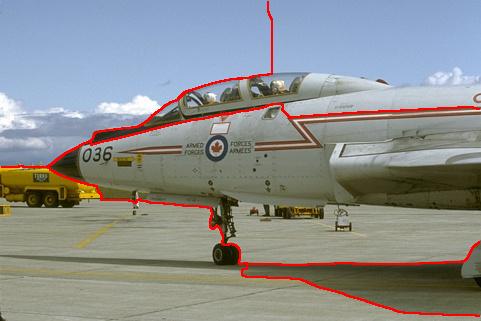}&
\includegraphics[width=0.33\columnwidth]{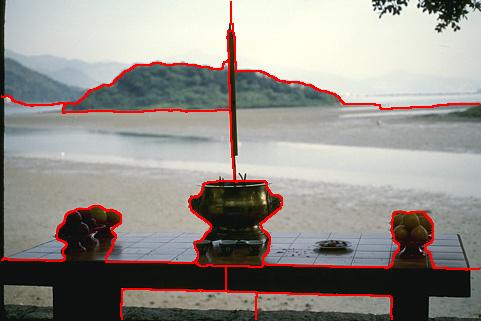}&
\includegraphics[width=0.33\columnwidth]{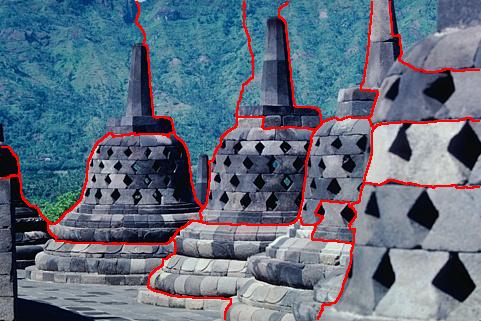}
\end{tabular}
\caption{Sample results on BSDS500. Top row: spectral clustering. Middle \& Bottom rows: our Kernel \& Spectral Cuts.}
\label{fig:scvskc}
\end{figure*}

\subsection{MRF helps Kernel \& Spectral Clustering} \label{exp:ncmeetmrf}

Here we add MRF regulation terms to typical normalized cut applications, 
such as unsupervised multi-label segmentation
\cite{arbelaez2011contour} and image clustering \cite{collins2014spectral}. Our kernel and spectral cuts are 
used to optimize the joint energy of normalized cut and MRF \eqref{eq:EA+MRF} or \eqref{eq:wapproxEA+MRF}.

\subsubsection{Normalized Cut with Potts Regularization} 

Spectral clustering \cite{Shi2000} typically solves a (generalized) eigen problem, followed by simple clustering method such as K-means on the eigenvectors. However, it is known that such paradigm results in undesirable segmentation in large uniform regions \cite{arbelaez2011contour,malik:ijcv01}, see examples in Fig. \ref{fig:scvskc}. Obviously such edge mis-alignment can be penalized by contrast-sensitive Potts term. Our spectral and kernel cuts get 
better segmentation boundaries. As is in \cite{dhillon2004kernel} we use spectral initialization.

Tab.\ref{tab:scvskc} gives quantitative results on BSDS500 dataset. 
Number of segments in ground truth is provided to each method. 
It shows that kernel and spectral cuts give better covering, PRI (probabilistic rand index) and VOI (variation of information) than spectral clustering. Fig.\ref{fig:scvskc} gives sample results. Kernel K-means \cite{dhillon2004kernel} gives results 
similar to spectral clustering and hence are not shown.

\subsubsection{Normalized Cuts with Label Cost \cite{LabelCosts:IJCV12}}

\begin{figure}[b]
\begin{subfigure}[t]{1\columnwidth}
\centering
\includegraphics[width=0.31\columnwidth]{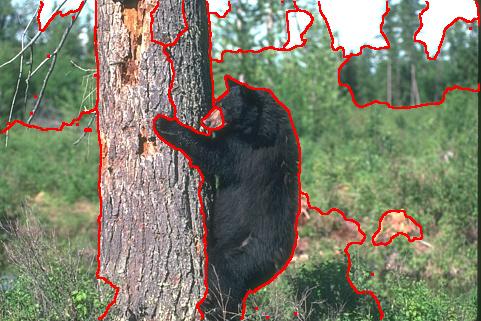}\;
\includegraphics[width=0.31\columnwidth]{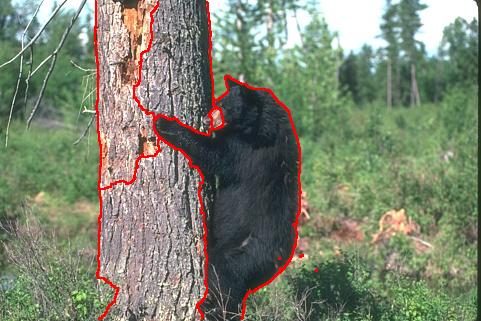}\;
\includegraphics[width=0.31\columnwidth]{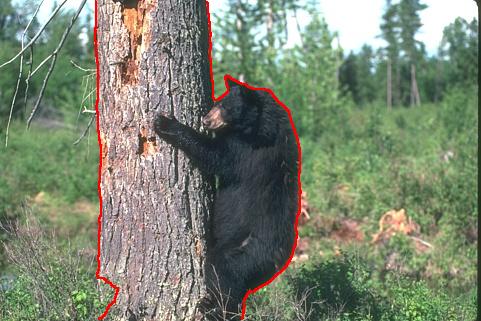}
\end{subfigure}\\
\begin{subfigure}[t]{1\columnwidth}
\centering
\includegraphics[width=0.31\columnwidth]{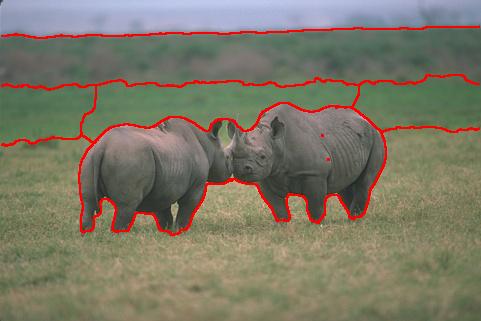}\;
\includegraphics[width=0.31\columnwidth]{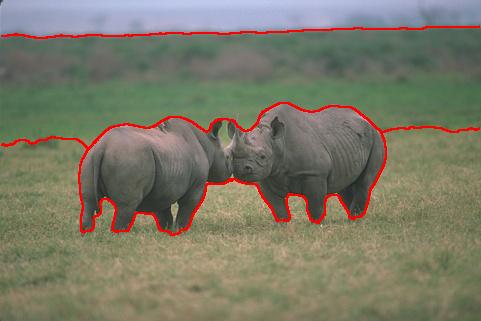}\;
\includegraphics[width=0.31\columnwidth]{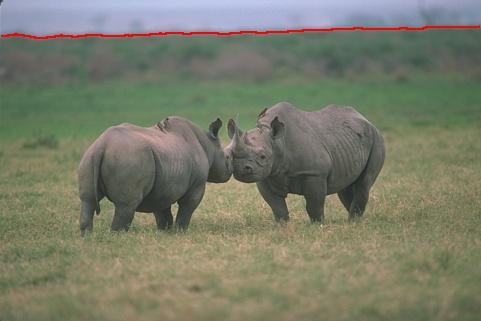}
\end{subfigure}\\
\begin{subfigure}[t]{1\columnwidth}
\centering
\includegraphics[width=0.31\columnwidth]{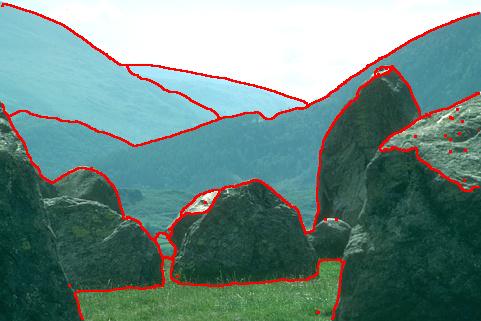}\;
\includegraphics[width=0.31\columnwidth]{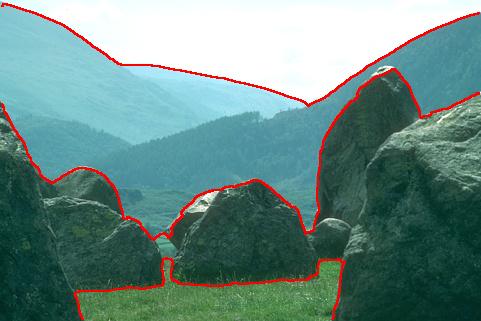}\;
\includegraphics[width=0.31\columnwidth]{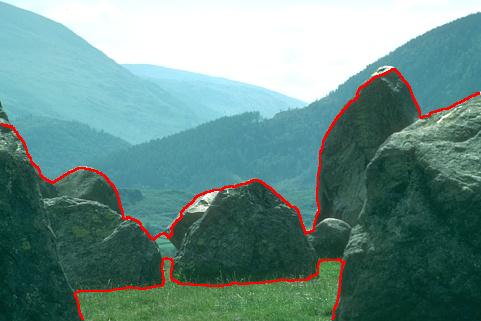}
\end{subfigure}
\caption{Segmentation using our kernel cut with label cost. We experiment with increasing value of label cost $h_k$ for each label (from left to right)}
\label{fig:labelcost}
\end{figure}

Unlike spectral clustering, our kernel and spectral cuts do not need the number of segments beforehand. 
We use kernel cut to optimize a combination of the normalized cut, Potts model and label costs terms. The label cost \eqref{eq:labelcost} 
penalizes each label by constant $h_k$. The energy is minimized by $\alpha$-expansion and $\alpha\beta$-swap moves in 
Sec.\ref{sec:move-making}. We sample initial models from patches, as in~\cite{LabelCosts:IJCV12}. Results with different label cost are 
shown in Fig.\ref{fig:labelcost}. Due to sparsity prior, our kernel and spectral cuts automatically prune \textit{weak} models and determine the number of segments, yet yield regularized segmentation. We use \KNN/ affinity for normalized cut and mPb \cite{arbelaez2011contour}  based Potts regularization.

\begin{table}
\begin{center}
\begin{tabular}{ |c|c|c|c|} 
 \hline
 method& Covering & PRI & VOI  \\ \hline
 Spectral Clustering & 0.34 & 0.76& 2.76 \\ \hline
 Our Kernel Cut & {0.41} & \textbf{0.78}& {2.44} \\ \hline
 Our Spectral Cut & \textbf{0.42} & \textbf{0.78}& \textbf{2.34} \\ 
 \hline
\end{tabular}
\end{center}
 \caption{Results of spectral clustering (K-means on eigenvectors) and our Kernel Cut \& Spectral Cuts on BSDS500 dataset. For this experiment mPb-based kernel is used \cite{arbelaez2011contour}.}
 \label{tab:scvskc}
\end{table}

\subsubsection{Normalized Cut with High Order Consistency Term \cite{kohli2009robust,Park:ECCV2012,onecut:iccv13}} It is common that images come with multiple tags, such as those in Flickr platform or the LabelMe dataset \cite{oliva2001modeling}. We study how to utilize tag-based group prior for image clustering \cite{collins2014spectral}.

We experiment on the LabelMe dataset \cite{oliva2001modeling} which contains 2,600 images of 8 scene categories (coast, mountain, forest, open country, street, inside city, tall buildings and highways). We use the same GIST feature, affinity matrix and group prior as used in \cite{collins2014spectral}. We found the group prior to be noisy. The dominant category in each group occupies only 60\%-90\% of the group. The high-order consistency term is defined on each group. For each group, we introduce an energy term that is akin to the \textit{robust} $P^n$-Potts \cite{kohli2009robust}, which can be exactly minimized within a single $\alpha\beta$-swap or $\alpha$-expansion move. Notice that here we have to use robust consistency potential instead of rigid ones.

Our kernel cut minimizes NC  plus the \textit{robust} $P^n$-Potts term. Spectral cut minimizes energy of \eqref{eq:approxEA+MRF}. Normalized mutual information (NMI) is used as the measure of clustering quality. Perfect clustering with respect to ground truth has NMI value of 1.

\begin{figure}[b]
\centering
\includegraphics[width=0.5\columnwidth]{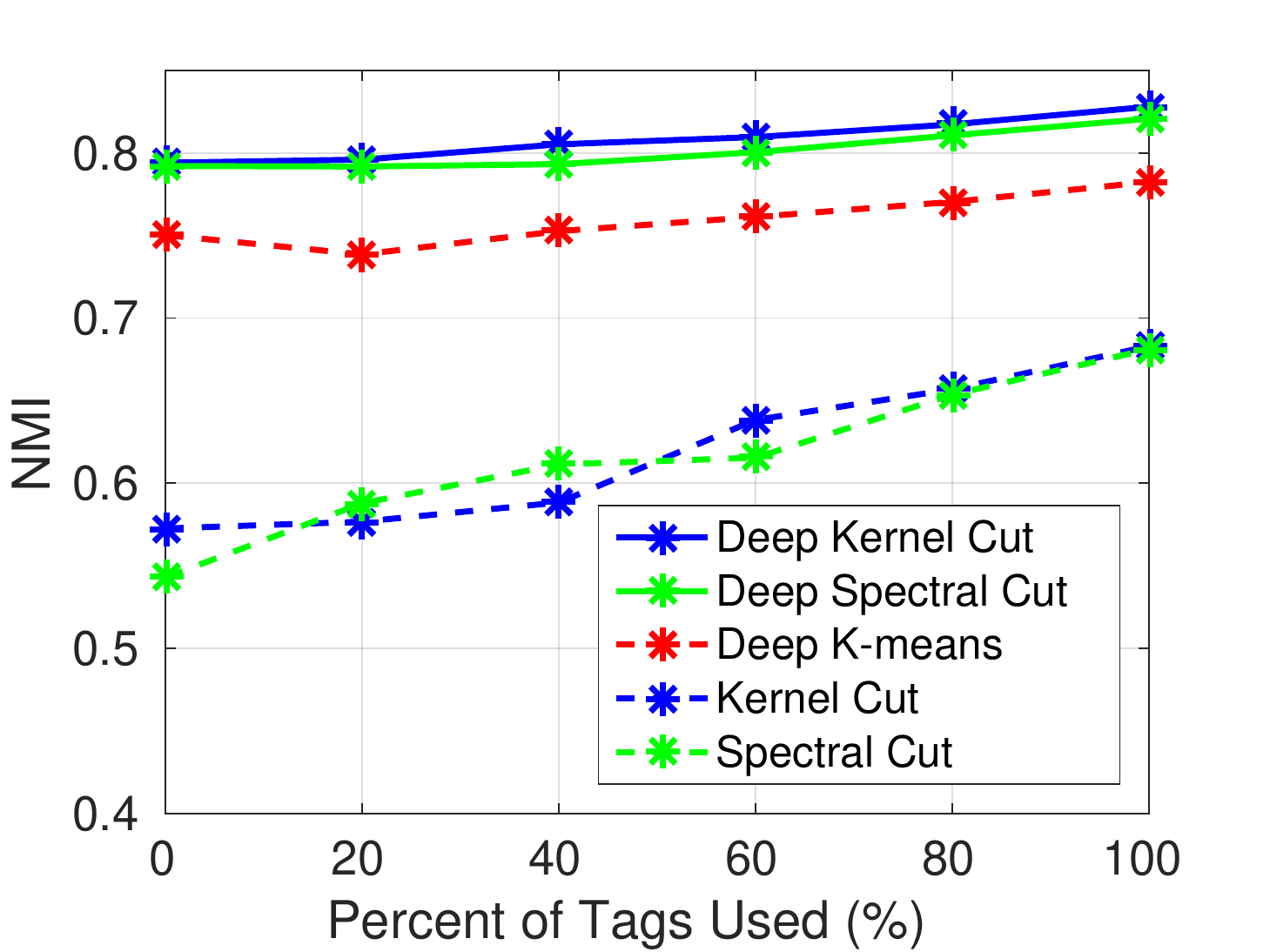}\includegraphics[width=0.5\columnwidth]{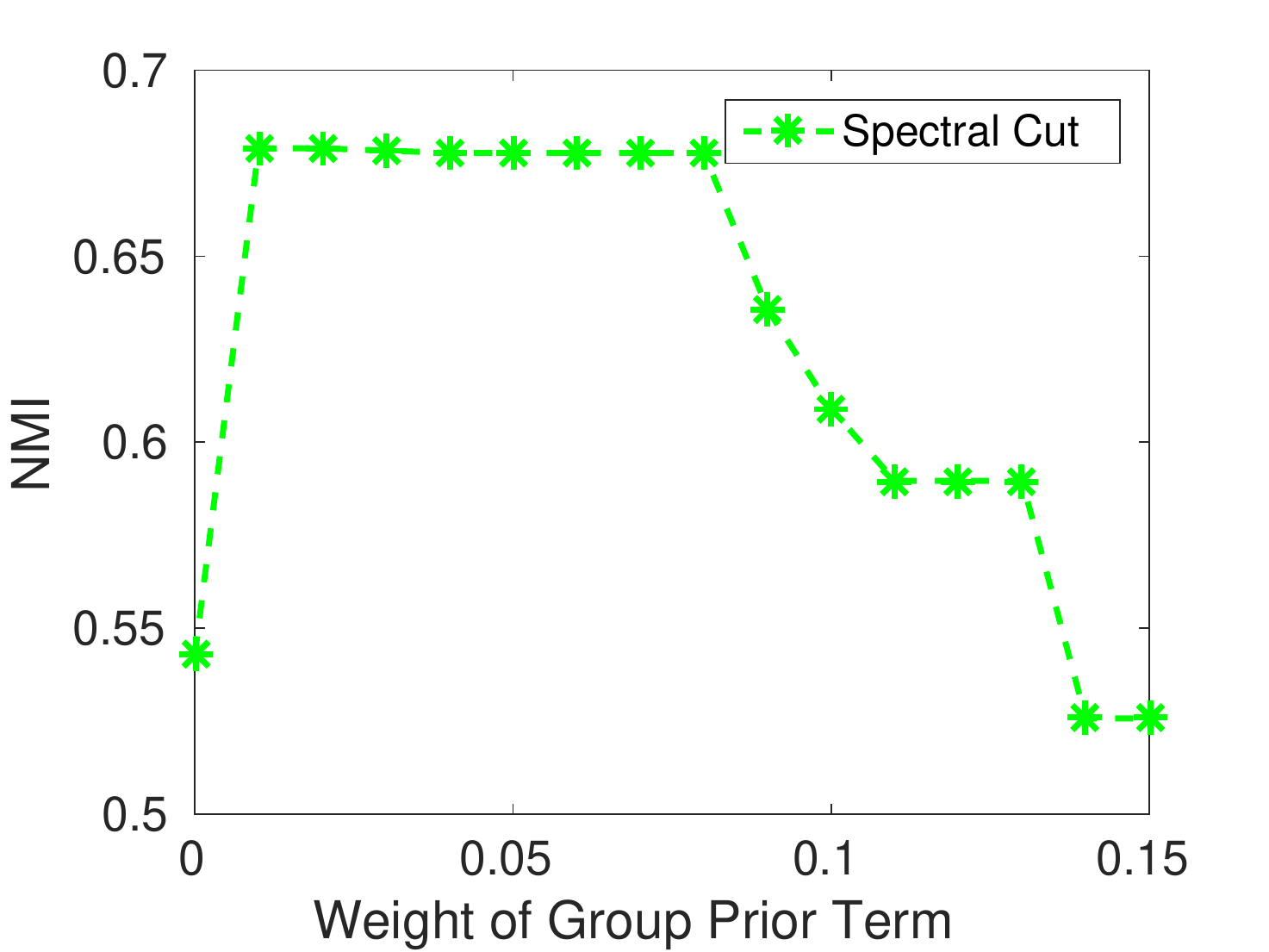}
\caption{Incorporating group prior achieves better NMI for image clustering. Here we use tags-based group prior. Our method achieved better NMI when more images are tagged. The right plot shows how the weight of bin consistency term affects our method.}
\label{fig:groupprior}
\end{figure}

Spectral clustering and kernel K-means \cite{dhillon2004kernel} give NMI value of 0.542 and 0.572 respectively. Our kernel cut and spectral cut significantly boost the NMI to  0.683 and 0.681. Fig.\ref{fig:groupprior} shows the results with respect to different amount of image tags used. The left most points correspond to the case when no group prior is given. We optimize over the weight of high order consistency term, see Fig.\ref{fig:groupprior}. Note that it's not the case the larger the weight the better since the grouping prior is noisy.

We also utilize deep features, which are 4096 dimensional fc7 layer from AlexNet \cite{krizhevsky2012imagenet}. We either run plain K-means, or construct a \KNN/ kernel on deep features. These algorithms are denoted as deep K-means, deep spectral cut or deep kernel cut in Fig. \ref{fig:groupprior}. Incorporating group prior indeed improved clustering. The best NMI of 0.83 is achieved by our kernel cut and spectral cut for KNN kernel on deep features.

\subsection{Kernel \& Spectral Clustering helps MRF} \label{exp:mrfmeetnc}

In typical MRF applications we replace the log-likelihood terms by average association or normalized cut. We evaluate our Kernel Cut (fixed width kernel or \KNN/) in the context of interactive segmentation, and compare with the commonly used GrabCut algorithm \cite{GrabCuts:SIGGRAPH04}. In Sec. \ref{sec:robustness}, we show that our kernel cut is less sensitive to choice of regularization weight $\rp$. We further report results on the GrabCut dataset of 50 images and the Berkeley dataset in Sec. \ref{sec:datasetresult}. We experiment with both (i) contrast-sensitive edge regularization, (ii) length regularization and (iii) color clustering (i.e., no regularization) so as to assess to what extent the algorithms benefit from regularization.


From Sec. \ref{sec:expxy} to Sec. \ref{sec:expmotion}, we also report segmentation results of our kernel cut with high-dimensional features $I_p$, including location, texture, depth, and motion respectively.



\subsubsection{Robustness to regularization weight}
\label{sec:robustness}

\begin{figure}[h]
\centering
\begin{tabular}{  c  }
        \includegraphics[width=\columnwidth]{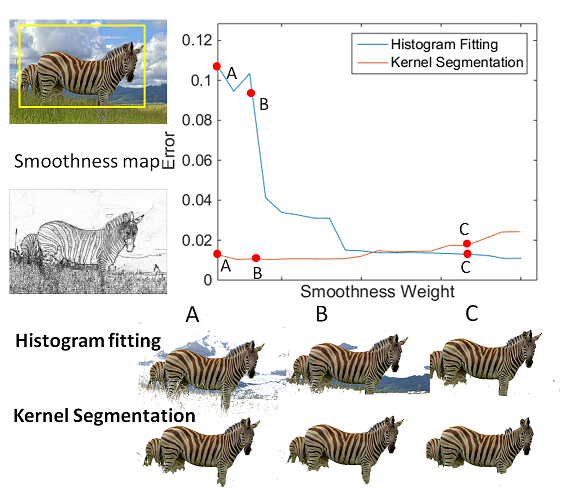}
\\ (a) \\
\includegraphics[width=\columnwidth]{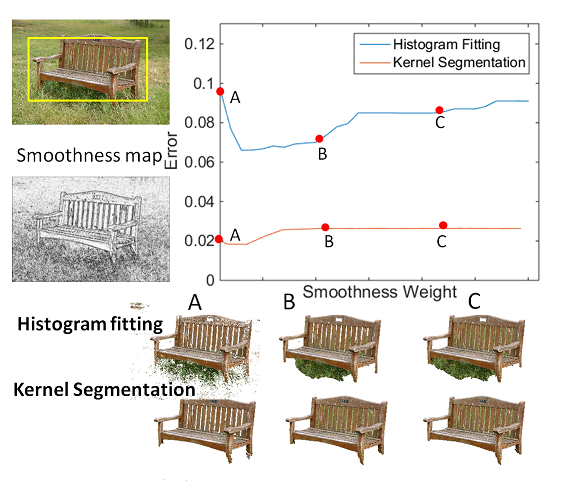} \\
(b)
\end{tabular}
\centering
 \caption{Illustration of robustness to smoothness weight.}
\label{fig:chair-zebra}
\end{figure}

We first run all algorithms without smoothness. Then, we experiment with several values of $\rp$ for the contrast-sensitive edge term.  In the experiments of Fig. \ref{fig:chair-zebra} (a) and (b), we used the yellow boxes as initialization. For a clear interpretation of the results, we did not use any additional hard constraint. In Fig. \ref{fig:chair-zebra}, ''KernelCut-KNN-AA'' means Kernel Cut with KNN kernel for average association (\AA/). Without smoothness, our Kernel Cut yields much better results than Grab Cut. Regularization significantly benefited the latter, as the decreasing blue curve in (a) indicates. For instance, in the case of the zebra image, model fitting yields a plausible segmentation when assisted with a strong regularization. However, in the presence of noisy edges and clutter, as is the case of the chair image in (b), regularization does not help as much. Note that for small regularization weights $\rp$
our method is substantially better than model fitting. Also, the performance of our method is less dependent on regularization weight and does not require fine tuning of $\rp$. 

\subsubsection{Segmentation on GrabCut \& Berkeley datasets.}
\label{sec:datasetresult}
First, we report results on the GrabCut database (50 images) using the bounding boxes provided 
in \cite{lempitsky2009box}. For each image the error is the percentage of mis-labeled pixels. 
We compute the average error over the dataset. 
\begin{figure}[t]
        \centering
        \includegraphics[width=0.8\linewidth]{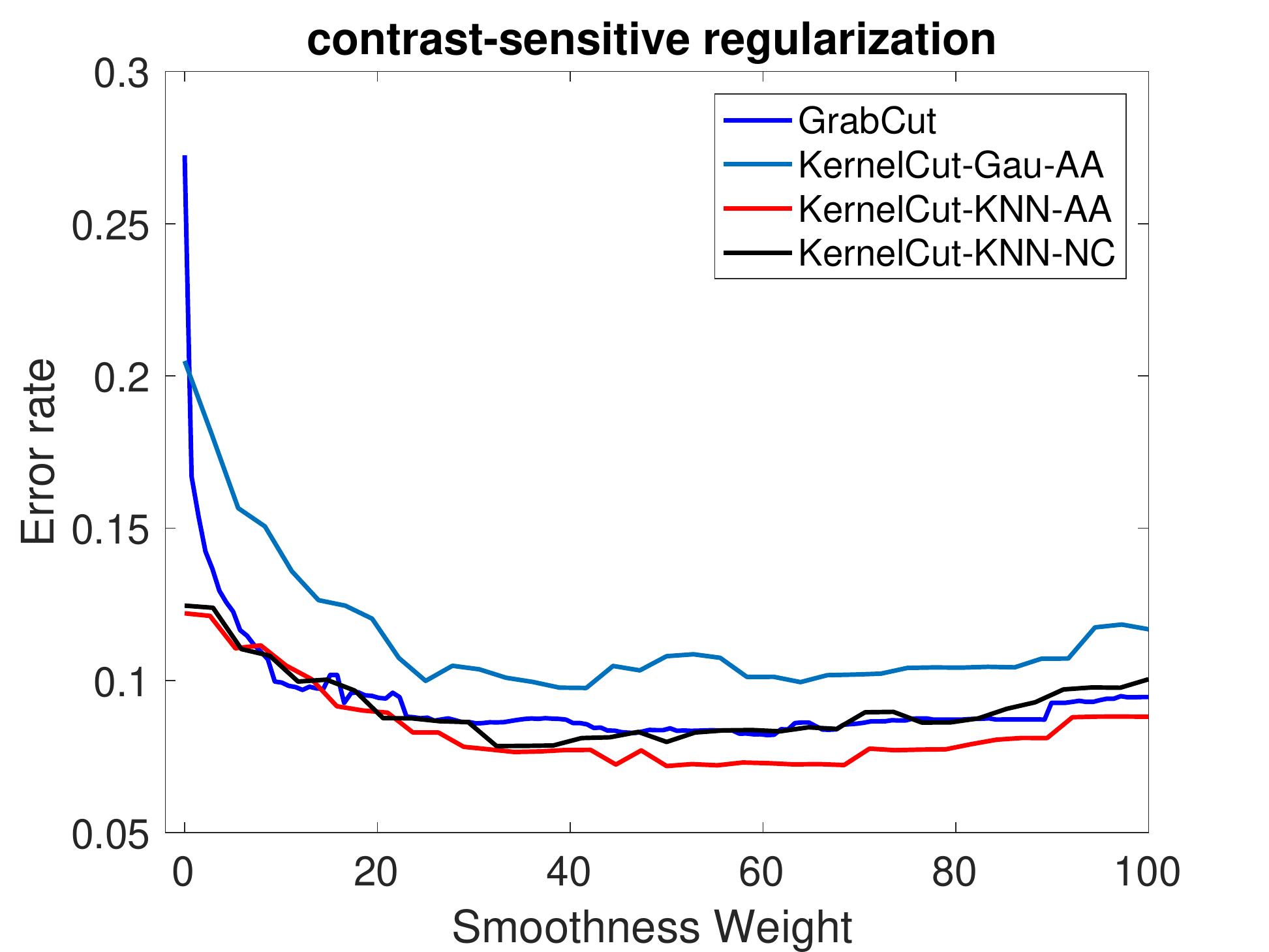}
        \includegraphics[width=0.8\linewidth]{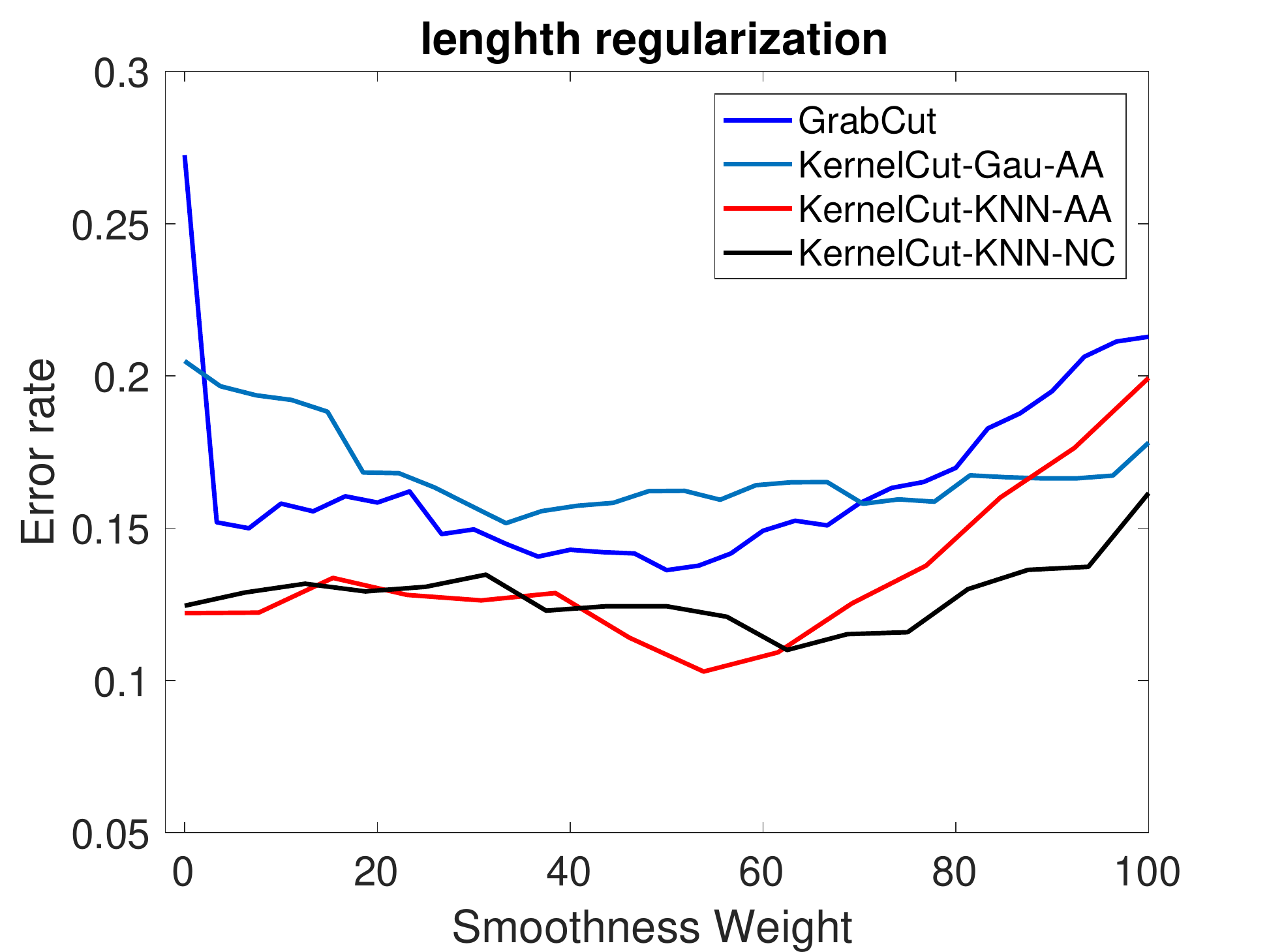}
                \caption{Average error vs. regularization weights for different variants of our KernelCut on the GrabCut dataset.}
        \label{fig:grabcuterrors}
\end{figure}

\begin{figure}[t]
        \centering
        \includegraphics[width=0.8\linewidth]{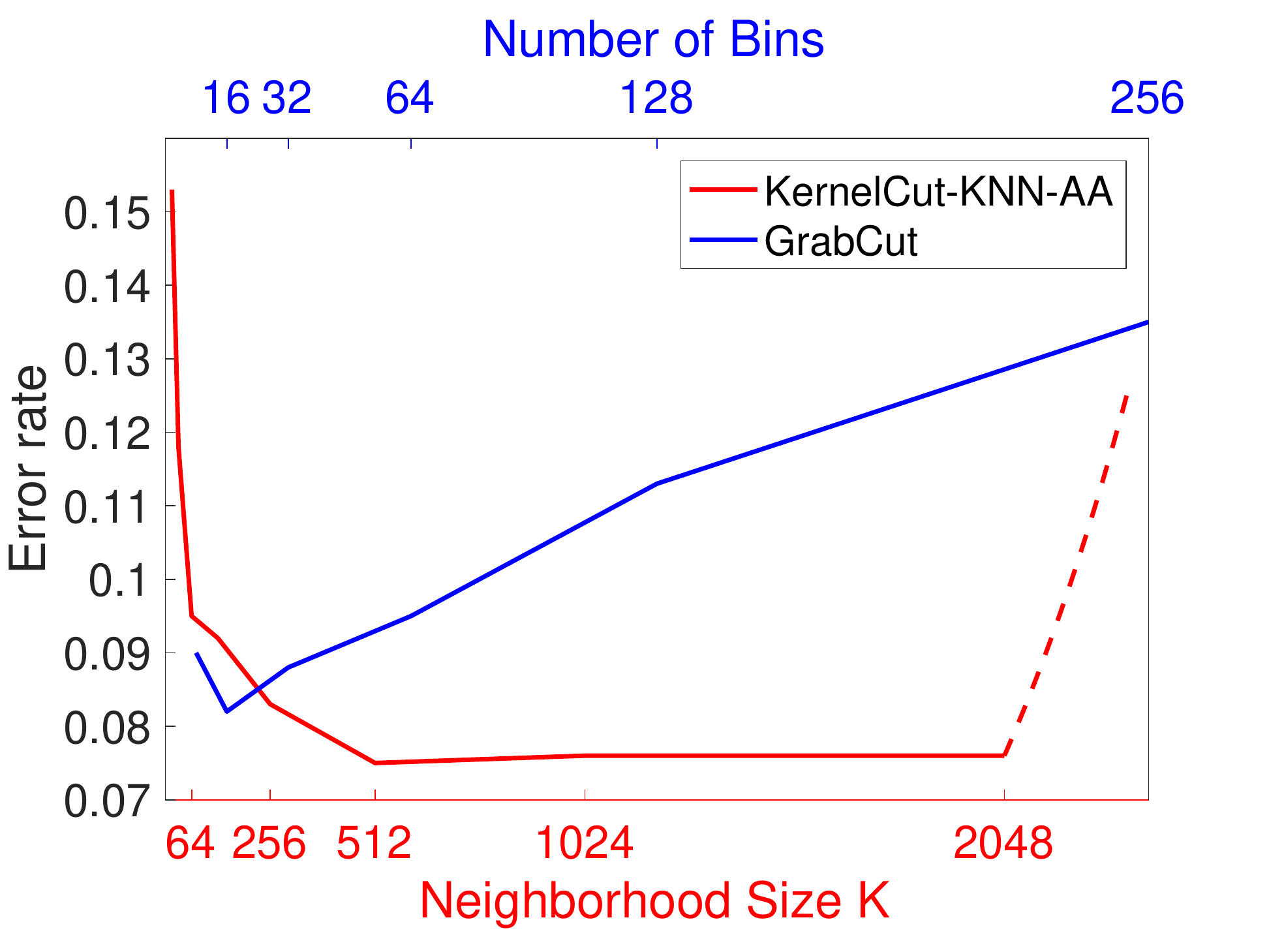}
                \caption{Our method aKKM is robust to choice of $K$ while GrabCut is sensitive to bin size for histograms.}
        \label{fig:kexperiments}
\end{figure}

We experiment with four variants of our Kernel Cut, depending on whether to use fixed width Gaussian kernel or KNN kernel, and also the choice of normalized cut or average association term. We test different smoothness weights and plot the error curves\footnote{The smoothness weights for different energies are not directly comparable; Fig. \ref{fig:grabcuterrors} shows all the curves for better visualization.} in Fig.\ref{fig:grabcuterrors}. 
Table \ref{tb:errorrates} reports the best error for each method. For contrast-sensitive regularization GrabCut gets good results ($8.2\%$). However, without edges (Euclidean or no regularization) GrabCut gives much higher errors
($13.6\%$ and $27.2\%$). In contrast, KernelCut-KNN-AA (Kernel Cut with adaptive KNN kernel for \AA/) gets only $12.2\%$ doing a better job in color clustering without 
any help from the edges. In case of contrast-sensitive regularization, our method outperformed GrabCut ($7.1\%$ vs. $8.2\%$)
but both methods benefit from strong edges in the GrabCut dataset. Fig .\ref{fig:kexperiments} shows that our Kernel Cut is also robust to the hyper-parameter, i.e. $K$ for nearest neighbours, unlike GrabCut.
\begin{table}[t]
{\footnotesize
\hfill{}
 \begin{tabular}{ c | cccc}  \hline
{\bf boundary} & \multicolumn{4}{c}{{\bf color clustering term}} \\ 
{\bf smoothness}     &GrabCut& \begin{tabular}[c]{@{}c@{}}KernelCut\\-Gau-AA\end{tabular} & \begin{tabular}[c]{@{}c@{}}KernelCut\\-Gau-NC\end{tabular}  & \begin{tabular}[c]{@{}c@{}}KernelCut\\-KNN-AA\end{tabular}  \\ \hline
none                        & 27.2 & 20.4  & 17.6 & \textbf{12.2} \\ \hline
Euclidean length   & 13.6 & 15.1  & 16.0 & \textbf{10.2}\\ \hline
contrast-sensitive &   8.2 &   9.7  & 13.8 &  \textbf{7.1}\\  \hline
  \end{tabular}
\hfill{}}
\caption{Box-based interactive segmentation (Fig.\ref{fig:grabcutexamples}). 
Error rates (\%) are averaged over 50 images in GrabCut dataset. KernelCut-Gau-NC means KernelCut for fixed width Gaussian kernel based normalized cut objective.}
\label{tb:errorrates}
\end{table}
\begin{figure}
        \centering
        \includegraphics[width=\columnwidth]{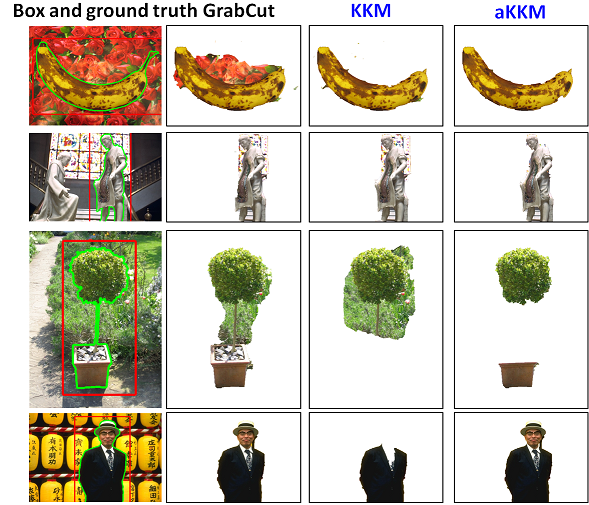}
                \caption{Sample results for GrabCut and our kernel cut with 
             fixed width Gaussian or adaptive width KNN kernel, see Tab.\ref{tb:errorrates}.}
        \label{fig:grabcutexamples}
\end{figure}

Figure \ref{fig:grabcutexamples} gives some sample results. The top row shows a failure case for GrabCut 
where the solution aligns with strong edges. The second row shows a challenging image where our KernelCut-KNN-AA works well. The third and fourth rows show failure cases for fixed-width Gaussian kernel in kernel cut due to 
Breiman's bias (Th.\ref{th:gini_bias}) where image segments of uniform color are separated; see green bush and black suit. Adaptive kernel (KNN) addresses this bias. 

\begin{figure}
\centering
\includegraphics[width=\columnwidth]{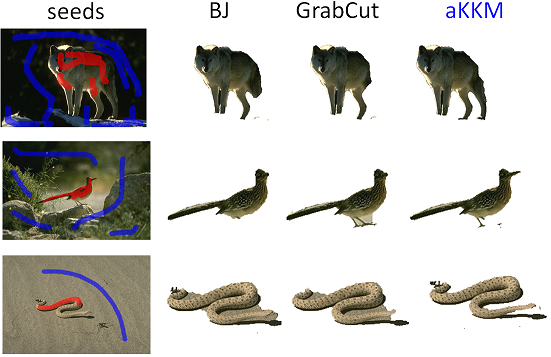}
                \caption{Sample results for BJ \cite{BJ:01}, 
                GrabCut \cite{GrabCuts:SIGGRAPH04}, and our kernel cut for adaptive KNN kernel, 
                see Tab.\ref{tb:errorrates_seeds}.}
        \label{fig:seedsexamples}
\end{figure}

We also tested seeds-based segmentation on a different database \cite{mcguinness:2010} with ground truth, 
see Tab.\ref{tb:errorrates_seeds} and Fig.\ref{fig:seedsexamples}.

\begin{table}[t]
{\small
\hfill{}
 \begin{tabular}{ l | ccc}  \hline
\multirow{2}{*}{\bf{boundary smoothness}} & \multicolumn{3}{c}{{\bf color clustering term}} \\ 
        & BJ & GrabCut &  \begin{tabular}[c]{@{}c@{}}KernelCut\\-KNN-AA\end{tabular}   \\ \hline
none      & 12.4 & 12.4 & \textbf{7.6} \\ \hline
contrast-sensitive & 3.2   & 3.7  & \textbf{2.8}\\  \hline
  \end{tabular}
\hfill{}}
\caption{Seeds-based interactive segmentation (Fig.\ref{fig:seedsexamples}). Error rates (\%) are averaged over 82 images from Berkeley database. Methods get the same seeds entered by four users. 
We removed 18 images with multiple nearly-identical objects (see Fig.\ref{fig:dbsamples}) 
from 100 image subset in \cite{mcguinness:2010}. (GrabCut and KernelCut-KNN-AA give 3.8 and 3.0 errors on the whole database.)}
\label{tb:errorrates_seeds}
\end{table}

\subsubsection{Segmentation of similar appearance objects}
\label{sec:expxy}
%
%

\begin{figure}[t]
\begin{center}
   \setlength{\fboxsep}{0pt}
\setlength{\fboxrule}{1pt}
\includegraphics[width=0.22\columnwidth]{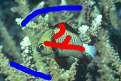}
  \includegraphics[width=0.22\columnwidth]{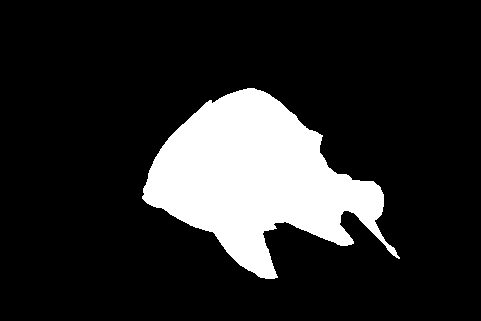}
   \fbox{\includegraphics[width=0.22\columnwidth]{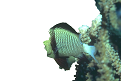}}
   \fbox{\includegraphics[width=0.22\columnwidth]{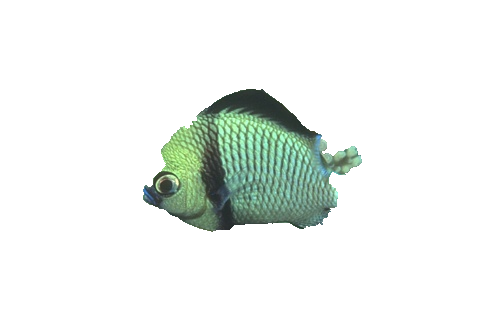}}
   \\
   \includegraphics[width=0.22\columnwidth]{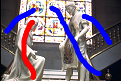}
  \includegraphics[width=0.22\columnwidth]{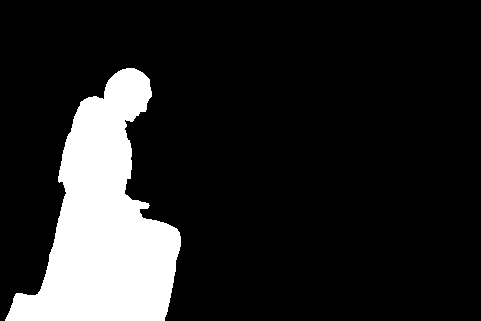}
   \fbox{\includegraphics[width=0.22\columnwidth]{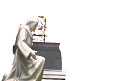}}
   \fbox{\includegraphics[width=0.22\columnwidth]{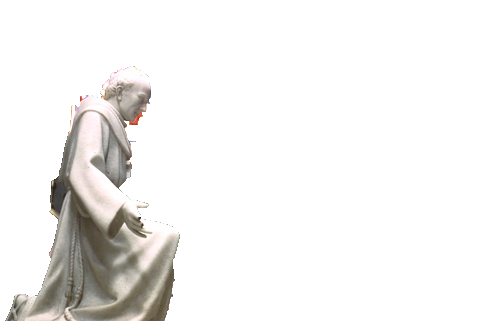}}
   \\~~~~(a) seeds\;\; (b) ground truth (c) GrabCut \;(d)Kernel Cut
\end{center}
   \caption{Sample results using RGBXY+XY.}
\label{fig:dbsamples}
\end{figure}

Even though objects may have similar appearances or look similar to the background 
(\eg the top row in Fig.\ref{fig:dbsamples}), we assume that the objects of interest are 
compact and have different locations. This assumption motivates using XY 
coordinates of pixels as extra features for distinguishing similar or camouflaged objects. XY features 
have also been used in \cite{Nieuwenhuis:PAMI13} to build space-variant color distribution. 
However, such distribution used in MRF-MAP 
inference \cite{Nieuwenhuis:PAMI13} would still over-fit the data \cite{kkm:iccv15}.
Let $I_p\in \mathcal{R}^5$ be the augmented color-location features 
$I_p=[l_p,a_p, b_p, \rs x_p, \rs y_p]$ at pixel $p$
where $[l_p, a_p, b_p]$ is its color, $[x_p,y_p]$ are its image coordinates, and $\rs$ is a scaling parameter. 
Note that the edge-based Potts model~\cite{BJ:01} also uses the XY information. 
Location features in the clustering and regularization terms have complementary effect: the former
solves appearance camouflage while the latter gets edge alignment.

\begin{figure}[t]
\begin{center}
   \includegraphics[width=0.4\linewidth]{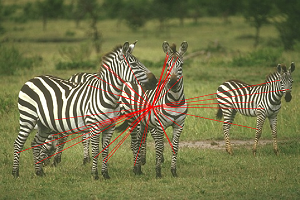}\quad\includegraphics[width=0.4\linewidth]{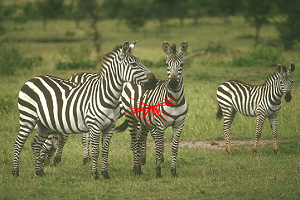}
\end{center}
   \caption{Visualization of a pixel's K-Nearest-Neighbours for RGB feature (left) or RGBXY feature (right).}
\label{fig:knngraph}
\end{figure}

\begin{figure}[h]
\begin{center}
   \includegraphics[width=0.65\linewidth]{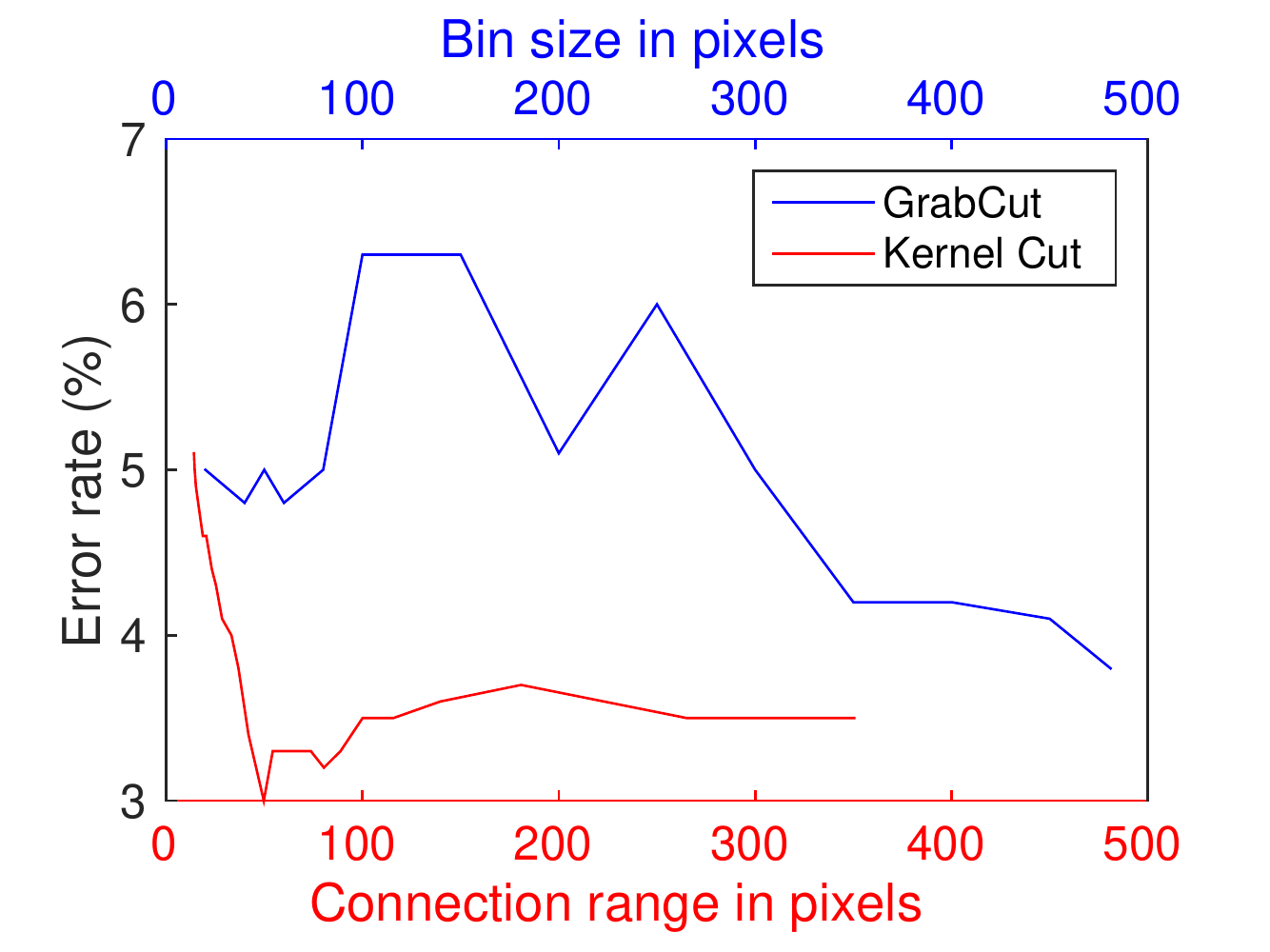}
\end{center}
   \caption{Error on Multi-objects dataset. We vary spatial bin-size for GrabCut and weight $\rs$ in $[l, a, b,\rs X,\rs Y]$ for Kernel Cut. The connection range is the average geometric distance between a pixel and its $k^{th}$ nearest neighbor. 
   {\bf The right-most point of the curves corresponds to the absence of XY features.} GrabCut does not benefit from XY features.
    Kernel Cut achieves the best error rate of 2.9\% with connection range of 50 pixels.}
\label{fig:rgbxyxy}
\end{figure}

We test the effect of adding XY into feature space for GrabCut and Kernel Cut. 
We try various $\rs$ for Kernel Cut. Fig.\ref{fig:knngraph} shows the effect of different $\rs$ on \KNN/s of a pixel. For histogram-based GrabCut we change spatial bin size for the XY channel, ranging from 30 pixels to the image size. We report quantitative results on 18 images with similar objects and camouflage from the Berkeley 
database \cite{martin2001database}. Seeds are used here. Fig. \ref{fig:rgbxyxy} shows average errors for multi-object dataset, see example segmentations in Fig. \ref{fig:dbsamples}.



\begin{figure}[h]
\begin{center}
    \begin{subfigure}[b]{0.35\linewidth}
        \includegraphics[width=\textwidth]{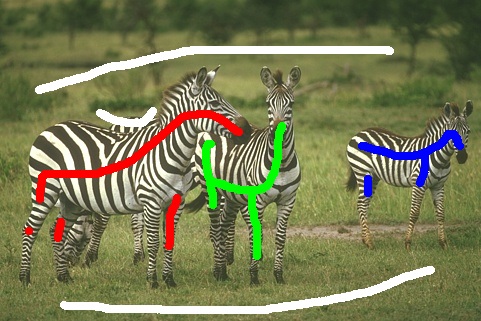}
        \caption{Seeds}
        \label{fig:multiseeds}
    \end{subfigure}
    ~ 
    \begin{subfigure}[b]{0.35\linewidth}
        {%
\setlength{\fboxsep}{0pt}%
\setlength{\fboxrule}{1pt}%
\fbox{\includegraphics[width=\textwidth]{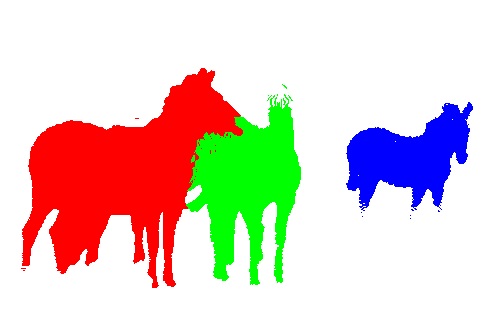}}%
}%

        \caption{Our solution}
        \label{fig:multisolution}
    \end{subfigure}
    \\ 
    \begin{subfigure}[b]{1.0\linewidth}
    \center
        \includegraphics[width=0.4\textwidth]{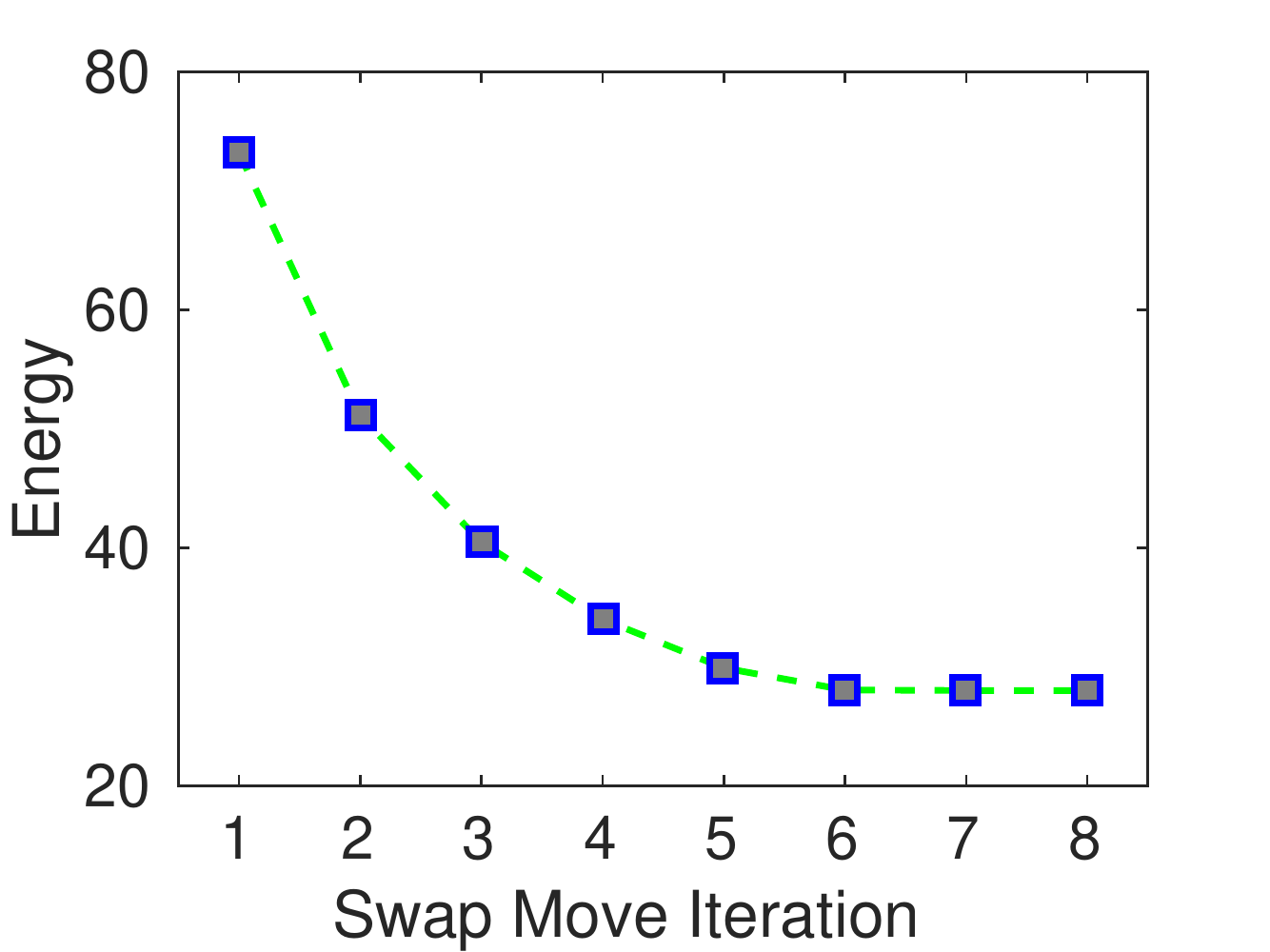}
        \caption{Energy minimization for \NC/ plus smoothness}
        \label{fig:swapmove}
    \end{subfigure}
    
\end{center}
   \caption{Multi-label segmentation for similar objects. }
\label{fig:multi}
\end{figure}

Fig. \ref{fig:multi} gives multi-label segmentation of similar objects in one image with seeds using our algorithm. We optimize kernel bound with move-making for \NC/ and smoothness term combination, 
as discussed in Sec. \ref{sec:KB}. Fig. \ref{fig:multi} (c) shows energy convergence.

\subsubsection{Texture segmentation}
\label{sec:exptexture}

\begin{figure}
\includegraphics[width=\columnwidth]{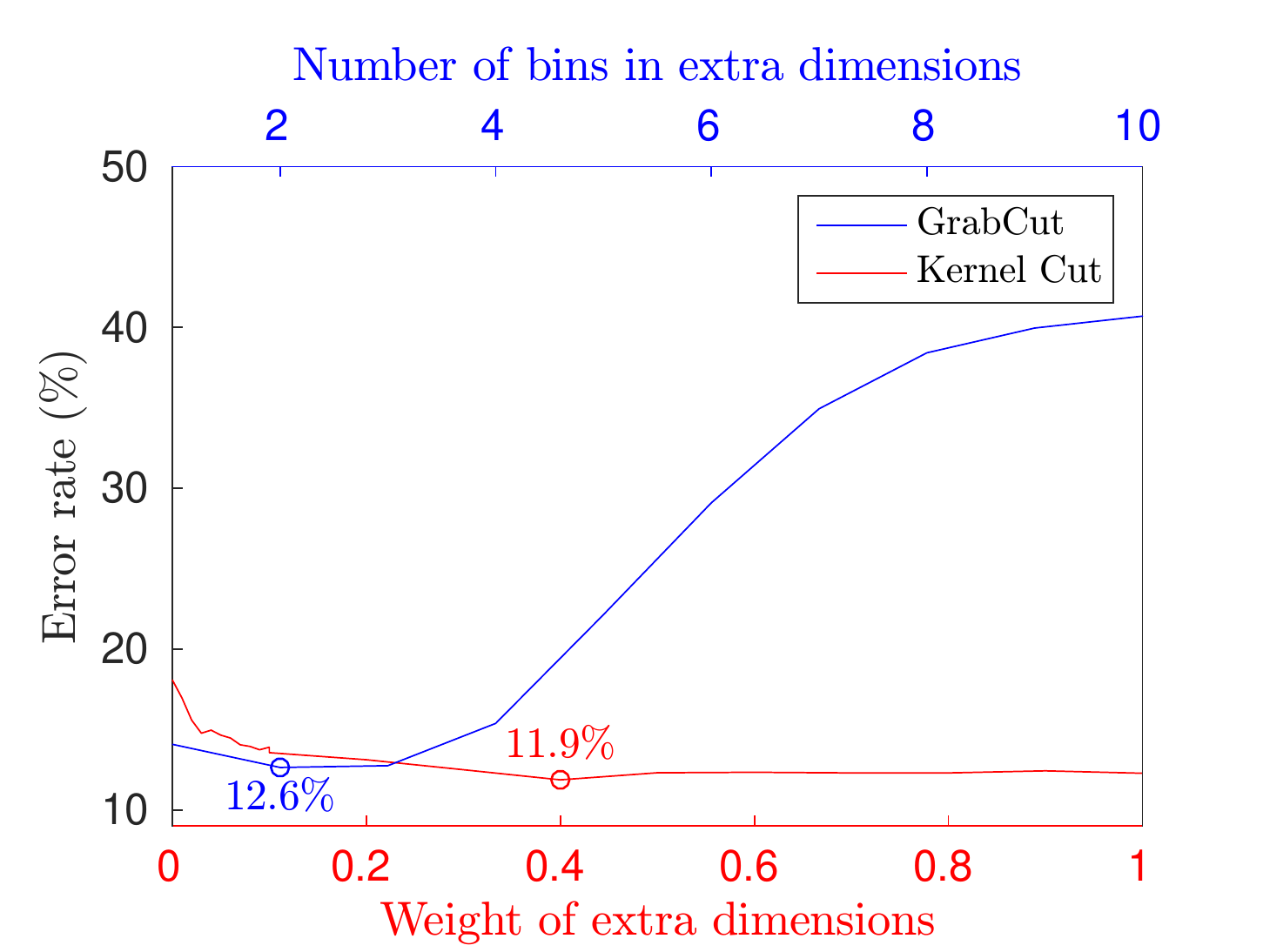}
\caption{The average errors of GrabCut a Kernel Cut methods 
for texture segmentation over 50 desaturated images from 
GrabCut database~\cite{GrabCuts:SIGGRAPH04}. 
We optimize the result of GrabCut with respect to smoothness weight 
and bin sizes in the intensity dimension. We optimize the result of Kernel Cut 
with respect to smoothness weight.}
\label{fig:texture_stat}
\end{figure}

The goal of this experiment is to demonstrate  scalability of our methods to highly dimensional data.
First, desaturated images from GrabCut database~\cite{GrabCuts:SIGGRAPH04} 
are convolved with 48 filters from~\cite{Zisserman:2005statistical}.
This yields a 48-dimensional descriptor for each pixel. 
Secondly, these descriptors are clustered into 32 textons by K-means. 
Thirdly, for each pixel we build a 32-dimensional normalized histogram of 
textons in $5\times5$ vicinity of the pixel.
Then the gray-scale intensity\footnote{We found that for the GrabCut database
adding texture features to RGB does not improve the results. }
 of a pixel is augmented by the corresponding texton histogram scaled by a factor $w$. 
Finally, resulting 33-dimensional feature vectors are used for segmentation. 
We show the result of Kernel Cut with respect to $w$ in Fig.\ref{fig:texture_stat}.
We compare our results with GrabCut with various bin sizes for texture features.

\subsubsection{Interactive RGBD Images Segmentation}
\label{sec:expdepth}

\newcommand{\mdapic}[2][]{\includegraphics[width=0.135\textwidth,clip=true,#1]{figures/rgbd_ex/#2}}
\newcommand{\mdapicwbox}[2][]{%
\begin{tikzpicture}
    \node[anchor=south west,inner sep=0] (image) at (0,0) {\includegraphics[width=0.135\textwidth,clip=true,#1]{figures/rgbd_ex/#2}};
    \node[anchor=south west,inner sep=0] (image) at (0,0) {\includegraphics[width=0.135\textwidth,clip=true,#1]{figures/rgbd_ex/#2_box}}; 
\end{tikzpicture}}
 
\begin{figure*}
\setlength\tabcolsep{1pt}
\setlength{\fboxsep}{0pt}%
\begin{tabular}{cccccccc} 
\mdapicwbox[trim=45pt 0 5pt 37.5pt]{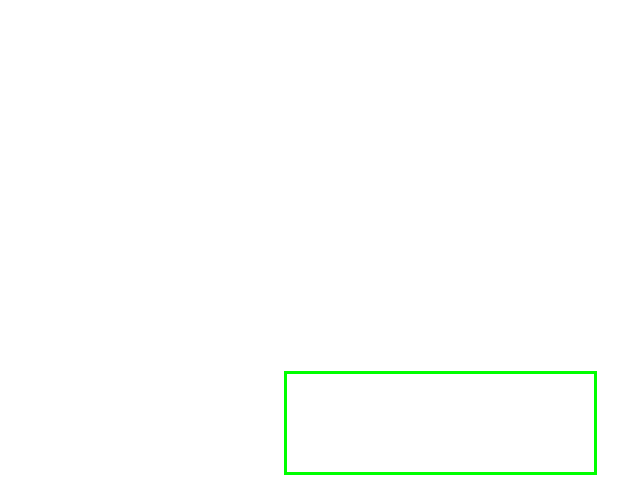} & 
\mdapicwbox{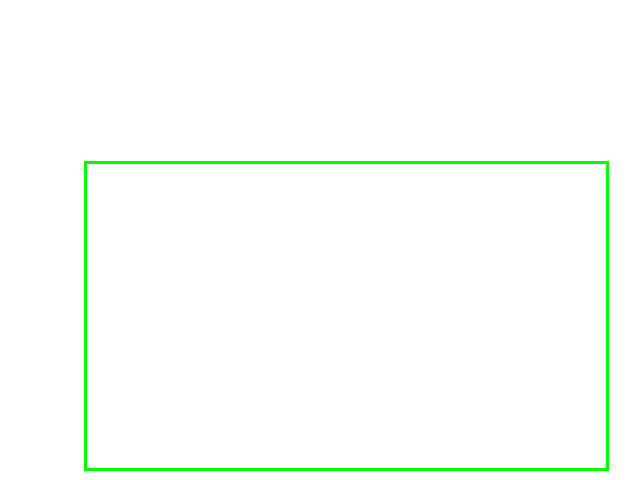} &
\mdapicwbox{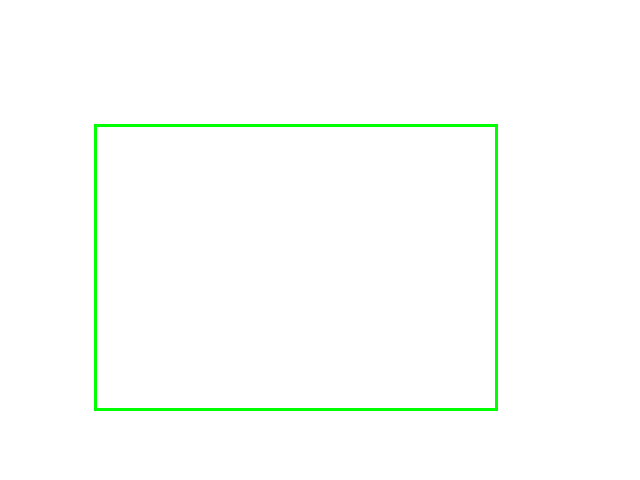} & 
\mdapicwbox[trim=0 20pt 50pt 17.5pt]{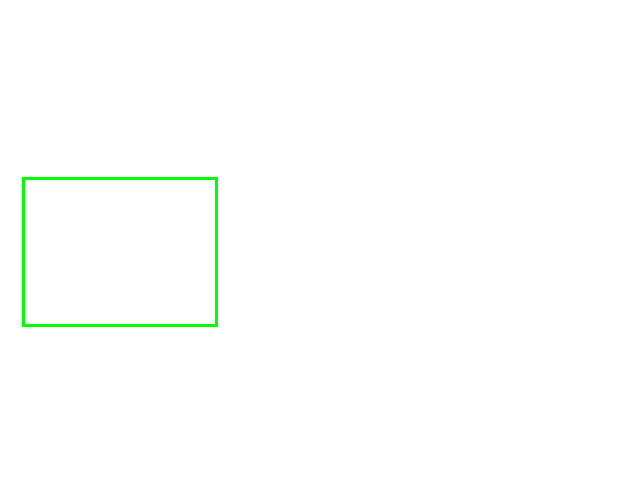} & 
\mdapicwbox[trim=30pt 5pt 0 17.5pt]{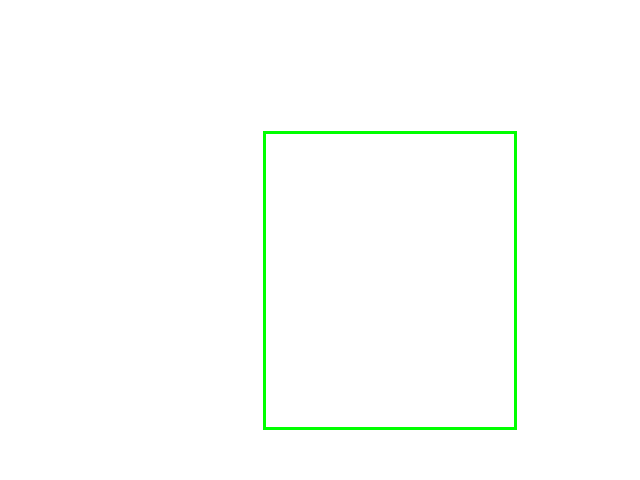} & 
\mdapicwbox{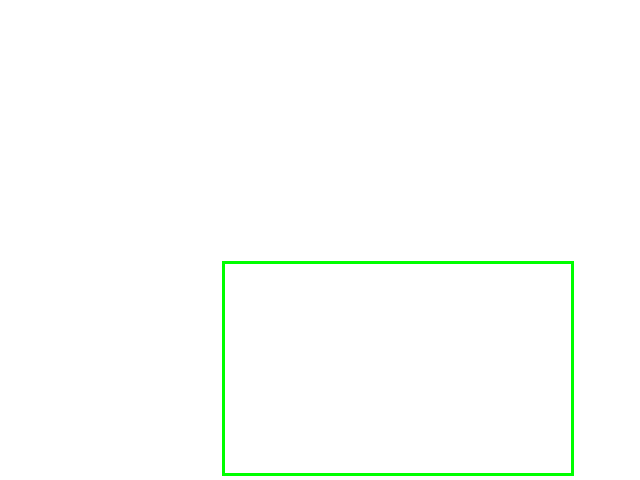} &
\mdapicwbox[trim=0 20pt 40pt 10pt]{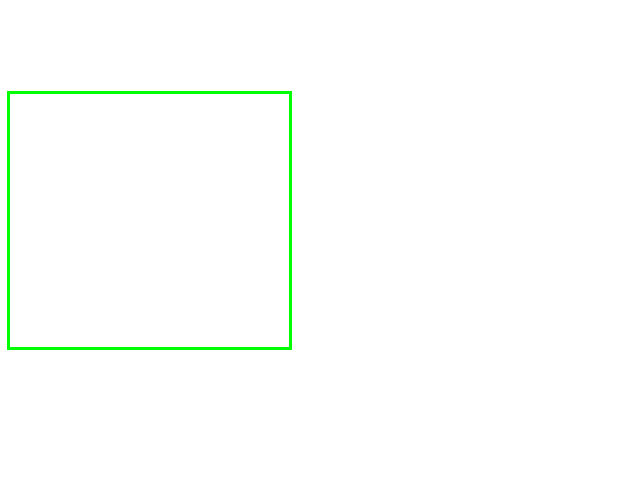} \\ 
\mdapic[trim=45pt 0 5pt 37.5pt]{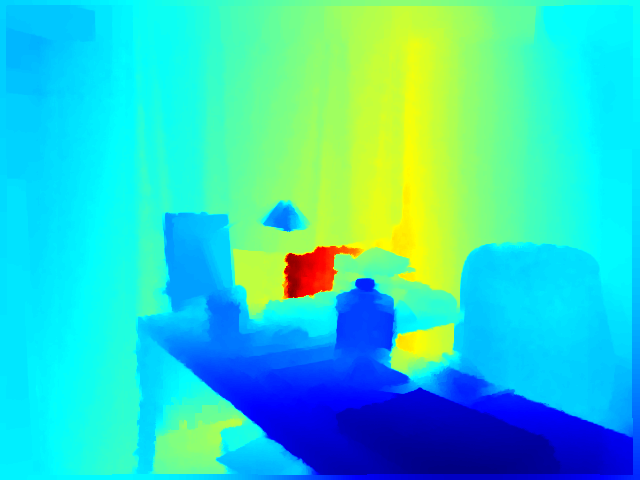} & 
\mdapic{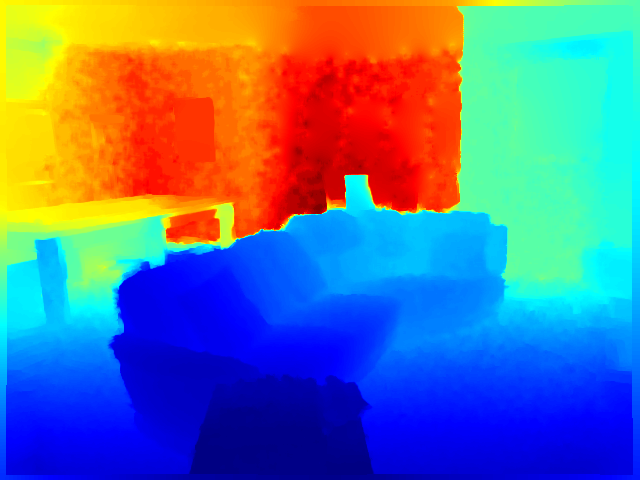} & 
\mdapic{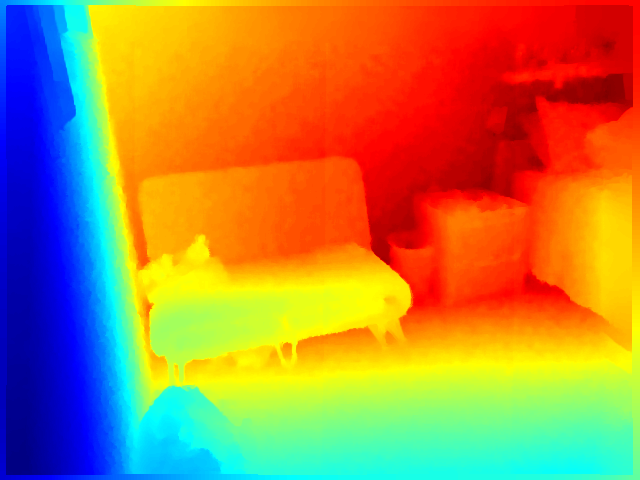} & 
\mdapic[trim=0 20pt 50pt 17.5pt]{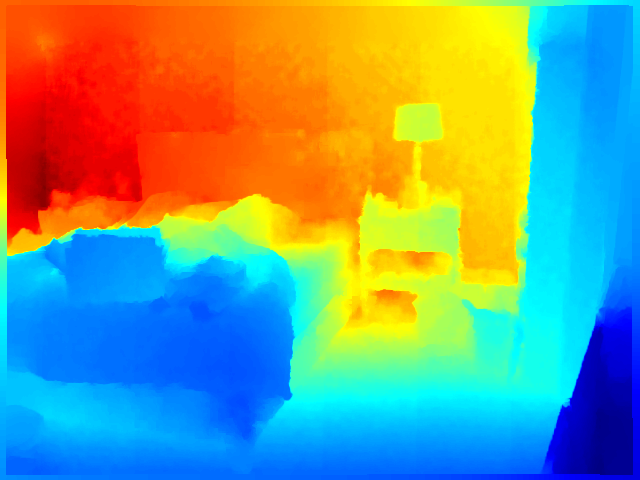} & 
\mdapic[trim=30pt 5pt 0 17.5pt]{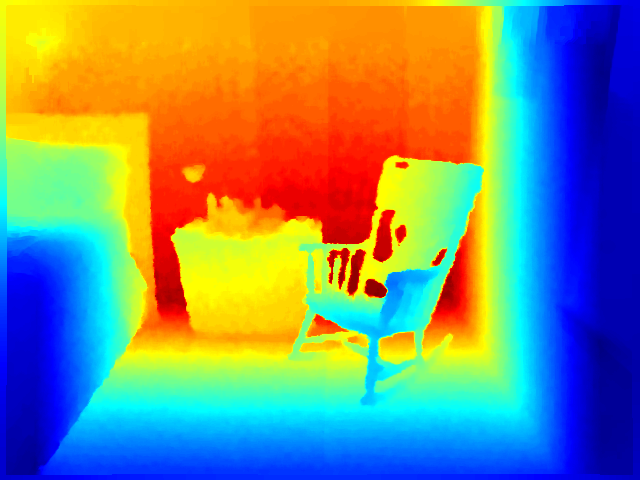} & 
\mdapic{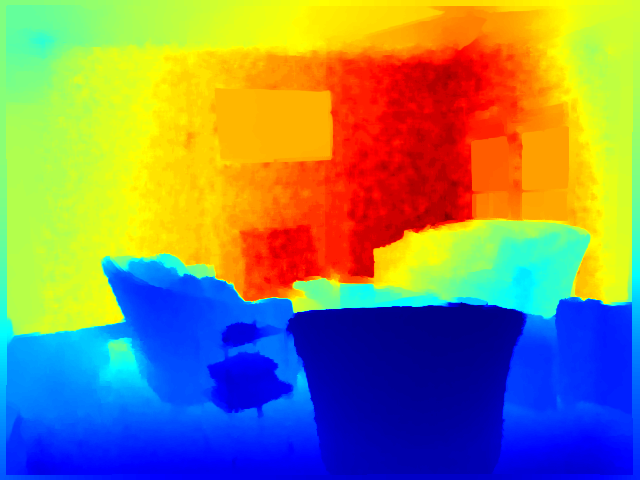} &
\mdapic[trim=0 20pt 40pt 10pt]{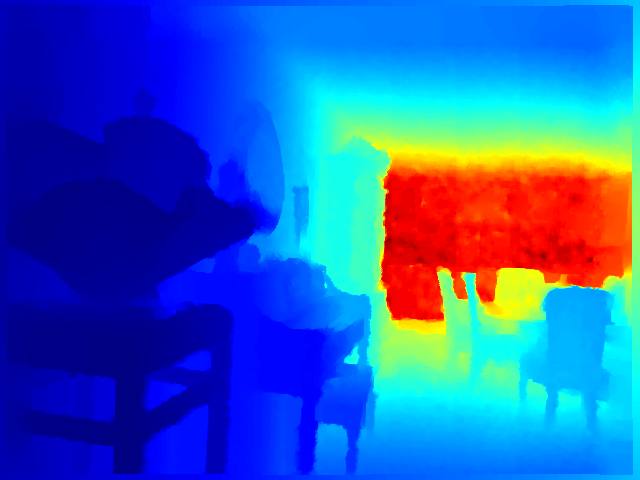} \\
\fbox{\mdapic[trim=45pt 0 5pt 37.5pt]{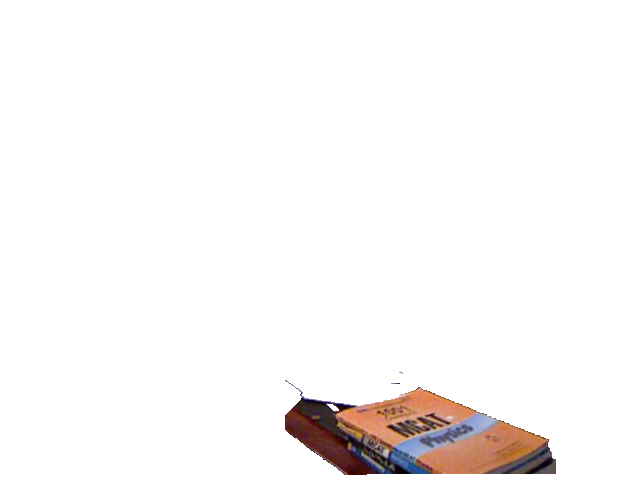}} & 
\fbox{\mdapic{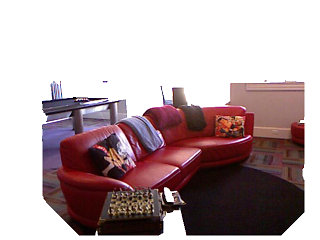}} & 
\fbox{\mdapic{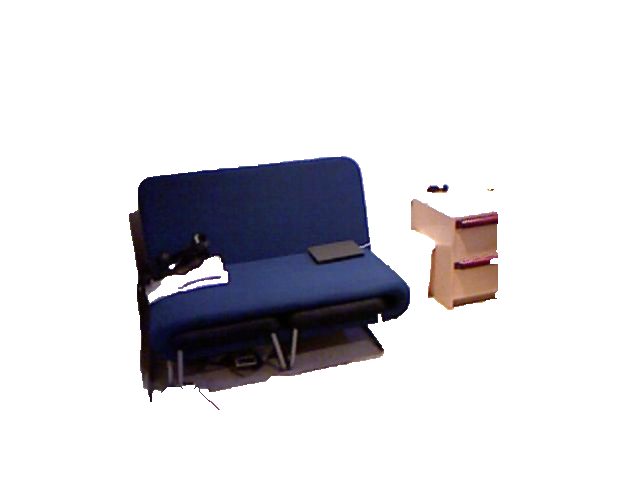}} & 
\fbox{\mdapic[trim=0 20pt 50pt 17.5pt]{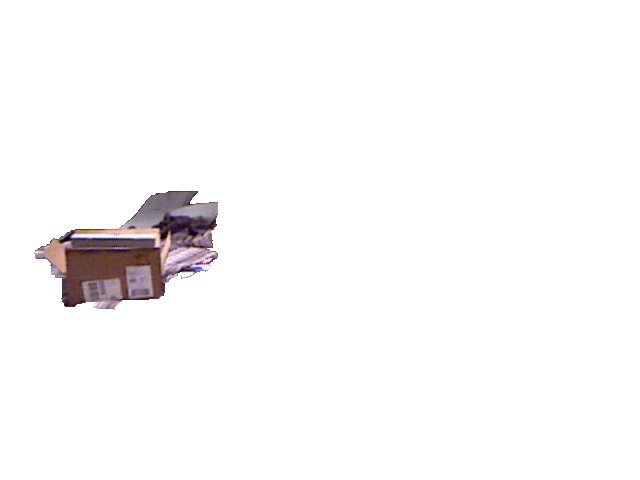}} & 
\fbox{\mdapic[trim=30pt 5pt 0 17.5pt]{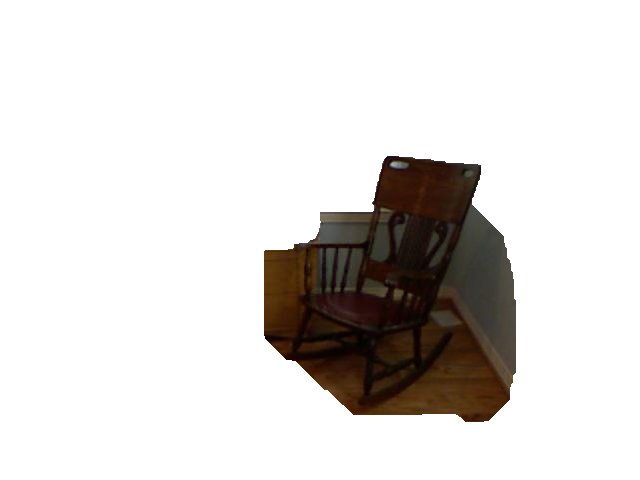}} & 
\fbox{\mdapic{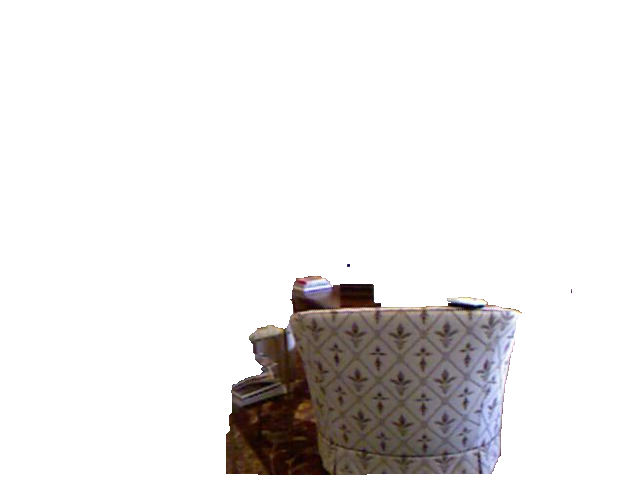} }& 
\fbox{\mdapic[trim=0 20pt 40pt 10pt]{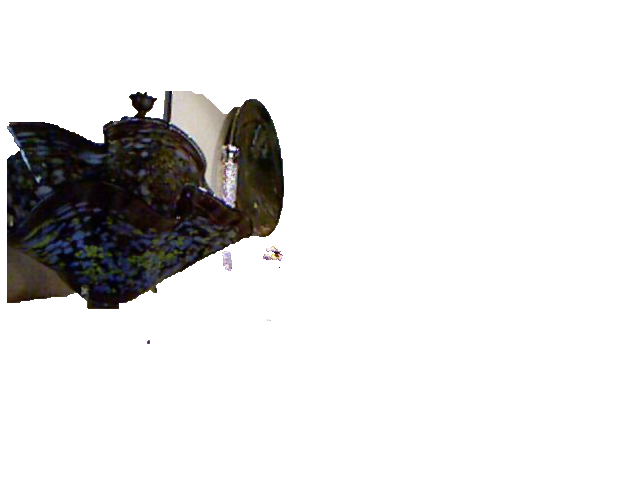}} \\ 
\fbox{\mdapic[trim=45pt 0 5pt 37.5pt]{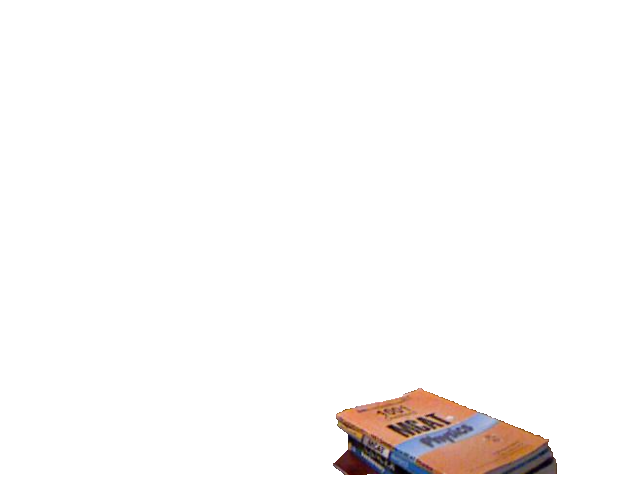}} & 
\fbox{\mdapic{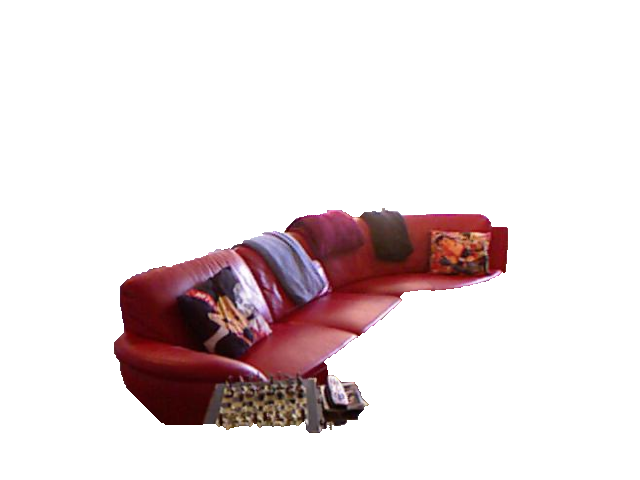}} & 
\fbox{\mdapic{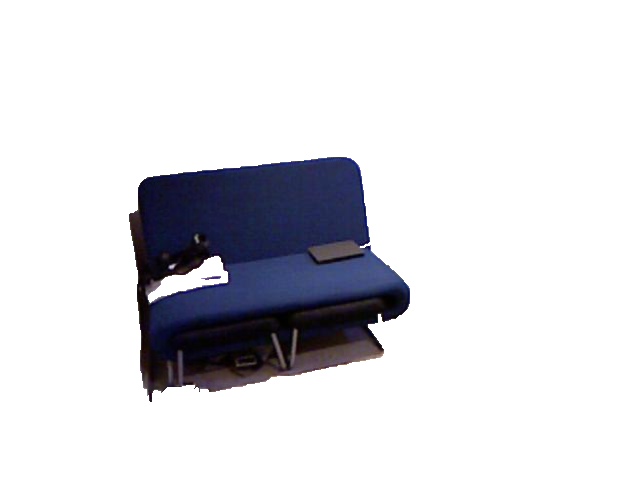}} & 
\fbox{\mdapic[trim=0 20pt 50pt 17.5pt]{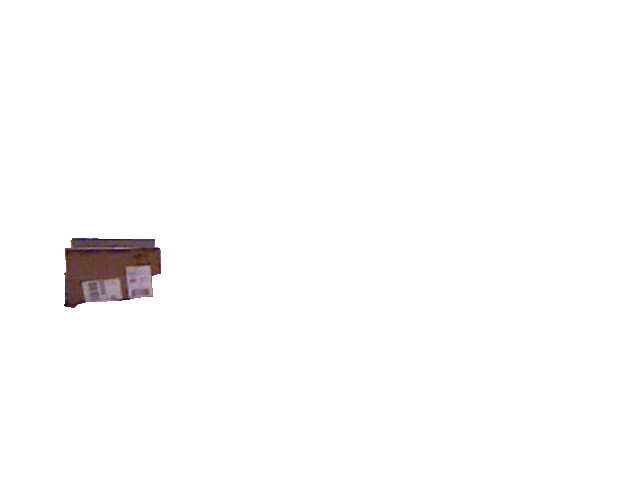}} & 
\fbox{\mdapic[trim=30pt 5pt 0 17.5pt]{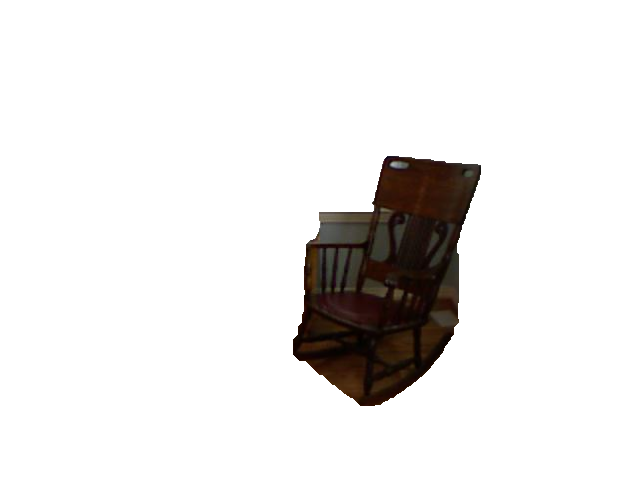}} &
\fbox{\mdapic{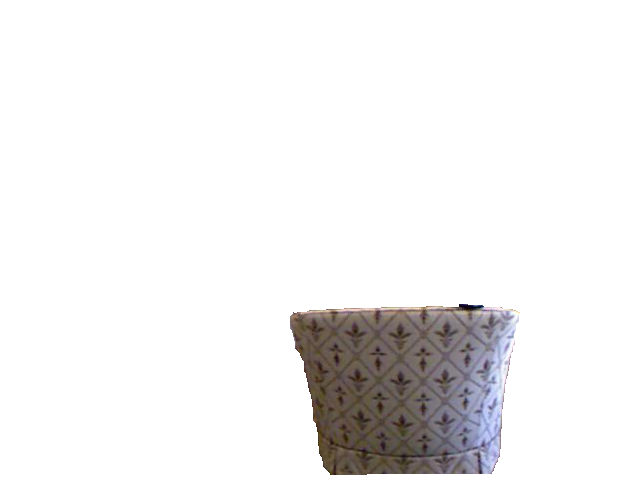}} &  
\fbox{\mdapic[trim=0 20pt 40pt 10pt]{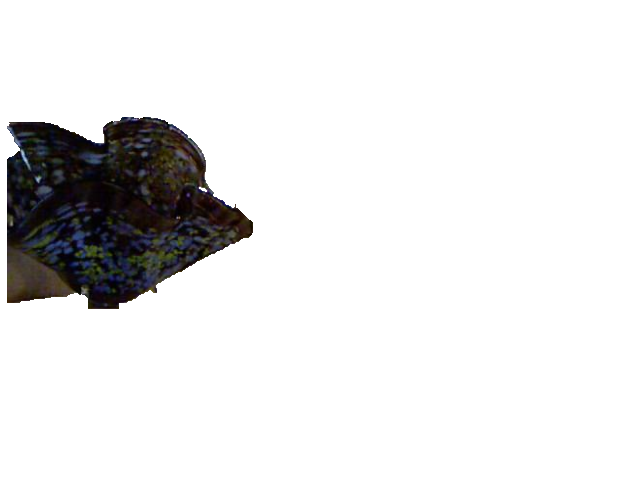}}
\end{tabular}
\caption{RGBD+XY examples. The first two rows show original images wit bounding box and color-coded depth channel. The third row shows the results of Grabcut, the forth row shows the results of Kernel Cut. Parameters of the  methods were independently selected to minimize average error rate over the database. The parameters of the algorithms were selected to minimize the average error over the dataset.}
\label{fig:rgbd_examples}
\end{figure*}

\begin{figure}
\includegraphics[width=\columnwidth]{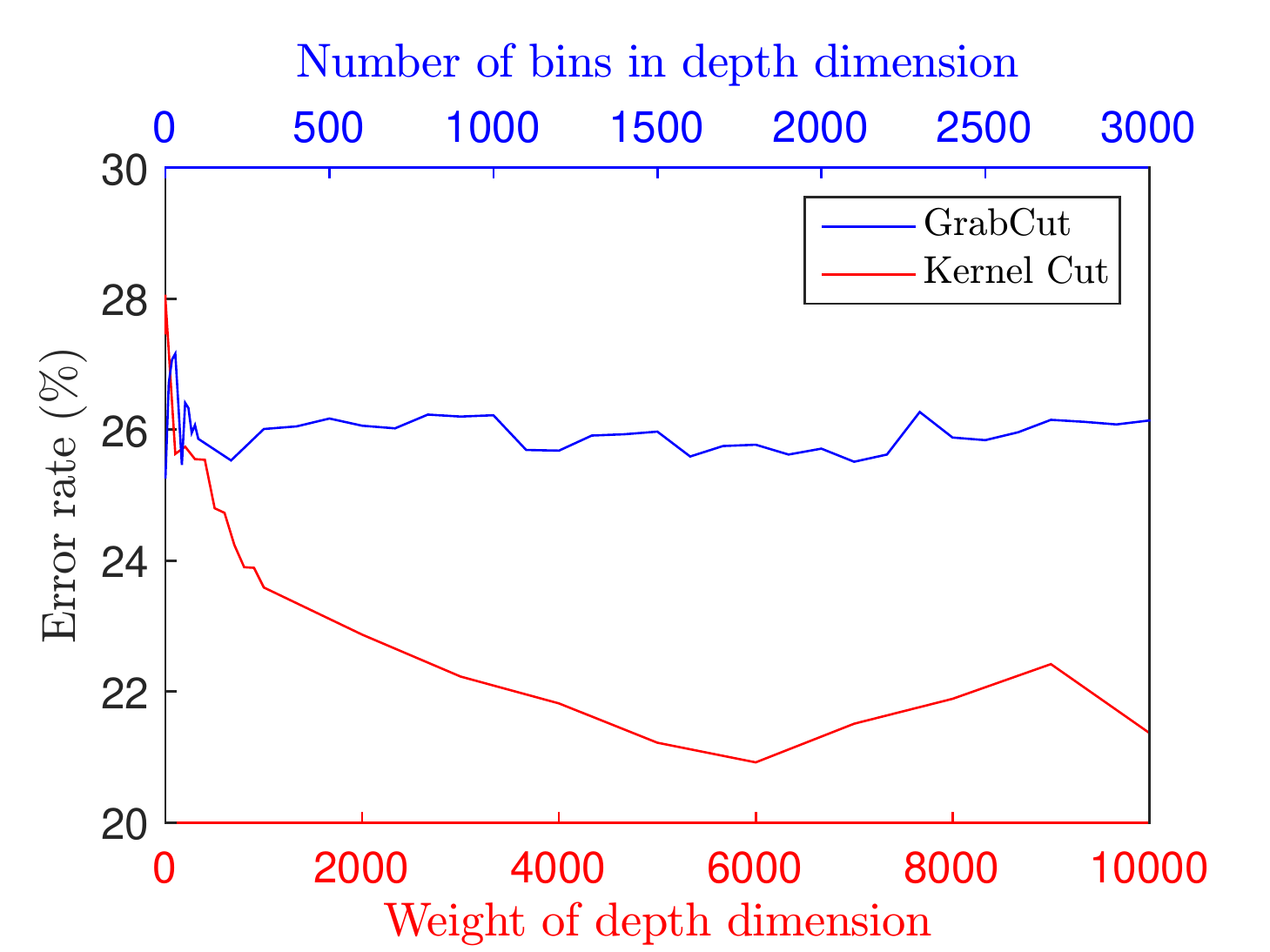}
\caption{The average errors of GrabCut a Kernel Cut methods over 64 images selected from NYUv2 database~\cite{RGBD:database}.}
\label{fig:rgbd_stat}
\end{figure}

Depth sensor are widely used in vision for
3D modelling~\cite{RGBD:dou20153d,RGBD:kinectfusion}, 
semantic segmentation~\cite{RGBD:dengsemantic,RGBD:gulshan2011humanising,RGBD:database,RGBD:ren2012rgb}, 
motion flow~\cite{RGBD:gottfried2011computing}. 
We selected 64 indoor RGBD images from semantic segmentation database NYUv2~\cite{RGBD:database} and
provided bounding boxes and ground truth. 
In contrast to \cite{GrabCuts:SIGGRAPH04}, the prepared dataset consists of low-quality images: 
there are camera motion artifacts, underexposed and overexposed regions.
Such artifacts make color-based segmentation harder.


We compare GrabCut to Kernel Cut over joint features $I_p=[L_p, a_p, b_p, \rs D_p]$ as in Sec.\ref{sec:expxy}. 
Fig.\ref{fig:rgbd_examples} shows the error statistics and segmentation examples.
While Kernel Cut takes advantage of the additional channel, GrabCut fails to improve.

\subsubsection{Motion segmentation} 
\label{sec:expmotion}

\begin{figure*}
\centering
\setlength{\fboxsep}{0pt}
\setlength{\fboxrule}{1pt}
   \fbox{\includegraphics[width=0.19\linewidth]{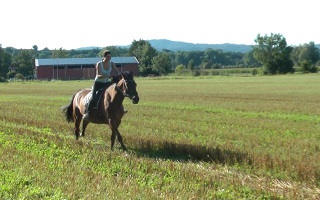}}
   \includegraphics[width=0.19\linewidth]{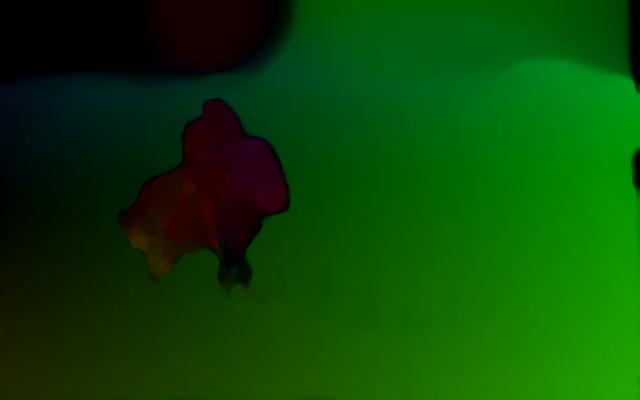}
    \fbox{ \includegraphics[width=0.19\linewidth]{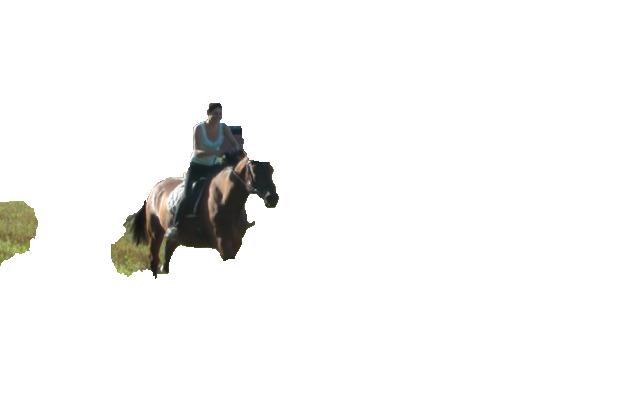}}
     \fbox{ \includegraphics[width=0.19\linewidth]{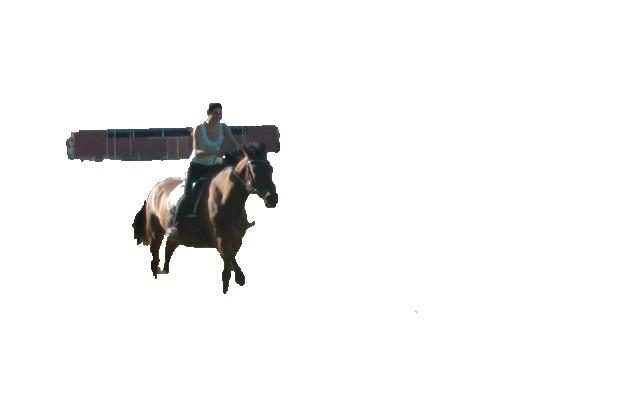}}
      \fbox{ \includegraphics[width=0.19\linewidth]{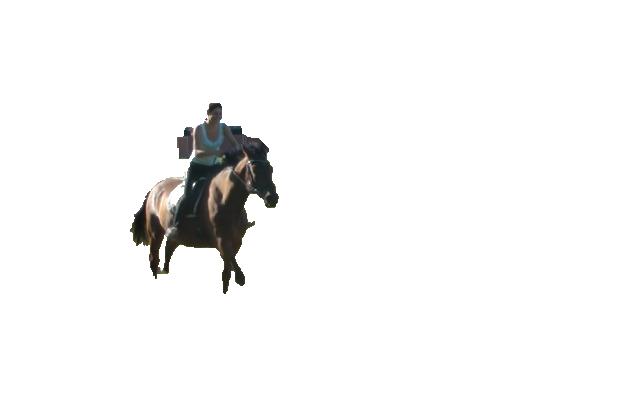}}
      \\
       \fbox{\includegraphics[width=0.19\linewidth]{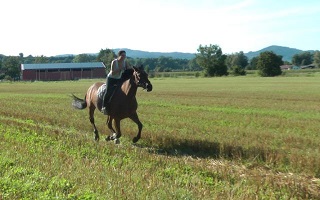}}
   \includegraphics[width=0.19\linewidth]{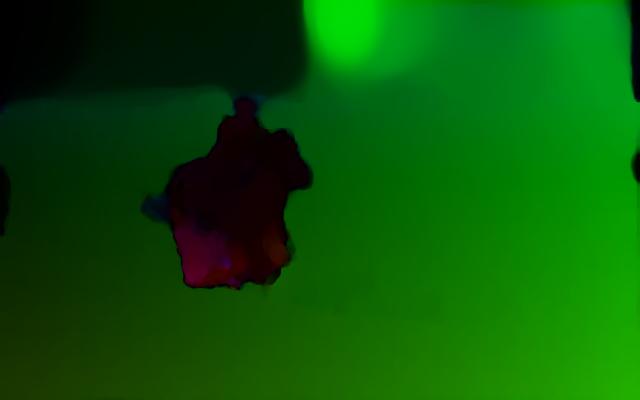}
    \fbox{ \includegraphics[width=0.19\linewidth]{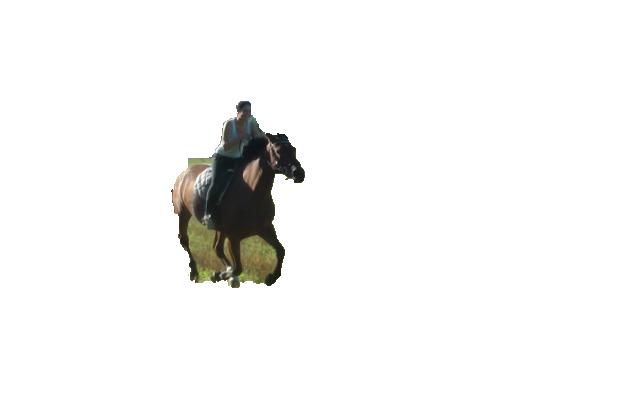}}
     \fbox{ \includegraphics[width=0.19\linewidth]{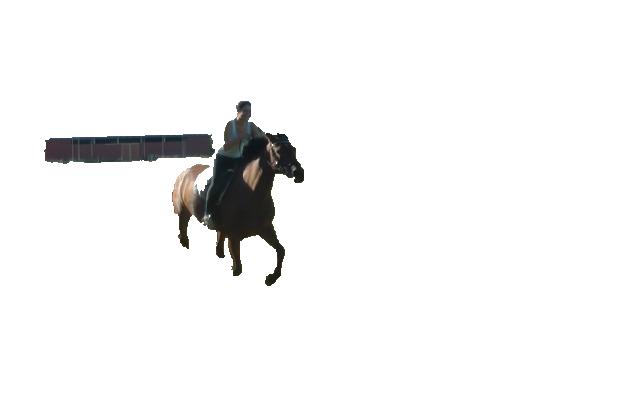}}
      \fbox{ \includegraphics[width=0.19\linewidth]{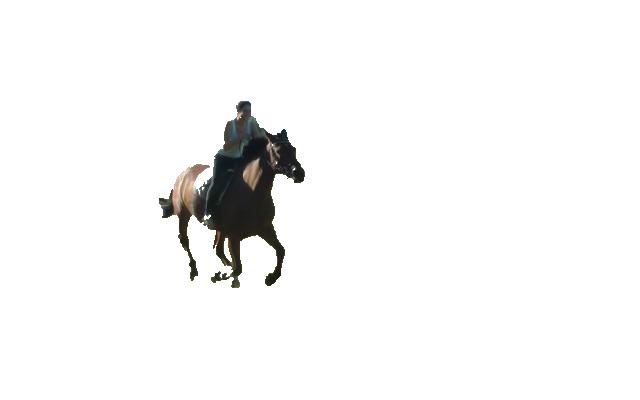}}
      \\
       \fbox{\includegraphics[width=0.19\linewidth]{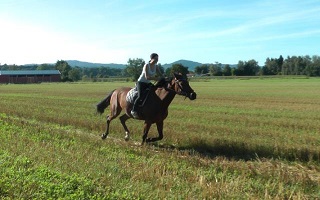}}
   \includegraphics[width=0.19\linewidth]{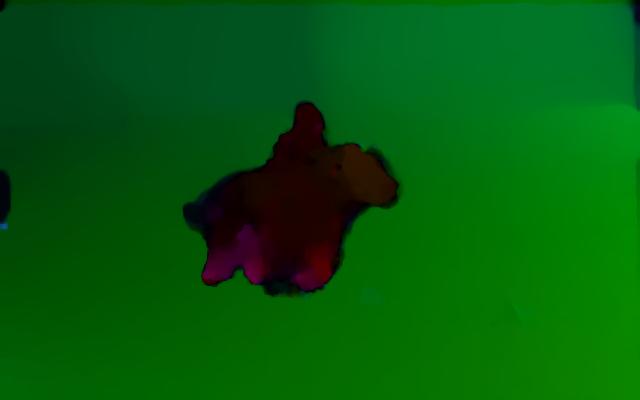}
    \fbox{ \includegraphics[width=0.19\linewidth]{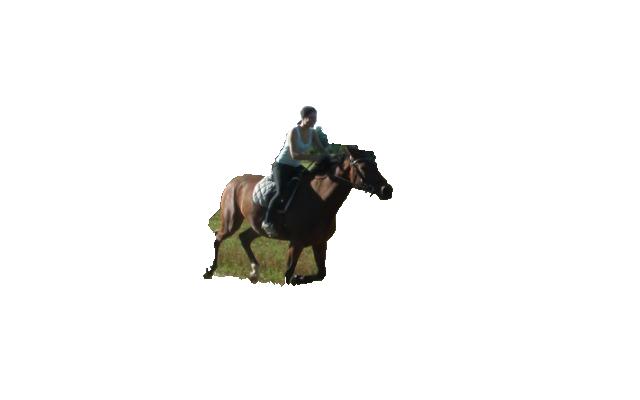}}
     \fbox{ \includegraphics[width=0.19\linewidth]{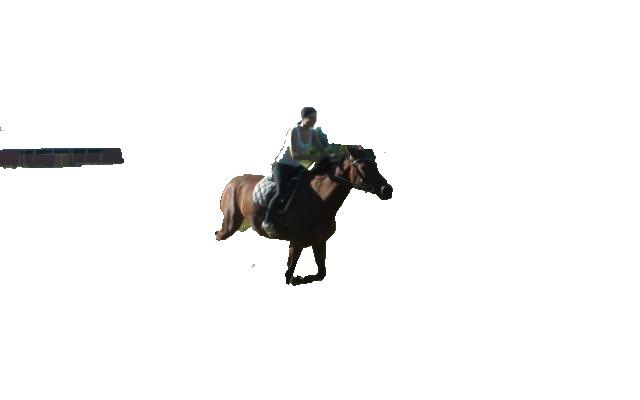}}
      \fbox{ \includegraphics[width=0.19\linewidth]{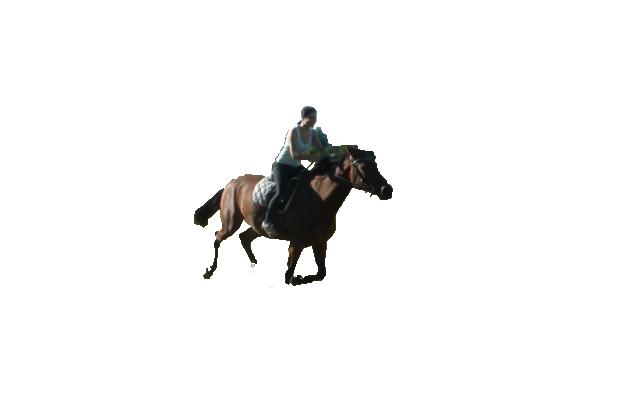}}
       \\
       \fbox{\includegraphics[width=0.19\linewidth]{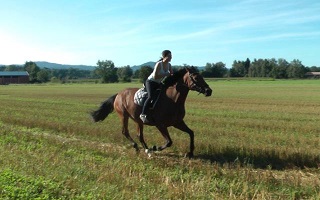}}
   \includegraphics[width=0.19\linewidth]{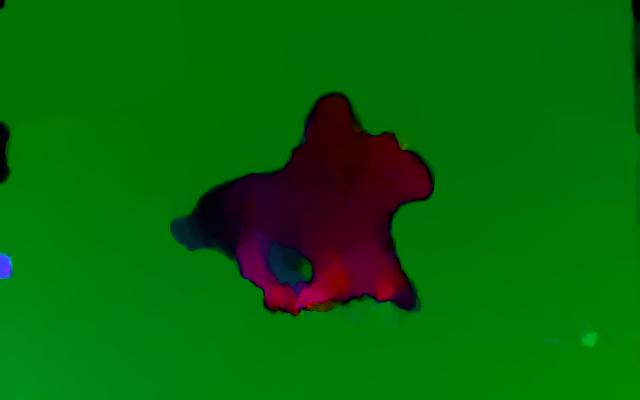}
    \fbox{ \includegraphics[width=0.19\linewidth]{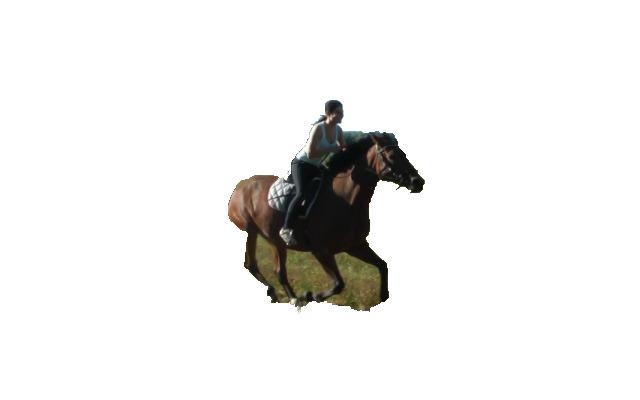}}
     \fbox{ \includegraphics[width=0.19\linewidth]{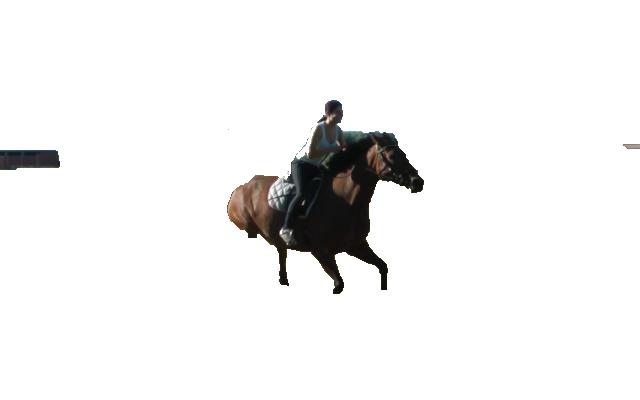}}
      \fbox{ \includegraphics[width=0.19\linewidth]{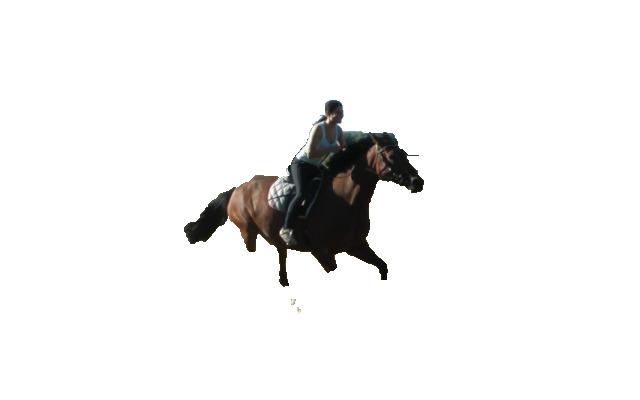}}
      \\
      \flushleft
      ~~~~~~~~~~~(a) frames ~~~~~~~~~~~~~~~~(b) optical flow \cite{Bro11a}~~~~~~~~~~~~~~(c) \textbf{M}+\textbf{XY}~~~~~~~~~~~~~~~~~~~~
      (d) \textbf{RGB}+\textbf{XY}~~~~~~~~~~~~~~~~(e) \textbf{RGBM}+\textbf{XY}
   \caption{Motion segmentation using our framework for the sequence \textit{horses01} in FBMS-59 dataset \cite{brox2010object}. Motion feature alone (\textbf{M}+\textbf{XY} in (c)) is not sufficient to obtain fine segmentation. Our framework successfully utilize motion feature (optical flow) to separate the horse from the barn, which have similar appearances.}
\label{fig:horses}
\end{figure*}

\begin{figure*}
\centering
\setlength{\fboxsep}{0pt}
\setlength{\fboxrule}{1pt}
   \fbox{\includegraphics[width=0.23\linewidth]{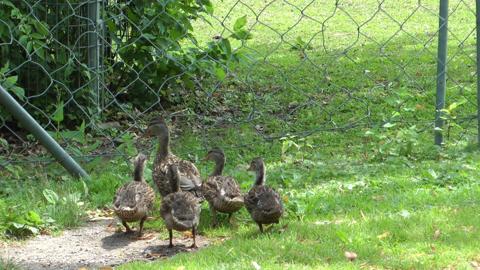}}
   \includegraphics[width=0.23\linewidth]{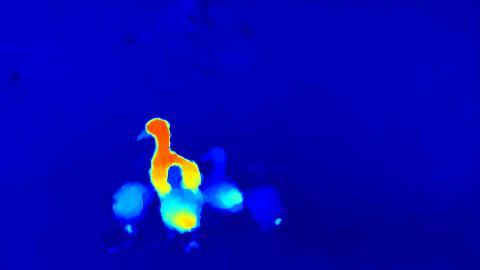}
    \fbox{ \includegraphics[width=0.23\linewidth]{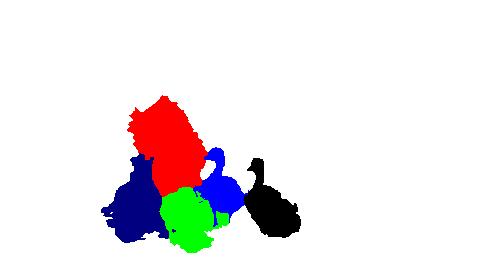}}
     \fbox{ \includegraphics[width=0.23\linewidth]{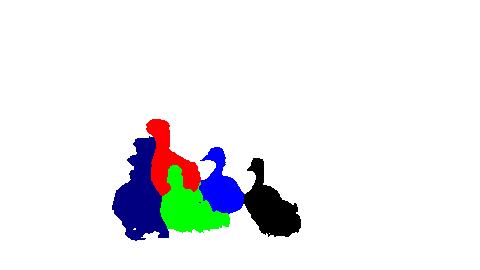}}
     \\
      \fbox{\includegraphics[width=0.23\linewidth]{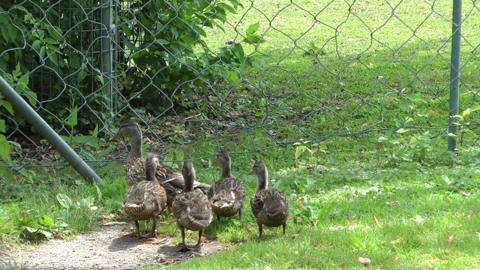}}
   \includegraphics[width=0.23\linewidth]{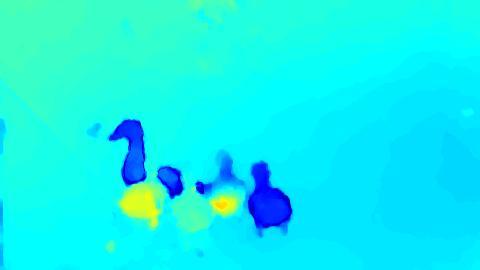}
    \fbox{ \includegraphics[width=0.23\linewidth]{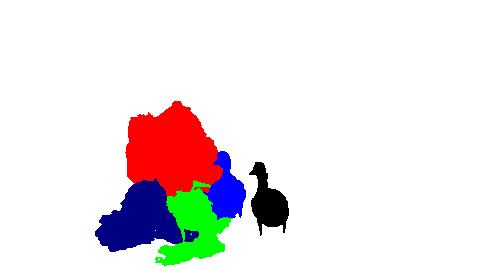}}
     \fbox{ \includegraphics[width=0.23\linewidth]{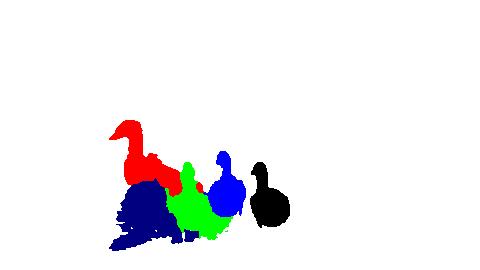}}
     \\
      \fbox{\includegraphics[width=0.23\linewidth]{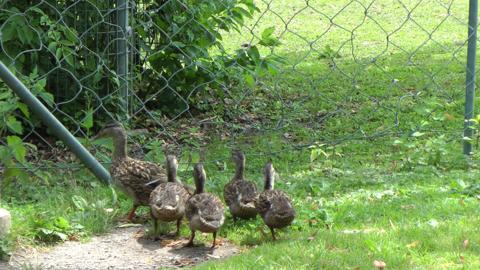}}
   \includegraphics[width=0.23\linewidth]{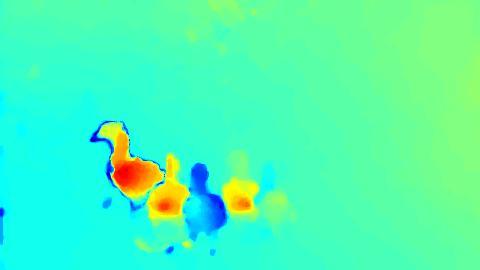}
    \fbox{ \includegraphics[width=0.23\linewidth]{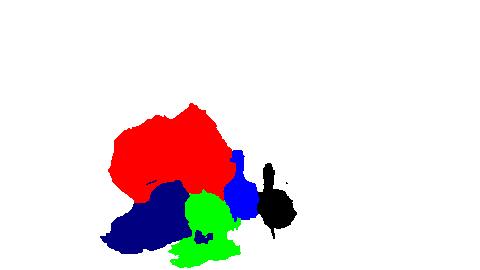}}
     \fbox{ \includegraphics[width=0.23\linewidth]{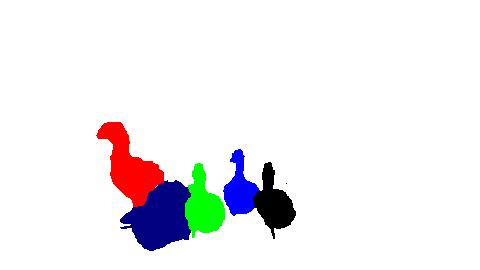}}
     \\
      \fbox{\includegraphics[width=0.23\linewidth]{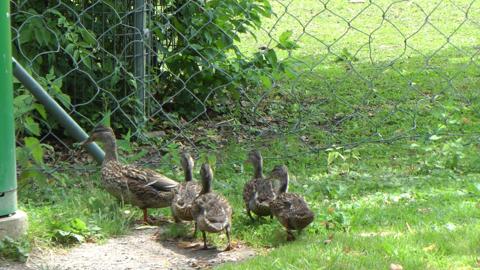}}
   \includegraphics[width=0.23\linewidth]{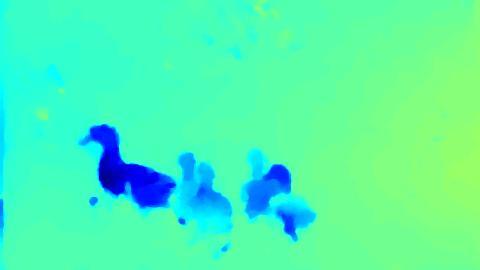}
    \fbox{ \includegraphics[width=0.23\linewidth]{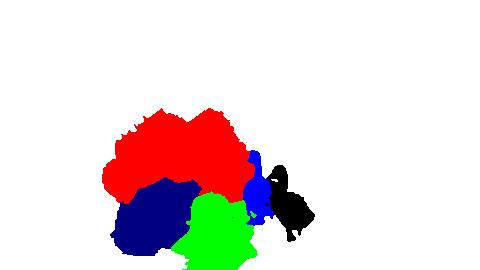}}
     \fbox{ \includegraphics[width=0.23\linewidth]{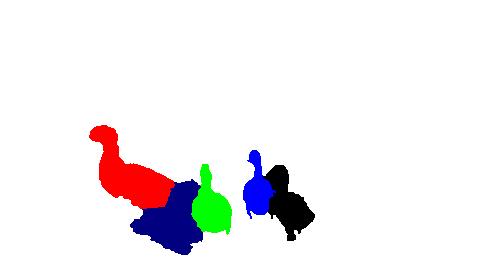}}
      \flushleft
      ~~~~~~~~~~~~~~~~~~~~(a) frames ~~~~~~~~~~~~~~~~~~~~~~~~~(b) optical flow \cite{Bro11a}~~~~~~~~~~~~~~~~~(c) \textbf{RGBXY}+\textbf{XY}~~~~~~~~~~~~~~~~
      (d) \textbf{RGBXYM}+\textbf{XY}
      \caption{Multi-label motion segmentation using our framework for the sequence \textit{ducks01} in FBMS-59 dataset \cite{brox2010object}. This video is challenging since the ducks here have similar appearances and even spatially overlap with each other. However, different ducks come with different motions, which helps our framework to better separate individual ducks.}
\label{fig:ducks}
\end{figure*}

\begin{figure*}
\centering
\begin{tabular}{cc}
\multirow{1}{*}[30mm]{\rotatebox[origin=c]{90}{Motion Flow}} & \fbox{\includegraphics[width=0.7\textwidth]{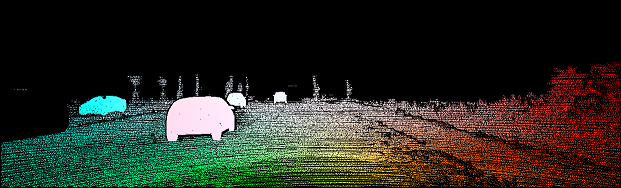}}\\
\multirow{1}{*}[25mm]{\rotatebox[origin=c]{90}{RGB+XY}} & \fbox{\includegraphics[width=0.7\textwidth]{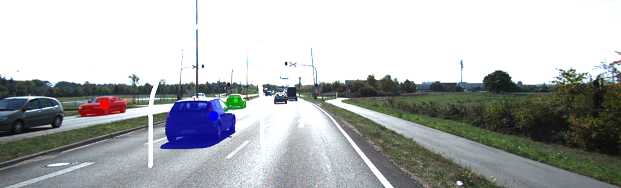}}\\
\multirow{1}{*}[25mm]{\rotatebox[origin=c]{90}{MXY+XY}} &\fbox{\includegraphics[width=0.7\textwidth]{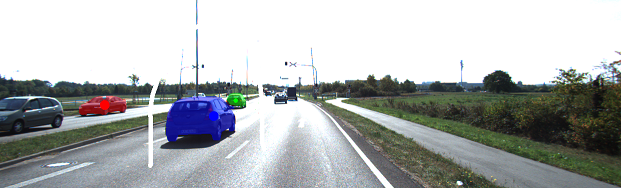}}
\end{tabular}
\caption{Motion segmentation for image 000079\_10 from \emph{KITTI}~\cite{Menze2015CVPR} dataset. The first row shows the motion flow. Black color codes the pixels that do not have motion information. The second row shows color-based segmentation. The third row shows motion based segmentation with location features. We also tried M+XY segmentation, but it does not work as well as MXY+XY above. The results for RGBMXY+XY were not significantly different from MXY+XY.}
\label{fig:kitti}
\end{figure*}

Besides the location and depth features, we also test segmentation with motion features. 
Figs. \ref{fig:horses}, \ref{fig:ducks} and \ref{fig:kitti} compare motion segmentations using 
different feature spaces: RGB, XY, M (optical flow) and their combinations (RGBM or RGBXY or RGBXYM). 
Abbreviation \textbf{+XY} means Potts regularization. We apply kernel cut (Alg.\ref{alg:kernelcut}) 
to the combination of \NC/ with the Potts term. 

{\bf Challenging video examples:} For videos in FBMS-59 dataset \cite{brox2010object}, our algorithm runs on individual frames instead of 3D volume. Segmentation of previous frame initializes the next frame. The strokes are provided \textit{only} for the first frame. We use the optical flow algorithm in \cite{Bro11a} to generate M features. 
Selected frames are shown in Figs. \ref{fig:horses} and \ref{fig:ducks}. 
Instead of tracks from all frames in \cite{broxiccv11}, our segmentation of each frame 
uses only motion estimation between two consecutive frames.
Our approach jointly optimizes normalized cut and Potts model. In contrast, 
\cite{broxiccv11} first clusters semi-dense tracks via spectral clustering \cite{brox2010object} 
and then obtains dense segmentation via regularization.


\graphicspath{{./figures/}} 

{\bf Kitti segmentation example:} We also experiment with Kitti dataset~\cite{Menze2015CVPR}. Fig.\ref{fig:kitti} shows the multi-label segmentation using either color information RGB+XY (first row) or motion MXY+XY (second row). The ground-truth motion field works as M channel. Note that the motion field is known only for approximately 20\% of the pixels. To build an affinity graph, we construct a \KNN/ graph from pixels that have motion information. The regularization over 8-neighborhood on the pixel grid
interpolates the segmentation labels during the optimization procedure.

\renewcommand{\theequation}{A-\arabic{equation}}
  \setcounter{equation}{0}
  \section*{APPENDIX A (Weighted \KM/ and \AA/) }

This Appendix reviews weighted kernel K-means (\wkKM/) in detail. In particular, it describes
generalizations of \KM/ procedures  \eqref{eq:exp_kKMproc} and \eqref{eq:imp_kKMproc} to the weighted case
and shows that they also correspond to linear bound optimization by extending 
Theorem~\ref{th:kKM_bound1} in Sec.~\ref{sec:KM_and_BO}.
We provide an alternative derivation for the kernel bound for \NC/. The Appendix also discusses equivalence of 
weighted \AA/ with arbitrary affinity to \wkKM/ with p.s.d. kernel and explains the corresponding {\em diagonal shift}.

{\bf Weighted kernel K-means (\wkKM/):} As discussed in Section~\ref{sec:kKM},
weighted K-means corresponds to objective \eqref{eq:wkKM}
\begin{eqnarray} 
F^w(S) & = & \sum_k \sum_{p \in S^k} w_p \|\phi_p-\mu^w_{S^k} \|^2  \label{eq:kKMweighted}   \\
             & \utc & -\sum_k \;\; \frac{ {S^k}'W \,{\cal K}\, W S^k }{{S^k}'w}   \label{eq:AAweighted}
\end{eqnarray}
where $\| . \| $ is the Euclidean norm, $w:=\{w_p|p\in\Omega\}$ are predefined weights,  $\phi_p\equiv\phi(I_p)$ is an {\em embedding} 
of data points in some high-dimensional space,  and $\mu^w_{S^k}$ is a weighted cluster mean \eqref{eq:wmean}.
Consistently with Sec.~\ref{sec:kKM} we use diagonal matrix $W=diag(w)$ and embedding matrix $\phi :=[\phi_p]$ 
implying identities \eqref{eq:identities} and matrix formulation \eqref{eq:AAweighted} 
with p.s.d. kernel $${\cal K} \;\;=\;\; \phi'\phi$$ of dot products ${\cal K}_{pq} = \phi_p'\phi_q$.
The constant connecting equivalent objectives \eqref{eq:kKMweighted} and \eqref{eq:AAweighted} is $\sum_{p\in\Omega}w_p \|\phi_p\|^2$.

In the context of weighted energy \eqref{eq:kKMweighted} the basic \KM/ algorithm \cite{Duda:01} 
is the {\em block-coordinate descent} for {\em mixed} objective \eqref{eq:wkmixed} 
\begin{eqnarray} 
F^w(S,m)  & :=  & \sum_k \sum_{p \in S^k} w_p\|\phi_p- m_k \|^2    \nonumber \\ 
            & \utc & \sum_k \left(  w \|m_k\|^2 - 2W \phi' m_k   \right)' S^k   \label{eq:wmixedA} 
\end{eqnarray}
where the second linear algebraic formulation\footnote{It is obtained by opening the square of the norm 
and applying algebraic identities \eqref{eq:identities}. Formulation \eqref{eq:mixed} 
omits the same constant as \eqref{eq:AAweighted}. } generalizes \eqref{eq:kmixed0} and
highlights modularity (linearity) with respect to $S$. 
Variables $m_k$ can be seen as ``relaxed'' segment means $\mu^w_{S^k}$ in \eqref{eq:kKMweighted}. 
Yet, energies \eqref{eq:wmixedA} and  \eqref{eq:kKMweighted}
are equivalent since their global minimum is achieved at the same optimal segmentation $S$. Indeed,
\begin{eqnarray} 
\mu^w_{S^k} & = & \arg\min_{m_k} \sum_{p \in S^k} w_p\|\phi_p- m_k \|^2  \nonumber \\   \Rightarrow \;\;\;\;\;\;
F^w(S)      & = & \min_m F^w(S,m).     \label{eq:variousA}
\end{eqnarray}

{\bf Weighted  \KM/ and bound optimization:}
The weighted case of procedure \eqref{eq:exp_kKMproc} replaces $\mu_{S^k}$ \eqref{eq:mu_phi}  
by weighted mean $\mu^w_{S^k}$ \eqref{eq:wmean}
\begin{equation} \label{eq:wKMproc}
\left(\parbox{13ex}{\centering\bf\small weighted \KM/ \\ procedure}\right)\quad\;\;\;
S_p \;\;\;\leftarrow\;\;\; \arg\min_k \|  \phi_p - \mu^w_{S^k_t}  \|. \quad\quad
\end{equation}
Assuming $\phi_p \in {\mathcal R}^m$ each iteration's complexity ${\cal O} (|\Omega|Km)$ is 
linear with respect to the number of data points $|\Omega|$.

Implicit \kKM/ procedure \eqref{eq:imp_kKMproc} generalizes to the weighted case as follows. 
Similarly to Sec.8.2.2 in \cite{TaylorCristianini:04} and our derivation of \eqref{eq:imp_kKMproc} in Sec.\ref{sec:kKM}, 
the square of the objective in \eqref{eq:wKMproc} and \eqref{eq:wmean} give
$$ \cancel{\| \phi_p\|^2} - 2{\phi_p}'\mu_{S^k_t} + \| \mu_{S^k_t}\|^2  \; = \;
-2\frac{{\phi_p}'\phi WS^k_t}{{S^k_t}'w} + \frac{{S^k_t}'W\phi'\phi WS^k_t}{({S^k_t}'{w})^2}.$$
Since the crossed term is a constant at $p$, the right hand side gives an equivalent objective 
for computing $S_p$ in \eqref{eq:wKMproc}. Using ${\cal K} = \phi'\phi$ and 
indicator vector ${\mathbf 1}_p$ for element $p$ we get generalization of \eqref{eq:imp_kKMproc} 
\begin{equation} \label{eq:wKKMproc}
\left(\parbox{9.5ex}{\centering\bf\small weighted \kKM/ \\ procedure}\right) \;\;
S_p\;\; \leftarrow\;\; \arg\min_k 
\frac{{S^k_t}'W{\cal K}WS^k_t}{({S^k_t}'{w})^2}   -   2\frac{{{\mathbf 1}_p}'{\cal K}WS^k_t}{{S^k_t}'w}.
\end{equation}
In contrast to procedure \eqref{eq:wKMproc}, this approach has iterations of quadratic complexity 
${\cal O}(|\Omega|^2)$. However,
it avoids the explicit use of high-dimensional embeddings $\phi_p$ replacing them by the kernel matrix 
${\cal K}$ in all computations, {\em a.k.a.} the {\em kernel trick}. 

Generalizing Theorem~\ref{th:kKM_bound1} in Sec.~\ref{sec:KM_and_BO} 
we can interpret weighted \KM/ procedures (\ref{eq:wKMproc},\ref{eq:wKKMproc})  as linear bound optimization.
\begin{theor}[{\bf bound for} (\ref{eq:wkKM0},\ref{eq:kKMweighted})] \label{th:wKM_bound}
Weighted \KM/ procedures \eqref{eq:wKMproc} or \eqref{eq:wKKMproc} can be seen as  
bound optimization methods for weighted K-means objective $F^w(S)$ \eqref{eq:wkKM0} or \eqref{eq:kKMweighted} using auxiliary function
\begin{equation} \label{eq:wKMbound}
a_t(S) \;\;:=\;\;  F^w(S,\mu^w_t)
\end{equation}
at any current segmentation $S_t = \{S^k_t\}$ with means $\mu^w_t = \{\mu^w_{S^k_t}\}$.
\end{theor}
\begin{proof}
Indeed,  \eqref{eq:variousA} implies $a_t(S) \geq F^w(S)$. Since $a_t(S_t) =F^w(S_t)$ then $a_t(S)$ is a proper
bound for $F^w(S)$. Re-segmentation step \eqref{eq:wKMproc} produces optimal segments $S_{t+1}$ 
minimizing the bound  $a_t(S)$. Re-centering step minimizing $F^w(S_{t+1},m)$ 
for fixed segments produces means $\mu^w_{t+1}$ defining the bound $a_{t+1}(S)$ for the next iteration.  
\end{proof}

Since algebraic formulations \eqref{eq:wmixedA} and \eqref{eq:AAweighted} 
omit the same constant we also get the following Corollary.
\begin{coro}[{\bf bound for} (\ref{eq:wkKM},\ref{eq:AAweighted})] \label{cor:wKKM_bound}
Weighted \KM/ procedures  \eqref{eq:wKMproc} or \eqref{eq:wKKMproc} can be seen as  
bound optimization methods for \wkKM/ objective $F^w(S)$ in \eqref{eq:AAweighted} using auxiliary function
\begin{equation} \label{eq:wkKMbound1}
a_t(S) \;\; := \;\; \sum_k \left(  w \|\mu^w_{S^k_t}\|^2 - 2W \phi' \mu^w_{S^k_t}   \right)' S^k     
\end{equation}
\begin{equation} \label{eq:wkKMbound2}
\equiv \;\;\;\;  \sum_k \left(  w \frac{{S^k_t}'W{\cal K} WS^k_t}{(w' S^k_t)^2}  -   W {\cal K} WS^k_t  \frac{2 }{w'S^k_t}   \right)' S^k 
\end{equation}
at any current segmentation $S_t := \{S^k_t\}$ with means $\mu^w_t := \{\mu^w_{S^k_t}\}$.
\end{coro}
\begin{proof}
The first expression follows from Th.~\ref{th:wKM_bound} and formula \eqref{eq:wmixedA} for $F^w(S,m)$ at $m=\mu^w_t$.
Also, \eqref{eq:wmean} and ${\cal K} = \phi'\phi$ imply the second expression for bound $a_t(S)$.
\end{proof}

The second expression for $a_t(S)$ in Corollary~\ref{cor:wKKM_bound} allows to obtain a linear bound
for \NC/ objective \eqref{eq:NC2}. For simplicity, assume positive definite  affinity $A$. 
As follows from \cite{bach:nips03,dhillon2004kernel,Dhillon:PAMI07} and a simple derivation in our Sec.~\ref{sec:NC-to-wKKM},
normalized cut \eqref{eq:NC2} with p.d. affinity $A$ is equivalent to \wkKM/  (\ref{eq:wkKM},\ref{eq:AAweighted})
with weights  and kernel  in  \eqref{eq:Atilde}
$$w=d:=A{\mathbf 1} \;\;\;\;\;\;\;\;\;\;\;  \text{and}   \;\;\;\;\;\;\;\;\;\;\;\;\;\;\;  {\cal K} = W^{-1} A W^{-1}.$$
Then,  \eqref{eq:wkKMbound2} implies the following linear bound for \NC/
$$\sum_k   \left(d \;\frac{{S^k_t}'AS^k_t}{(d' S^k_t)^2} - AS^k_t \frac{2}{d'S^k_t} \right)' S^k$$ 
that agrees with the {\em kernel bound} for \NC/ in Table~\ref{tab:kernel_bounds}.

\begin{figure*}
\centering
\includegraphics[width=5.4in]{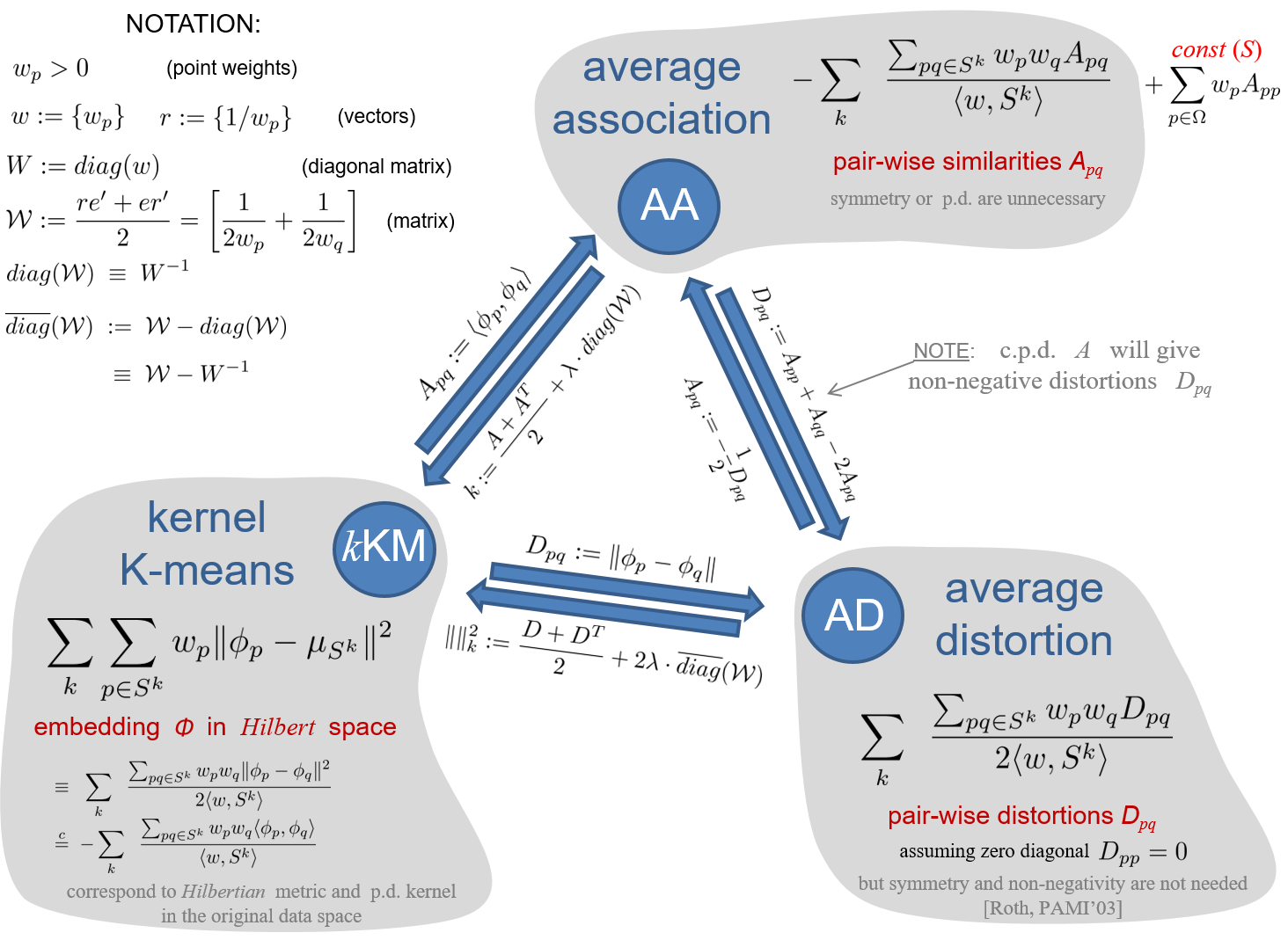} 
\centering
 \caption{Equivalence of {\em kernel K-means} (\kKM/), {\em average distortion} (\AD/), {\em average association} (\AA/)
for the general case of weighted points. Typically, \kKM/ is associated with {\em positive semi-definite} (p.s.d.) or 
{\em conditionally positive-definite} (c.p.d.) kernels \cite{Scholkopf:NIPS01} and Hilbertian distortions \cite{Hein:PR04}. 
The above formulations of \AA/ and \AD/ make no assumptions for association matrix $A$ and distortions $D$ 
except for zero diagonal in $D$. Then, equivalence of \AD/ and \AA/ objectives (up to a constant) is straightforward. 
Roth et al. \cite{buhmann:pami03} reduce non-weighted case of \AD/ to \kKM/. 
For arbitrary $D$ they derive Hilbertian distortion $\|\|^2_k$ with an equivalent \AD/ energy (up to a constant) 
and explicitly construct the corresponding embedding $\phi$. We show Hilbertian metric $\|\|^2_k$ for 
the general weighted case of \AD/, see \AD/$\rightarrow$\kKM/ above. 
Dhillon et al. \cite{dhillon2004kernel,Dhillon:PAMI07} prove that {\em normalized cuts} 
is a special case of weighted \kKM/ and construct the corresponding p.d. kernel. Sec.\ref{sec:NC-to-wKKM} shows a simpler
reduction of normalized cuts to weighted \kKM/. Similarly to \cite{buhmann:pami03}, 
an equivalent p.d. kernel could be constructed for any association matrix $A$, see \AA/$\rightarrow$\kKM/ above.
Note that the formulas for $A$-equivalent p.d. kernel and $D$-equivalent Hilbertian metric require some sufficiently 
large diagonal shift $\ds$. Roth et al. \cite{buhmann:pami03} relate proper $\ds$ to the smallest eigenvalue of $A=-\frac{D}{2}$.
Our weighted formulation requires the eigenvalue of $diag(\mathcal{W})^{-\frac{1}{2}}\cdot A \cdot diag(\mathcal{W})^{-\frac{1}{2}}$.
\label{fig:dclustering2}}
\end{figure*}

{\bf Equivalence of weighted \AA/ and \AD/ to \wkKM/:}
Figure~\ref{fig:dclustering2} extends Figure~\ref{fig:dclustering1} by relating 
weighted generalizations of standard pairwise clustering objectives to \wkKM/.
The equivalence of objectives in Figure~\ref{fig:dclustering2} can be verified by simple algebra.
One additional simple property below is also needed. It is easy to prove.
\begin{prop} {\bf (\eg Roth et al. \cite{buhmann:pami03})} \label{fact:psd_diagonal_shift}
For any symmetric matrix $M$ define $$\tilde{M} := M+ \ds I$$
where $I$ is an identity matrix. 
Then, matrix $\tilde{M}$ is positive semi-definite (psd) for any scalar $\ds \geq - \lambda_0(M)$ 
where $\lambda_0$ is the smallest eigenvalue of its argument.  
\end{prop}

 \begin{figure*}
\centering
\includegraphics[width=6in]{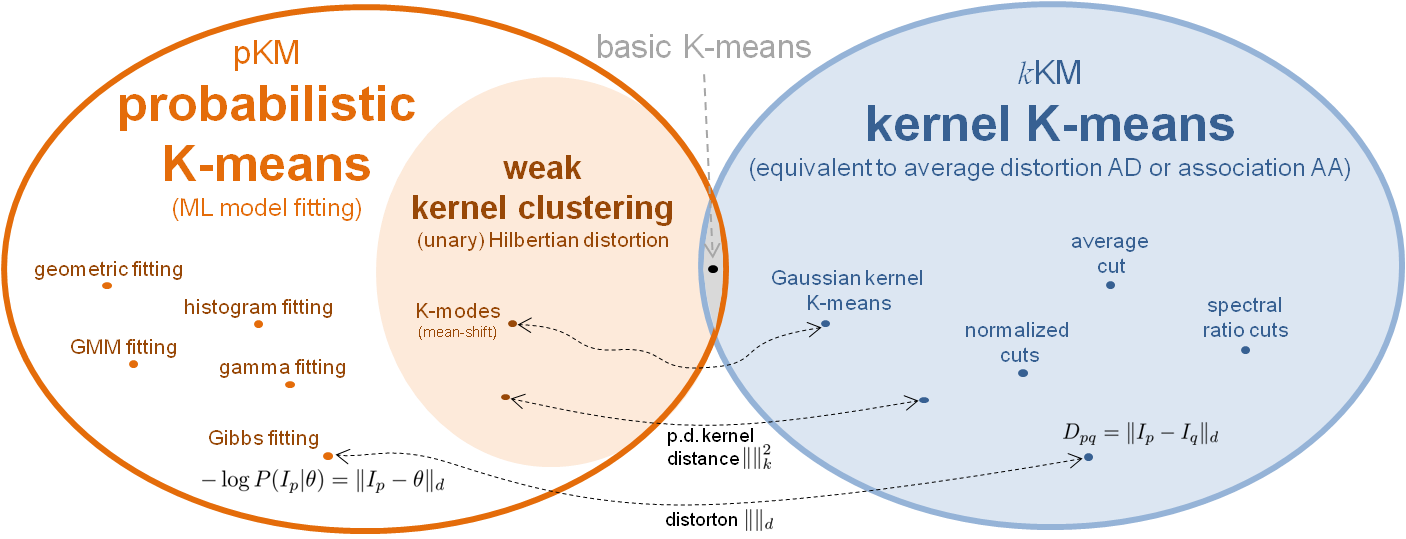} 
\begin{tabular}{ c@{\hspace{12ex}}c }  
{\color{magenta} pointwise distortions (likelihoods) energy} & 
{\color{blue} pairwise distortions energy} \\
(A) $\;\; \sum_k \sum_{p \in S^k} w_p \|I_p-\theta_{S^k} \|_d $  & 
(B) $\;\; \sum_k \frac{\sum_{pq \in S^k} w_p w_q \|I_p-I_q \|_d}{2 \sum_{p\in S^k} w_p} \;\equiv\; 
\sum_k \frac{{S^k}'WDWS^k}{2w'S^k}$
\end{tabular}
\centering
 \caption{Clustering with (A) pointwise and (B) pairwise distortion energies generalizing \eqref{eq:distortion} and \eqref{eq:AD} 
for points with weights $w=\{w_p\}$. Pointwise distortion relates a data point and a model
as log-likelihood function $\|I_p-\theta\|_d = -\ln\P(I_p|\theta)$. Pairwise distortion is defined by matrix $D_{pq}=\|I_p-I_q\|_d$.
Weighted \AD/ or  \AA/ for arbitrary metrics are equivalent to weighted \kKM/, see Figures \ref{fig:dclustering1} and \ref{fig:dclustering2}. 
As shown in  \cite{buhmann:pami03,dhillon2004kernel}  {\em average cut}, {\em normalized cut} \cite{Shi2000}, and 
{\em spectral ratio cut} \cite{sratiocut:1994} are examples of (weighted) \kKM/.  
\label{fig:dclustering}}
\end{figure*}


\renewcommand{\theequation}{B-\arabic{equation}}
  \setcounter{equation}{0}
 \section*{APPENDIX B (Weak kernel K-means)}
 
For Hilbertian distortions $\|\|_d=\|\|^2_k$ with p.s.d. kernels we can show that pairwise 
\kKM/ approach \eqref{eq:kKmeans2} is ``stronger'' than a pointwise \pKM/ approach \eqref{eq:distortion} 
using the same metric. In this case \pKM/ can be called {\em weak kernel K-means}, see Figure \ref{fig:dclustering}. 
Equivalent \kKM/ formulation \eqref{eq:kKmeans1} guarantees more complex decision boundaries, 
see Fig.\ref{fig:synexp}(h), compared to \pKM/ energy \eqref{eq:distortion} 
\begin{equation} \label{eq:weak_kKmeans}
\sum_k \sum_{p \in S^k} \|I_p-m_k \|^2_k
\end{equation}
$$ \equiv \; \sum_k \sum_{p \in S^k}\|\phi (I_p)-\phi(m_k) \|^2 \quad $$
with isometric kernel distance $\|\|_k$ and some point $m_k$ 
in the original space, Fig.\ref{fig:synexp}(g). Indeed, any $m_k$ in the original space corresponds to some search point $\phi(m_k)$
in the high-dimensional embedding space, while the opposite is false. Thus, optimization of \eqref{eq:kKmeans2} 
and \eqref{eq:kKmeans1} has larger search space than \eqref{eq:weak_kKmeans}. It is also easy to check that
energy \eqref{eq:weak_kKmeans} is an upper bound for \eqref{eq:kKmeans2} at any $S$.
For this reason we refer to distortion energy \eqref{eq:weak_kKmeans}
with kernel distance $\|\|_k$ and explicit $m_k$ in the original space as {\em weak kernel K-means}. 
Pointwise energy \eqref{eq:weak_kKmeans} is an example of \pKM/ \eqref{eq:distortion}, while
pairwise energy \eqref{eq:kKmeans2} with the same kernel metric is \kKM/.

Note that weak kernel K-means \eqref{eq:weak_kKmeans}  for Gaussian kernel corresponds to
{\em K-modes} closely related to {\em mean-shift}  \cite{cheng:pami95,dorin:pami02}, as discussed below. 
Some results comparing K-modes (weak \kKM/)
to regular \kKM/ are shown in Fig.\ref{fig:synexp}(g,h). Figure \ref{fig:dclustering} illustrates general relations between 
kernel K-means \eqref{eq:kKmeans2} and probabilistic K-means (\ref{eq:distortion},\ref{eq:pKmeans}). It includes
a few examples discussed earlier and some more examples discussed below.

{\bf K-modes and mean-shift}: 
Weak kernel K-means using unary formulation \eqref{eq:weak_kKmeans}
with kernel distance $\|\|_k$ and explicit optimization of $m_k$ in the original data space 
is closely related to {\em K-modes} approach to clustering continuous \cite{carreira:arXiv13} 
or discrete \cite{huang1998,chaturvedi2001} data. 
For example, for Gaussian kernel distance $\|\|_k$ energy \eqref{eq:weak_kKmeans} becomes
\begin{equation*}
- \sum_k\sum_{p\in S^k} e^{\frac{-\|I_p-m_k\|^2}{2\sigma^2}}
\end{equation*}
or, using Parzen densities $\P_{\sigma} (\;\cdot\; |S^k)$ for points $\{I_p|p\in S^k\}$,
\begin{equation}\label{eq:kmodes}
= \;\;\;- \sum_k \; |S^k|\cdot \P_{\sigma} (m_k|S^k) .
\end{equation}
Clearly, optimal $m_k$  are modes of Parzen densities in each segment. 
K-modes objective \eqref{eq:kmodes} can be seen as an energy-based 
formulation of mean-shift clustering \cite{cheng:pami95,dorin:pami02} with a fixed number of clusters. 
Formal objective allows to combine color clustering via \eqref{eq:kmodes} with 
geometric regularization in the image domain \cite{Ismail:tip10}. If needed, the number of clusters (modes) 
can be regularized by adding {\em label cost} \cite{LabelCosts:IJCV12}. 
In contrast, mean-shift segmentation \cite{dorin:pami02} clusters RGBXY space combining 
color and spatial information.  The number of clusters is indirectly controlled by the bandwidth.

Note that K-modes energy \eqref{eq:kmodes}
follows from a weak \kKM/ approach \eqref{eq:weak_kKmeans} for arbitrary positive normalized kernels.
Such kernels define different Parzen densities, but they all lead to energy \eqref{eq:kmodes} where optimal 
$m_k$ are modes of the corresponding densities. Therefore, different kernels in \eqref{eq:weak_kKmeans}
give different modes in \eqref{eq:kmodes}.

Many optimization methods can be used for K-modes energy. 
For example, it is possible to use iterative (block-coordinate descent) approach
typical of K-means methods: one step reclusters points and the other step
locally refinement the modes, \eg using mean-shift operation \cite{Ismail:tip10}.
For better optimization, local refinement of the mode in each cluster can be replaces by the best mode search 
tracing all points within each cluster using mean-shift. RANSAC-like sampling procedure can be used for some
compromise between speed and quality.
It is also possible to use exhaustive search for the strongest mode in each cluster over observed discrete features
and then locally refine each cluster's mode with mean-shift.

It is also interesting that discrete version of K-modes for histograms \cite{huang1998,chaturvedi2001} 
define modes $m_k = (m^1_k,...,m^j_k,...) $ 
combining {\em marginal modes} for all attributes or dimensions $j$. 
Implicitly, they use distortion $\|\|_k$ for discrete kernel $k(x,y) = \sum_j [x^j=y^j]$ where $[\cdot]$
are {\em Iverson brackets}. Marginal modes could be useful for aggregating sparse high-dimensional data.

Analogously, we can also define a continuous kernel for {\em marginal modes} as
\begin{equation} \label{eq:mmodes}
 k(x,y) = \sum_j e ^{\frac{-(x^j-y^j)^2}{2\sigma^2}}. 
\end{equation}
Note that this is different from the standard Gaussian kernel
$$ e ^{\frac{-\|x-y\|^2}{2\sigma^2}}   = \prod_j e ^{\frac{-(x^j-y^j)^2}{2\sigma^2}},$$
which leads to regular modes energy \eqref{eq:kmodes}. It is easy to check that kernel \eqref{eq:mmodes}
corresponds to weak \kKM/ energy
\begin{equation*}
- \sum_k \sum_j \sum_{p\in S^k} e^{\frac{-(I_p^j-m^j_k)^2}{2\sigma^2}} 
\end{equation*}
$$ = \;\;\;- \sum_k |S_k|\cdot\sum_j \P_{\sigma}^j (m^j_k|S^k) $$
where $P_{\sigma}^j$ is a marginal Parzen density for dimension $j$.

\renewcommand{\theequation}{C-\arabic{equation}}
  \setcounter{equation}{0}
 \section*{APPENDIX C (Entropy Clustering)}

First, we show that  \eqref{eq:pKmeans} reduces to \eqref{eq:entropy} for descriptive models.
Indeed, assume $\P(\theta)\equiv\P (\cdot|\theta)$ is a continuous density of a sufficiently descriptive class 
(e.g. GMM). For any function $f(x)$ Monte-Carlo estimation gives for any subset $\S\subset\Omega$
$$\sum_{p\in\S} f (I_p) \approx |\S|\cdot \int f(x)\dS(x) dx \equiv |\S|\cdot\langle f,\dS \rangle$$ 
where $\dS$ is a ``true'' density for intensities $\{I_p|p\in\S\}$ and $\langle,\rangle$ is a dot product. 
If $f =  - \log \P (\thS)$ and $\dS\approx\P (\thS)$ then \eqref{eq:pKmeans} implies  \eqref{eq:entropy} for
differential entropy $H(\S):= H(\P (\thS))$. 
For histograms $\P_h(\S)\equiv \P_h (\cdot|\S)$ entropy-based interpretation  \eqref{eq:entropy} of \eqref{eq:pKmeans} 
is exact for discrete entropy 
$$H(\S):=-\sum_x \P_h (x|\S)\cdot \log \P_h (x|\S) \equiv - \langle \P_h(\S) ,  \log \P_h(\S) \rangle.$$

Intuitively, minimization of the entropy criterion \eqref{eq:entropy} favors clusters with tight or ``peaked'' distributions.
This criterion is widely used in categorical clustering \cite{li:icml04} or decision trees \cite{breiman1996,Louppe2013}
where the entropy evaluates histograms over ``naturally'' discrete features. Below we discuss limitations of the entropy
clustering criterion with either discrete histograms or continuous GMM densities in the context of color feature spaces.

\begin{figure}[t!]
        \centering
        \includegraphics[width=0.9\columnwidth]{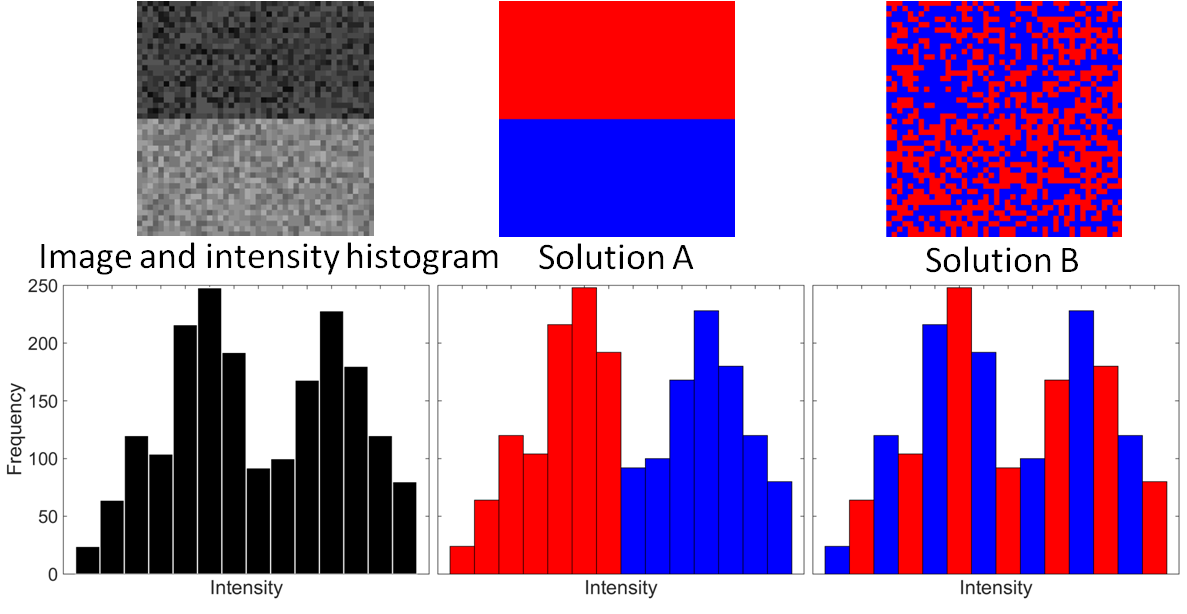}
                \caption{{\em Histograms in color spaces.} Entropy criterion \eqref{eq:entropy} with histograms 
can not tell a difference between A and B: bin permutations do not change the histogram's entropy.}
        \label{fig:synpatch}
\end{figure}

{\bf The case of histograms:} In this case the key problem for color space clustering is illustrated in Fig.\ref{fig:synpatch}.
Once continuous color space is broken into bins, the notion of proximity between the colors in the nearby bins is lost.
Since bin permutations do not change the histogram entropy, criterion \eqref{eq:entropy}
can not distinguish the quality of clusterings A and B in Fig.\ref{fig:synpatch}; some
permutation of bins can make B look very similar to A.

{\bf The case of GMM densities:}  In this case the problem of entropy clustering \eqref{eq:entropy} is different.
In general, continuous density estimators commonly use Gaussian kernels, which preserve the notion of continuity in the color space. 
Indeed, the (differential) entropy for any reasonable continuous density estimate will see a significant difference
between the clusters in A and B, see Fig.\ref{fig:synpatch}. 

We observe that the main issue for entropy criterion \eqref{eq:entropy} with GMM densities is related to optimization problems.
In this case high-order energies \eqref{eq:entropy} or \eqref{eq:pKmeans} require joint optimization of discrete variables $S_p$ 
and a large number of additional continuous parameters for optimum GMM density $\P (\cdot|\theta_{\S})$. That is, the use of 
complex parametric probability models leads to complex high-order {\em mixed} objective functions. Typical block coordinate descent
methods \cite{Zhu:96,GrabCuts:SIGGRAPH04} iterating optimization of $S$ and $\theta$ are sensitive to local minima, 
see Figures \ref{fig:firstexamples} and \ref{fig:synexp}(e). Better solutions like Figure \ref{fig:synexp}(f) have lower energy, 
but they can not be easily found unless initialization is very good. 

These problems of \pKM/ with histograms or GMM may explain why descriptive model fitting is not common in the learning community 
for clustering high-dimensional continuous spaces. Instead of \pKM/
they often use a different extension of K-means, that is {\em kernel K-means} (\kKM/) or related pairwise clustering criteria
like Normalize Cut (\NC/), see Sec.\ref{sec:kKM} and \ref{sec:NC-to-wKKM}.


\renewcommand{\theequation}{D-\arabic{equation}}
  \setcounter{equation}{0}
\section*{APPENDIX D (Pseudo Bound Lemma)}
  \label{app:pbo}

\begin{lem} \label{lm:pseudo_bound}
Consider function $e:\{0,1\}^{|\Omega|}\rightarrow {\mathcal R}^1$ defined by any symmetric matrix $A$ 
and (strictly) positive vector $w$ as $$e(X) = -\frac{X'AX}{w'X}.$$
Function $T_t$ is a pseudo-bound of $e(X)$ at $X_t$ for $W:=diag(w)$ 
\begin{equation} \label{eq:pseudo-bound-T}
T_t(X,\ds) \;:=\; \nabla e(X_t)' X \;+\;\ds \left( \frac{({\bf 1}-2X_t)'WX}{w' X_t} + 1 \right)
\end{equation}
where $\nabla e(X) \;=\; w \;\frac{X'AX}{(w'X)^2} \;-\; A X \frac{2}{w'X}$.
Furthermore, $T_t(X, \ds)$ is an auxiliary function for $e(X)$ for all 
$\ds\geq -\lambda_0(W^{\frac{-1}{2}}AW^{\frac{-1}{2}})$
where $\lambda_0$ denotes the smallest eigenvalue of the matrix.
\end{lem}
\begin{proof}
Diagonal shift $\ds W$ for matrix $A$ defines function 
$$\e_{\ds}(X) := -\frac{X'(\ds W + A)X}{w'X} \equiv e(X) - \ds\frac{X'WX}{w'X}.$$
According to Lemma~\ref{lm:concave} function $\e_{\ds}$ is concave for any $\ds\geq -\lambda_0$ 
since $(\ds W + A)$ is p.s.d. for such $\ds$. Thus, \eqref{eq:Tbound} defines a Taylor-based 
linear upper bound for $\e_{\ds}$ for $\ds\geq -\lambda_0$
$$ \nabla \e_{\ds} (X_t)'\,X.$$
We have $\e_{\ds}(X) = e(X) - \ds$ for Boolean $X$ where $X'WX=w'X$. Thus,
the following upper bound is valid for optimizing $e(X)$ over $X\in\{0,1\}^{|\Omega|}$ 
at $\ds\geq -\lambda_0$
\begin{equation} \label{eq:ebound}
 T_t(X,\ds) \;=\; \nabla \e_{\ds} (X_t)'\,X + \ds.
\end{equation}
where the definition of $\e_{\ds}$ above yields gradient expression
\begin{eqnarray}
\nabla \e_{\ds} (X_t) &=& \nabla e(X_t) -\ds \frac{2\,(\cancel{w'X_t})\,WX_t-(\cancel{X_t'WX_t})\,w}{(w'X_t)^{\cancel{2}}}   \nonumber \\
& = & \nabla e(X_t) + \ds  \frac{W({\bf 1} - 2X_t)}{w'X_t}       \nonumber
\end{eqnarray}
where $W{\bf 1}=w$ for matrix $W=diag(w)$ and $X'_tWX_t=w'X_t$ for any $X_t$ since all iterations explore only Boolean solutions. 
\end{proof}


\renewcommand{\theequation}{E-\arabic{equation}}
  \setcounter{equation}{0}
\section*{APPENDIX E (proof of {\em Gini Bias} Theorem \ref{th:gini_bias})}
  \label{app:gini}
Let $d_{\Omega}$  be a continuous probability density function over domain $\Omega\subseteq{\mathcal R}^n$
defining conditional density 
\begin{equation} \label{eq:ds}
\dS(x):=d_{\Omega}(x|x\in\S)
\end{equation}
for any non-empty subset $\S\subset\Omega$ and expectation 
$$\E z := \int z(x)d_\Omega(x)dx$$
for any function $z:\Omega\to{\mathcal R}^1$. 

Suppose $\Omega$ is partitioned into two sets $\S$ and $\barS$ such that $\S\cup\barS=\Omega$ 
and $\S\cap\barS=\emptyset$. Note that $\S$ here and in the statement of Theorem \ref{th:gini_bias} is not 
a discrete set of observations, which is what $\S$ means in the rest of the paper.
Theorem \ref{th:gini_bias} states a property of a fully continuous version of \ Gini criterion~\eqref{eq:gini_criterion}
that follows from an additional application of Monte-Carlo estimation allowing to replace discrete set cardinality $|\S|$ 
by probability $w$ of a continuous subset $\S$
$$w:=\int_Sd_\Omega(x)dx=\int d_\Omega(x)\cdot [x\in S]dx = \E [x\in S].$$
Then, minimization of $E_G(S)$ in \eqref{eq:gini_criterion} corresponds to maximization of the following objective function
\begin{equation} \label{eq:gini:objective}
    L(\S) := w \int \dS^2(x)dx + (1-w) \int \dSbar^2(x)dx.
\end{equation}
Note that conditional density $\dS$ in \eqref{eq:ds}  can be written as
\begin{equation}    \label{eq:gini:conditional}
    \dS(x)\;=\;d_\Omega(x)\cdot \frac{[x\in \S]}{w}
\end{equation}
where $[\cdot]$ is an indicator function. Eqs. \eqref{eq:gini:conditional} and \eqref{eq:gini:objective} give
$$L(\S)=\frac1{w}\int d_\Omega^2(x)[x\in \S]dx + \frac1{1-w}\int d_\Omega^2(x)[x\in \barS]dx. $$
Introducing notation $$I:= [x\in \S] \;\;\;\;\;\;\mbox{and}\;\;\;\;\;\; F:= d_\Omega(x)$$ allows to further rewrite objective function $L(\S)$ as
$$L(\S)=\frac{\E IF}{\E I}+\frac{\E F  (1- I)}{1-\E I}.$$

Without loss of generality assume that
$\frac{\E F (1- I)}{1-\E I} \le \frac{\E FI}{\E I}$ (the opposite case would yield a similar result). We now need the following lemma.
\begin{lem}\label{lm:ratio-law} 
Let $a,b,c,d$ be some positive numbers, then $$\frac ab \le \frac cd \implies \frac ab \le \frac{a+c}{b+d} \le \frac cd.$$
\begin{proof} Use reduction to a common denominator. \end{proof}
\end{lem}
\noindent Lemma \ref{lm:ratio-law} implies inequality
\begin{equation} \label{eq:gini:ratio}
\frac{\E F (1- I)}{1-\E I} \le \E F \le\frac{\E FI}{\E I}
\end{equation}
which is needed to prove the Proposition below.
\begin{prop} \textbf {(Gini-bias)}  \label{prop:appendixD}
Assume that subset $\S_\varepsilon \subset\Omega$ is
\begin{equation} \label{eq:se}
\S_\varepsilon:=\{ x:d_\Omega(x) \ge \sup_xd_\Omega(x)-\varepsilon\}.
\end{equation}
Then
\begin{equation} \label{eq:gini:main}
\sup_\S L(\S)=\lim_{\varepsilon\to0} L(\S_\varepsilon)=\E d_\Omega + \sup_xd_\Omega(x).
\end{equation}
 
\begin{proof} Due to monotonicity of expectation we have
\begin{align}
&  \frac{\E F I}{\E I} \le \frac{\E \left(I \sup_xd_\Omega(x)\right)}{\E I} =
 \sup_xd_\Omega(x). \label{eq:gini:up-incr}
\end{align}
Then \eqref{eq:gini:ratio} and \eqref{eq:gini:up-incr} imply
\begin{align}
L(\S) &= \frac{\E FI}{\E I} + \frac{\E F (1- I)}{1-\E I} \le \sup_xd_\Omega(x)+\E F. \label{eq:gini:bound}
\end{align}
That is, the right part of \eqref{eq:gini:main} is an upper bound for $L(\S)$.

Let $I_\varepsilon\equiv[x\in\S_\varepsilon]$. It is easy to check that
\begin{equation} \label{eq:lim1}
\lim_{\varepsilon\to0}\frac{\E F (1- I_\varepsilon)}{1-\E I_\varepsilon}=\E F.
\end{equation}
Definition \eqref{eq:se} also implies 
$$\lim_{\varepsilon\to0}\frac{\E F  I_\varepsilon}{\E I_\varepsilon} \ge
 \lim_{\varepsilon\to0}\frac{\E (\sup_xd_\Omega(x)-\varepsilon)  I_\varepsilon}{\E I_\varepsilon} =\sup_xd_\Omega(x).$$
This result and \eqref{eq:gini:up-incr} conclude that 
\begin{equation} \label{eq:lim2}
\lim_{\varepsilon\to0}\frac{\E F  I_\varepsilon}{\E I_\varepsilon}=\sup_xd_\Omega(x).
\end{equation}
Finally, the limits in \eqref{eq:lim1} and \eqref{eq:lim2} imply 
\begin{align*}
\lim_{\varepsilon\to0} L(\S_\varepsilon)&=  \lim_{\varepsilon\to0} \frac{\E F (1- I_\varepsilon)}{1-\E I_\varepsilon}+\lim_{\varepsilon\to0} \frac{\E F I_\varepsilon }{\E I_\varepsilon}  \\
&= \E d_\Omega + \sup_xd_\Omega(x).
\end{align*}
This equality and bound \eqref{eq:gini:bound} prove \eqref{eq:gini:main}.
\end{proof}
\end{prop}

\section*{Acknowledgements}
We would like to greatly thank Carl Olsson (Lund University, Sweden) for many hours of stimulating discussions, 
as well as for detailed feedback and valuable recommendations at different stages of our work. We also
appreciate his tremendous patience when our thoughts were much more confusing than they might be now.
Ivan Stelmakh (PhysTech, Russia) also gave helpful feedback on our draft and caught several errors. 
Anders Eriksson (Lund University, Sweden) helped with related work on \NC/ with constraints.  
We also thank Jianbo Shi (UPenn, USA) for his feedback and 
his excellent and highly appreciated spectral-relaxation optimization code for normalized cuts.

{\small
\bibliographystyle{IEEEtran}
\bibliography{KC_2016} 
}

\begin{IEEEbiography}[{\includegraphics[width=1in,height=1.25in,clip,keepaspectratio]{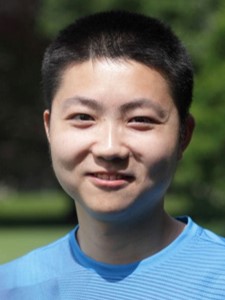}}]{Meng Tang} is a PhD candidate in computer science at the University of Western Ontario, Canada, supervised by Prof. Yuri Boykov. He obtained MSc in computer science in 2014 from the same institution for his thesis titled "Color Separation for Image Segmentation". Previously in 2012 he received B.E. in Automation from the Huazhong University of Science and Technology, China. He is interested in image segmentation and semi-supervised data clustering. He is also obsessed and has  experiences on discrete optimization problems for computer vision and machine learning.
\end{IEEEbiography}

\begin{IEEEbiography}[{\includegraphics[width=1in,height=1.25in,clip,keepaspectratio]{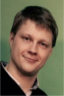}}]{Dmitrii Marin} received Diploma of Specialist from the Ufa State Aviational Technical University in 2011, and M.Sc. degree in Applied Mathematics and Information Science from the National Research University Higher School of Economics, Moscow, and graduated from the Yandex School of Data Analysis, Moscow, in 2013. In 2010 obtained a certificate of achievement at ACM ICPC World Finals, Harbin. 
His is a PhD candidate at the Department of Computer Science, University of Western Ontario under supervision of Yuri Boykov. His research is focused on designing general unsupervised and semi-supervised methods for accurate image segmentation and object delineation.
\end{IEEEbiography}

\begin{IEEEbiography}[{\includegraphics[width=1in,height=1.25in,clip,keepaspectratio]{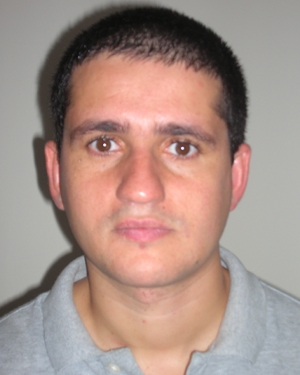}}]{Ismail Ben Ayed} received the PhD  degree
 (with the highest honor) in computer vision from the Institut National de la Recherche Scientifique (INRS-EMT), Montreal, QC, in 2007. 
He is currently Associate Professor at the Ecole de Technologie Superieure (ETS), University of Quebec, where he holds a research chair on Artificial Intelligence in Medical Imaging. Before joining the ETS, he worked for 8 years as a research scientist at GE Healthcare, London, ON, conducting research in medical image analysis. He also holds an adjunct professor appointment at the University of Western Ontario (since 2012). Ismail's research interests are in computer vision, optimization, machine learning and their potential applications in medical image analysis.
\end{IEEEbiography}

\begin{IEEEbiography}[{\includegraphics[width=1in,height=1.25in,clip,keepaspectratio]{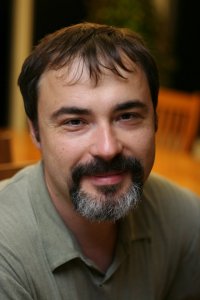}}]{Yuri Boykov}
received "Diploma of Higher Education" with honors at Moscow Institute of Physics and Technology (department of Radio Engineering and Cybernetics) in 1992 and completed his Ph.D. at the department of Operations Research at Cornell University in 1996.
He is currently a full professor at the department of Computer Science at the University of Western Ontario. His research is concentrated in the area of computer vision and biomedical image analysis. In particular, he is interested in problems of early vision, image segmentation, restoration, registration, stereo, motion, model fitting, feature-based object recognition, photo-video editing and others. He is a recipient of the Helmholtz Prize (Test of Time) awarded at International Conference on Computer Vision (ICCV), 2011 and Florence Bucke Science Award, Faculty of Science, The University of Western Ontario, 2008.
\end{IEEEbiography}

\end{document}